\documentclass[11pt]{article}
\usepackage[utf8]{inputenc}
\usepackage[T1]{fontenc}

\usepackage[verbose=true,
letterpaper,
textheight=8in,
textwidth=6in,
margin=1in,
headheight=12pt,
headsep=25pt,
footskip=30pt]{geometry}

\usepackage{microtype}
\usepackage{mathtools}
\usepackage{booktabs}
\usepackage{natbib}
\usepackage{graphicx}
\usepackage{subcaption}
\usepackage{amsmath,amssymb,amsfonts,amsxtra,bm}

\usepackage{amsthm}
\makeatletter
\newtheorem*{rep@theorem}{\rep@title}
\newcommand{\newreptheorem}[2]{%
\newenvironment{rep#1}[1]{%
 \def\rep@title{#2 \hspace{0.05cm}\ref{##1}}%
 \begin{rep@theorem}}%
 {\end{rep@theorem}}}
\makeatother

\usepackage{xspace}
\usepackage{paralist}
\usepackage{enumerate}
\usepackage{enumitem}
\usepackage{hyperref}
\usepackage{url}
\usepackage{colortbl}
\usepackage{makecell,multirow}       %
\usepackage{nicefrac}       %
\usepackage{thmtools}  
\usepackage{xspace}
\usepackage{footnote, tablefootnote}
\usepackage{float}
\floatstyle{plaintop}
\restylefloat{table}
\usepackage{epstopdf}
\usepackage{latexsym}
\usepackage{bbm}
\usepackage{cleveref}
\usepackage{algorithm,algorithmic}
\usepackage{ss}
\usepackage{tikz}
\usepackage{amssymb}
\usepackage{tcolorbox}
\usepackage{wrapfig}
\usepackage{pifont}
\usepackage{tabularx,ragged2e}

\definecolor{ao}{rgb}{0.0, 0.5, 0.0}
\definecolor{LightCyan}{rgb}{0.88,1,1} 
\definecolor{Lightpurple}{rgb}{0.9,0.9,1} 

\makeatletter
\newcommand{\leqnomode}{\tagsleft@true}
\newcommand{\reqnomode}{\tagsleft@false}

\makeatother

\numberwithin{equation}{section}
\clearpage{}%
\newcommand{\Ac}{\mathcal{A}}
\newcommand{\Xc}{\mathcal{X}}
\newcommand{\Yc}{\mathcal{Y}}

\newcommand{\Sc}{\mathcal{S}}
\newcommand{\Zc}{\mathcal{Z}}
\newcommand{\Pc}{\mathcal{P}}

\newcommand{\Mc}{\mathcal{M}}

\newcommand{\Fc}{\mathcal{F}}
\newcommand{\Oc}{\mathcal{O}}
\newcommand{\gr}{\nabla}
\newcommand{\E}{\mathbb{E}}

\newcommand{\reals}{\mathbb{R}}

\newcommand{\ip}[2]{\langle {#1},\, {#2} \rangle}
\newcommand{\norm}[1]{\| {#1} \|}

\DeclareMathOperator{\Proj}{Proj}

\makeatletter
\newsavebox\myboxA
\newsavebox\myboxB
\newlength\mylenA
\newcommand*\xbar[2][0.75]{%
    \sbox{\myboxA}{$\m@th#2$}%
    \setbox\myboxB\null
    \ht\myboxB=\ht\myboxA%
    \dp\myboxB=\dp\myboxA%
    \wd\myboxB=#1\wd\myboxA
    \sbox\myboxB{$\m@th\overline{\copy\myboxB}$}
    \setlength\mylenA{\the\wd\myboxA}
    \addtolength\mylenA{-\the\wd\myboxB}%
    \ifdim\wd\myboxB<\wd\myboxA%
       \rlap{\hskip 0.5\mylenA\usebox\myboxB}{\usebox\myboxA}%
    \else
        \hskip -0.5\mylenA\rlap{\usebox\myboxA}{\hskip 0.5\mylenA\usebox\myboxB}%
    \fi}
\makeatother

\DeclareMathOperator*{\argmin}{argmin}

\DeclareMathOperator*{\argmax}{argmax}

\makeatletter
\def\thanks#1{\protected@xdef\@thanks{\@thanks
        \protect\footnotetext{#1}}}
\makeatother

\newtheorem{theorem}{Theorem}
\newtheorem{Assumption}{Assumption}

\newtheorem{lemma}{Lemma}

\theoremstyle{definition}

\newtheorem{example}{Example}

\newtheorem{remark}{Remark}
\newcommand{\FullTitle}{Principled Penalty-based Methods for Bilevel Reinforcement Learning and RLHF}
\newcommand{\Mt}{\mathcal{M}_{\tau}}
\clearpage{}%

\title{Principled Penalty-based Methods for\\ Bilevel Reinforcement Learning and RLHF}

\begin{document}
\author{\name{Han Shen} \email{shenh5@rpi.edu}\\
  \addr{Rensselaer Polytechnic Institute}\\
  \name{Zhuoran Yang} \email{zhuoran.yang@yale.edu}\\
  \addr{Yale University}\\
  \name{Tianyi Chen} \email{chentianyi19@gmail.com}\\
  \addr{Rensselaer Polytechnic Institute}
   \thanks{
 The work of H. Shen and T. Chen was  supported by National
Science Foundation MoDL-SCALE project 2134168, the RPI-IBM Artificial Intelligence Research Collaboration (AIRC) and the Cisco Research Award.}
}

\maketitle

\begin{abstract}
Bilevel optimization has been recently applied to many machine learning tasks. However, 
    their applications have been restricted to the supervised learning setting, where static objective functions with benign structures are considered. But bilevel problems such as incentive design, inverse reinforcement learning (RL), and RL from human feedback (RLHF) are often modeled as dynamic objective functions that go beyond the simple static objective structures, which pose significant challenges of using existing bilevel solutions. 
    To tackle this new class of bilevel problems, we introduce the first principled algorithmic framework for solving bilevel RL problems through the lens of penalty formulation. We provide theoretical studies of the problem landscape and its penalty-based (policy) gradient algorithms. We demonstrate the effectiveness of our algorithms via simulations in the Stackelberg Markov game, RL from human feedback and incentive design.
\end{abstract}

\section{Introduction}

Bilevel optimization has emerged as an effective framework in machine learning for modeling decision-making problems involving incentives and misaligned objectives.
In a nutshell, bilevel optimization involves two coupled optimization problems in the upper and lower levels respectively, where they have different decision variables, denoted by $x$ and $y$ respectively. 
The lower-level problem serves as a constraint for the upper-level problem, e.g., in the upper level, we minimize a function $f(x,y)$ with the constraint that $y$ is a solution to the lower-level problem determined by $x$, i.e.,
$y \in  \Yc^*(x)$.
Here $\Yc^*(x)$ is the set of solutions to the lower-level problem determined by $x$.


Bilevel optimization enjoys a wide range of applications in machine learning, including hyper-parameter optimization \citep{maclaurin2015gradient,franceschi2018bilevel},   meta-learning \citep{finn2017maml,rajeswaran2019meta}, continue learning \citep{borsos2020coresets}, and adversarial learning \citep{jiang2021learning}.
Existing applications mostly concentrate on supervised learning setting, thus research on bilevel optimization has been predominantly confined to the static and smooth optimization setting \citep{franceschi2017forward,ghadimi2018approximation,zhang2023introduction}, where in both the upper and lower-level problems, the decision variables are typically unconstrained and the objective functions are (strongly-)convex functions. 
However, this setting is insufficient to model more complex game-theoretic behaviors with  sequential decision-making.  

Reinforcement learning (RL) \citep{SuttonRL}  is a principled framework for sequential decision-making problems and has achieved tremendous empirical success in recent years \citep{silver2017mastering, ouyang2022training}. 
In this work, we study the bilevel optimization problem in the context of RL, where the lower-level problem is an RL problem and the upper-level problem can be either smooth optimization or RL. 
Specifically, in the lower-level problem, the follower solves a Markov decision process (MDP) determined by the leader's decision variable $x$, and returns a  optimal policy of this MDP to the leader, known as the best response policy. 
The leader aims to maximize its own objective function, subject to the constraint that the follower always adopts the best response policy.  
This formulation of bilevel RL encompasses a range of applications such as Stackelberg Markov games \citep{stackelberg}, reward learning \citep{hu2020learning}, and RL from human feedback (RLHF) \citep{christiano2017deep}. 
As an example, in the RL from human feedback problem, the leader designs a reward $r_x$ for the follower's MDP,  with the goal that the resulting optimal policy yields the desired behavior of the leader.

Despite its various applications, the bilevel RL problem is difficult to solve.
Broadly speaking, the main technical challenge of bilevel optimization lies in handling the constraint, i.e., the  lower-level problem. 
The lower-level problem of bilevel  RL extends from smooth optimization to policy optimization in RL, and thus 
faces significant technical challenges.
Such an extension loses a few benign structures of optimization, such as strong convexity and uniform
Polyak-Łojasiewicz condition, which are critical for existing  bilevel optimization algorithms \citep{ghadimi2018approximation,ji2022will,shen2023penalty}. 

Specifically, there are \emph{two mainstream approaches} for bilevel optimization: (a) implicit gradient or iterative differentiation methods; and, (b) penalty-based methods.
The first approach aims to directly optimize the leader's objective under the lower-level constraints.
From the perspective of the leader's optimization problem, when finding a descent direction, the leader needs to quantify how the change of the leader's decision variable $x$ affects the follower's best response policy $\Yc^*(x)$. 
In (a), it is typically assumed that the lower-level objective function is strongly convex \citep{ji2021bilevel,chen2021tighter}, and thus the best response  is unique and the gradient of $\Yc^*(x)$ with respect  to $x$ can be computed using the implicit function theorem. 
Thus, implicit gradient method is essentially a gradient method  for the leader's objective, as a function of $x$, and the key is to differentiate the best response  $\Yc^*(x)$ in terms of $x$.  
However, in our bilevel RL case, the lower-level objective function is the discounted return in MDP, which is known to be non-convex \citep{agarwal2019optimality}. Thus, the implicit gradient are not well-defined. 
In (b), the bilevel problem is reformulated as a single-level problem by adding a penalty term to the leader's objective function. 
The penalty function penalizes the violation of the lower-level constraint. 
Thus, in the the  reformulated problem, we optimize the penalized objective with respect to both the leader and the follower's decision variables $x, y$ simultaneously. 
The penalty reformulation approach has been studied in \citep{jane2012exact,shen2023penalty,ye2022bome,kwon2023penalty} under the assumption that the lower-level objective function satisfies certain error bound conditions (e.g., uniform Polyak-Łojasiewicz inequalities).
Unfortunately, when it comes to bilevel RL, the lower-level discounted return objective does not satisfy these uniform error bound inequalities.
To develop the penalty approach for bilevel RL problems, it is unclear (i) what is an appropriate penalty function; (ii) 
how is the solution to the reformulated problem related to the original bilevel problem; and, (iii) how to solve the reformulated problem.
Therefore, directly extending applying  bilevel optimization methods to bilevel RL is not straightforward, and new theories and algorithms tailored to the  RL lower-level problem are needed, which are the subject of the paper.

We tackle this problem and provide an affirmative answer to   the following question: 
\begin{center}
    \large\textsf{Can we design a provably convergent first-order optimization algorithm for bilevel RL?}
\end{center}
To this end, 
we propose a novel algorithm that extends the idea of penalty-based bilevel optimization algorithm \citep{shen2023penalty} to tackle the specific challenges of bilevel RL. 
Our approach includes the design of two tailored penalty functions: \emph{value penalty and Bellman penalty},
which are crafted to capture the optimality condition of the lower-level RL problem.
The former is based on the optimal value function and the latter is based on the Bellman error. 
In addition, leveraging the geometry of the policy optimization problem, we
prove that an approximate solution to our reformulated problem is also an effective solution to the original bilevel problem. 
Furthermore, we establish the differentiability of the reformulated problem and  we propose a first-order policy-gradient-based algorithm that provably converges.
To our knowledge, we establish the first provably convergent first-order algorithm for bilevel RL.
 
Further enriching our research, we explore the extension of this bilevel RL framework to scenarios involving two RL agents in the lower-level problems with the goal of solving a zero-sum Markov game. Here, we introduce a value-based penalty function derived from the Nikaido-Isoda function for two-player games. 
The resulting algorithm is the first provably convergent algorithm for bilevel RL with a game constraint.
We believe our penalty reformulation approach provides a promising avenue for future research on bilevel RL with more complicated lower-level problems.

\subsection{Our contributions}
Existing bilevel optimization methods are not directly applicable to the bilevel RL problems due to the  fact that the lower-level objective function  does not entail the benign structures in supervised bilevel optimization.
The implicit and iterative gradient methods \citep{pedregosa2016hyper,franceschi2017forward} require a strongly-convex lower-level objective, which is violated in bilevel RL due to the ubiquitous non-convexity of the discounted-return objective. On the other hand, the penalty-based methods \citep{shen2023penalty,ye2022bome,kwon2023penalty} only require some weaker error bound conditions (e.g., uniform Polyak-Łojasiewicz inequalities).
Unfortunately, the lower-level discounted return objective does not satisfy these uniform error bound inequalities to our best knowledge. Though non-uniform PL inequalities have been established (e.g., in \citep{mei2020softmax}), it is not clear whether uniformity holds for bilevel algorithms. Therefore, the penalty reformulation of bilevel RL problems require further studies.

In this work, we develop a fully first-order algorithm to solve the bilevel RL problems. In developing the algorithm, we first consider how to reformulate the bilevel RL problem as a single-level RL problem with penalty functions. 
We will provide two penalty functions and show that solving the reformulated single-level problem locally/globally recovers the local/global solution of the original bilevel RL problem. Building on the reformulation, we propose a first order gradient-based algorithm that provably converges.
Furthermore, we extend the results to the two player zero-sum lower-level problem. We show a novel penalty reformulation using the Nikaido-Isoda function and propose a provably convergent algorithm. See Table \ref{tab:result summary} for a summary of convergence results.
Lastly, we conduct experiments on various applications covered by our general framework, including the Stackelberg Markov game, reinforcement learning from human feedback and incentive design.

\begin{table*}[h]
\centering
    \begin{tabular}{>{\columncolor{gray!10}}c|c|  c}
    \hline
    \hline
    &Section \ref{sec:penalty reform}  \& Section \ref{sec:algorithm}  & Section \ref{sec:bilevel rl zero sum} \\
    \hline
    Lower-level problem & single-agent RL & two-player zero-sum \\
    \hline
    Upper-level problem & \multicolumn{2}{c}{general objective} \\
    \hline
    Penalty functions& Value or Bellman penalty & Nikaido-Isoda (NI) \\
    \hline
    Penalty constant $\lambda$ & $\Omega(\delta_{value}^{-1})^\dagger$ or $\Omega(\delta_{bellman}^{-0.5})^\dagger$ & $\Omega(\delta_{NI}^{-1})^\dagger$\\
    \hline
    Inner-loop oracle algorithm & \multicolumn{2}{c}{Policy mirror descent (PMD)} \\
    \hline
    Iteration complexity &\multicolumn{2}{c}{$\Oc(\lambda\epsilon^{-1}\log(\lambda^2/\epsilon))^\ddagger$}\\
    \hline
    \hline
    \end{tabular}
    \caption{Summary of main convergence theorems. $\dagger$: the lower-bound of the penalty constant $\lambda$ to guarantee that certain lower-level optimality gap (value function gap, Bellman gap and NI function value) is smaller than accuracy $\delta_{value}$, $\delta_{bellman}$ and $\delta_{NI}$ respectively. These optimality gaps will be introduced in their respective sections; $\ddagger$: Here $\epsilon$ is the accuracy of our algorithm.} 
    \label{tab:result summary}
\end{table*}

\subsection{Related works}
\textbf{Bilevel optimization.} 
The bilevel optimization problem can be dated back to \citep{stackelberg}. 
The gradient-based bilevel optimization methods have gained growing popularity in the machine learning area; see, e.g., \citep{sabach2017jopt,franceschi2018bilevel,liu2020icml}. 
A prominent branch of gradient-based bilevel optimization is based on the implicit gradient (IG) theorem. The IG based methods have been widely studied under a strongly-convex lower-level function, see, e.g., \citep{pedregosa2016hyper,ghadimi2018approximation,hong2020ac,ji2020provably,chen2021tighter,khanduri2021near,shen2022single, Li2022fully,xiao2022alternating,giovannelli2022inexact,chen2023optimal,yang2023achieving}.  The iterative differentiation (ITD) methods, which can be viewed as an iterative relaxation of the IG methods, have been studied in, e.g., \citep{maclaurin2015gradient,franceschi2017forward,nichol2018firstorder,shaban2019truncated,liu2021towards,liu2022general,bolte2022automatic,grazzi2020iteration,ji2022will,shen2022single}. However, in our case the lower-level objective is the discounted return which is known to be non-convex \citep{agarwal2019optimality}. Thus it is difficult to apply the fore-mentioned methods here.

The penalty relaxation of the bilevel optimization problem, early studies of which can be dated back to \citep{clarke1983optimization,luo1996mathematical}, have gained interests from researchers recently (see, e.g., \citep{shen2023penalty,ye2022bome,lu2023first,kwon2023penalty,xiao2023generalized,lu2024slm}). Theoretical results for this branch of work are established under certain lower-level error bound conditions (e.g., uniform Polyak-Łojasiewicz inequalities) weaker than strong convexity. 
While in our case, the lower-level discounted return objective does not satisfy those uniform error bounds. 
Therefore, the established penalty reformulations may not be directly applied here. See Table \ref{tab:comparison with general bilevel opt} for more detailed comparison between this work and the general penalty-based bilevel optimization.

\newcolumntype{C}{>{\Centering\arraybackslash}X} 
\begin{table*}[h]
\small
\centering
    \begin{tabularx}{\textwidth}{>{\columncolor{gray!10}}C|C|  C}
    \hline
    \hline
     &Supervised penalty-based bilevel OPT   & This work on penalty-based bilevel RL  \\
    \hline
    Problem application & hyperparameter OPT, adversarial training, continue learning, etc. & Stackelberg Markov game, RL from preference, incentive design, etc \\
    \hline
    Penalty reformulation & Value penalty with assumed property & Value/Bellman/NI penalty with proven property \\
    \hline
    Algorithm & Gradient directly accessible & Need to derive close form gradient and estimate it \\
    \hline
    Iteration complexity & $\tilde{\Oc}(\lambda\epsilon^{-1})$ with inner-loop GD  & $\tilde{\Oc}(\lambda\epsilon^{-1})$ with inner-loop PMD\\
    \hline
    \hline
    \end{tabularx}
    \caption{A holistic comparison between the bilevel RL in this work and the general penalty-based bilevel optimization (OPT) (e.g., \citep{shen2023penalty,kwon2023penalty}), where "GD" stands for the gradient descent and "PMD" stands for the policy mirror descent.} 
    \label{tab:comparison with general bilevel opt}
\end{table*}

\vspace{+0.05cm}
\noindent\textbf{Policy-based RL.} The policy-based RL algorithms are generally based on the policy gradient theorem \citep{SuttonPG}. There has been a large body of literature studying the policy-based algorithms, including the Monte-Carlo sampling based policy gradient methods \citep{SuttonPG,baxter2001infinite}, the advantage actor-critic algorithm \cite{borkar1997actor,A3C}, proximal policy optimization \cite{schulman2017proximal}, and more generally the policy mirror descent methods \citep{lan2023policy,zhan2023policy}. The landscape of the RL objective and the (global) convergence of the policy gradient based algorithms have been extensively studied in, to list a few, \citep{agarwal2019optimality,zhang2019global,qiu2019finite,bhandari2020global,mei2020softmax,wu2020finite,junzi2021sample,cen2022fast,shen2023towards,ding2024last}.

\noindent\textbf{Applications of bilevel RL.} The bilevel RL formulation considered in this work covers several applications including reward shaping \citep{hu2020learning,zou2019reward}, reinforcement learning from preference \citep{christiano2017deep, xu2020preference,pacchiano2021dueling}, Stackelberg Markov game \citep{liu2021sharp,song2023can}, AI-economics with two-level deep RL \citep{zheng2022ai}, social environment design \citep{zhang2024social}, incentive design \citep{chen2016caching}, etc. A concurrent work \citep{chakraborty2023aligning} studies the policy alignment problem, and introduces a corrected reward learning objective for RLHF that leads to strong performance gain. While the PARL algorithm in \citep{chakraborty2023aligning} is based on the implicit gradient bilevel optimization method that requires the strong-convexity of the lower-level objective. On the other hand, PARL uses second-order derivatives of the RL objective, while our proposed algorithm is fully first order.

\section{Problem Formulations}\label{sec:formulation}

In this section, we will first introduce the generic bilevel RL formulation. Then we will show several specific applications of the generic bilevel RL problem.

\subsection{Bilevel reinforcement learning formulation}
Reinforcement learning studies the problem where an agent aims to find a policy that maximizes its accumulated reward under the environment's dynamic. In such problem, the reward function and the dynamic are fixed given the agent's policy. While in the problem that we are about to study, the reward or the dynamic oftentimes depend on another decision variable, e.g., the reward is parameterized by a neural network in RLHF; or in Stackelberg game, both the reward and the dynamic are affected by the leader's policy. 

Tailoring to this, we first define a so-called parameterized MDP. Given the parameter $x\in \reals^{d_x}$, define a parameterized MDP as $\Mc_\tau(x)\coloneqq \{\Sc, \Ac, r_x, \Pc_x, \tau h\}$ where $\Sc$ is a finite state space; $\Ac$ is a finite action space; $r_x(s,a)$ is the parameterized reward given state-action pair $(s,a)\in\Sc\times\Ac$; $\Pc_x$ is a parameterized transition distribution that specifies $\Pc_x(s'|s,a)$--the probability of transiting to $s'$ given $(s,a)$; a policy $\pi$ specifies $\pi(a|s)$ which is the probability of taking action $a$ given state $s$; and $\tau h$ is the regularization: $\tau\geq 0$ and $h=(h_s)_{s\in\Sc}$ where each $h_s:\Delta(\Ac)\mapsto \reals_{+}$ is a strongly-convex regularization function given $s$. When $\tau=0$, $\Mt(x)$ is an unregularized MDP. 

Given a policy $\pi$, the value function of $\Mt(x)$ under $\pi$ is defined as
\begin{align}\label{eq:vpi}
    V_{\Mt(x)}^\pi(s) \coloneqq \E\Big[\sum_{t=0}^\infty \gamma^t \big(r_x(s_t,a_t)-\tau h_{s_t}(\pi(s_t))\big) \big| s_0=s,\pi\Big]
\end{align}
where $\gamma\in[0,1)$, $\pi(s)\coloneqq\pi(\cdot|s)\in\Delta(\Ac)$ and the expectation is taken over the trajectory $(s_0,a_0\sim \pi(s_0),s_1\sim\Pc_x(\cdot|s_0,a_0),\dots)$. 
Given a state distribution $\rho$, we write $V_{\Mt(x)}^\pi(\rho)=\E_{s\sim\rho}[V_{\Mt(x)}^\pi(s)]$. 
We also define the $Q$ function as
\begin{align}
    Q_{\Mt(x)}^\pi(s,a) \coloneqq r_x(s,a)+\gamma \E_{s'\sim\Pc_x(\cdot|s,a)}\big[V_{\Mt(x)}^\pi(s')\big].
\end{align}
and $P_x^\pi(s_t=s|s_0)$ as the probability of reaching state $s$ at time  $t$ given initial state $s_0$ under a transition distribution $\Pc_x$ and a policy $\pi$. The probability $P_x^\pi(s_t=s|s_0,a_0)$ can be defined similarly.

Suppose the policy $\pi$ is parameterized by $y\in\Yc\subseteq\reals^{d_y}$. We define the policy class as $\Pi\coloneqq \{\pi_y:y\in\Yc\}\subseteq \Delta(\Ac)^{|\Sc|}$.
For $\Mt(x)$, its optimal policy is denoted as $\pi_y^*(x)\in\Pi$ satisfying $V_{\Mt(x)}^{\pi_y^*(x)}(s) \geq V_{\Mt(x)}^{\pi}(s)$ for any $\pi\in\Pi$ and $s$. 
With a function $f:\reals^{d_x}\times\reals^{d_y}\mapsto \reals$, we are interested in the following bilevel RL problem

{\centering
\begin{tcolorbox}[width=0.82\textwidth, right=-20pt]
\leqnomode
\vspace{-0.3cm}
\begin{align}
    ~~~~\mathcal{BM}: \min_{x,y} f(x,y),~~{\rm s.t.}~x\in\Xc,~ y\in \Yc^*(x)\coloneqq \argmin_{y\in\Yc}-V_{\Mt(x)}^{\pi_y}(\rho)
\end{align}
\vspace*{-0.4cm}
\reqnomode
\end{tcolorbox}}
\noindent where $\Xc \subseteq\reals^{d_x}$ and $\Yc\subseteq \reals^{d_y}$ are compact convex sets; and $\rho$ is a given state distribution with $\rho(s)>0$ on $\Sc$. The name `bilevel' refers to the nested structure in the optimization problem: in the upper-level, a function $f(x,y)$ is minimized subject to the lower-level optimality constraint that $\pi_y$ is the optimal policy for $\Mc_\tau(x)$. 

\subsection{Applications of bilevel reinforcement learning}\label{sec:application}
Next we show several example applications that can be modeled by a bilevel RL problem.

{\noindent \textbf{Stackelberg Markov game.}} Consider a Markov game where at each time step, a leader and a follower observe the state and make actions simultaneously. Then according to the current state and actions, the leader and follower receive rewards and the game transits to the next state. Such a MDP can be defined as $\Mc_\tau^g\coloneqq \{\Sc, \Ac_l, \Ac_f, r_l, r_f,\Pc,\tau h_l,\tau h_f \}$ where $\Sc$ is the state space; $\Ac_l/\Ac_f$ is the leader's/follower's action space; $r_l(s,a_l,a_f)$ and $r_f(s,a_l,a_f)$ are respectively the leader's and the follower's reward given $(s,a_l,a_f)\in\Sc\times\Ac_l\times\Ac_f$; $\Pc(s'|s,a_l,a_f)$ is the probability of transiting to state $s'$ given $(s,a_l,a_f)$; the leader's/follower's policy $\pi_x$/$\pi_y$ defines $\pi_x(a_l|s)$/$\pi_y(a_f|s)$--the probability of choosing action $a_l$/$a_f$ given state $s$; and $\tau h_l,\tau h_f$ are the regularization functions respectively for $\pi_x$ and $\pi_y$. 

Define the leader's/follower's value function as
\begin{align}
    V_\star^{\pi_x,\pi_y}(s) \coloneqq \E\Big[\sum_{t=0}^\infty \gamma^t \big(r_\star(s_t,a_{l,t},a_{f,t})-\tau h_{\star,s_t}(\pi_\star(s_t))\big) \big| s_0=s,\pi_x,\pi_y\Big],~~\star=l~~{\rm or }~~f
\end{align}
where $\gamma\in[0,1)$, $\pi_\star(s)\coloneqq\pi_\star(\cdot|s)\in\Delta(\Ac_\star)$ and the expectation is taken over the trajectory $(s_0,~a_{l,0}\sim \pi_x(s_0),~a_{f,0}\sim \pi_y(s_0),~s_{l,1}\sim\Pc(s_0,a_{l,0},a_{f,0}),\dots)$. The Q function can be defined as
$$Q_\star^{\pi_x,\pi_y}(s,a_l,a_f)\coloneqq r_\star(s,a_{l},a_{f})+\gamma\E_{s'\sim\Pc(s,a_l,a_f)}\big[V_\star^{\pi_x,\pi_y}(s')\big].$$

The follower's objective is to find a best-response policy to the leader's policy while the leader aims to find a best-response to the follower's best-response. The problem can be formulated as
\begin{align}\label{eq:SG original}
    \min_{x,y} -V_l^{\pi_x,\pi_y}(\rho),~~{\rm s.t.}~x\in\Xc,~ y\in \argmin_{y\in\Yc} -V_f^{\pi_x,\pi_y}(\rho).
\end{align}
With the proof deferred to Appendix \ref{sec:stackelberg proof}, this problem can be viewed as a bilevel RL problem with $f(x,y)=-V_l^{\pi_x,\pi_y}(\rho)$ and a $\Mt(x)$ in which $r_x(s,a_f)=\E_{a_l\sim\pi_x(s)}[r_l(s,a_l,a_f)]$ and $\Pc_x(\cdot|s,a_f)=\E_{a_l\sim\pi_x(s)}[\Pc(\cdot|s,a_l,a_f)]$.

\vspace{5pt}
{\noindent \textbf{Reinforcement learning from human feedback (RLHF).}}
In the RLHF setting, the agent learns a task without knowing the true reward function. Instead, humans evaluate pairs of state-action segments, and for each pair they label the segment they prefer. The agent's goal is to learn the task well with limited amount of labeled pairs.

The original framework of deep RL from human feedback in \citep{christiano2017deep} (we call it DRLHF) consists of two possibly asynchronous learning process: reward learning from labeled pairs and RL from learnt rewards. In short, we maintain a buffer of labeled segment pairs $\{(d_0,l_0,d_1,l_1)_i\}_{i}$ where each segment $d=(s_{t},a_{t},\dots,s_{t+T},a_{t+T})$ is collected with the agent's policy $\pi_y$ and $l_0,l_1$ is the label (e.g., $l_1=1,l_0=0$ indicates segment $d_1$ is preferred over $d_0$). DRLHF simultaneously learns a reward predictor $r_x$ with the data and trains an RL agent using the learnt reward. 
This process has a hierarchy structure and can be reformulated as a bilevel RL problem:
\begin{align}\label{eq:rlhf bilevel formulation}
    \min_{x,y} -\E_{d_0,d_1 \sim \pi_y}\big[l_0 \log P(d_0 \succ d_1 | r_x) + l_1 \log P(d_1 \succ d_0 | r_x)\big],~~{\rm s.t.}~~y\in\argmin_{y}-V_{\Mt(x)}^{\pi_y}(\rho).
\end{align}
where $P(d_0 \succ d_1 | r_x)$ is the probability of preferring $d_0$ over $d_1$ under reward $r_x$, given by the Bradley-Terry model:
\begin{align}
    P(d_0 \succ d_1 | r_x) = \frac{\exp (\sum_{s_t,a_t\in d_0}r_x(s_t,a_t))}{\exp (\sum_{s_t,a_t\in d_0}r_x(s_t,a_t))+\exp (\sum_{s_t,a_t\in d_1}r_x(s_t,a_t))}.
\end{align}
\begin{remark}[Connection with DPO \citep{rafailov2023direct}]
    The formulation in \eqref{eq:rlhf bilevel formulation} becomes similar to DPO \citep{rafailov2023direct} in a special case. Specifically when $\gamma=0, T=0$, $\pi_y$ is tabular and $h_s(\pi_y(s))=D_{KL}(\pi_y(s)||\pi_{ref}(s))$ where $\pi_{ref}$ is a given reference model, the lower level problem in \eqref{eq:rlhf bilevel formulation} is solved if and only if the equation $r_x(s,a) = \tau \log \frac{\pi_y(a|s)}{\pi_{ref}(a|s)}+\tau \log Z_{r_x}(s)$ holds, where $Z_{r_x}(s)$ is some partition function (see, e.g., \citep[eq. 5]{rafailov2023direct}). Plugging this equation back in the upper-level loss results in the DPO objective. The only difference is that the upper-level loss is on policy since the samples follow $\pi_y$, while the DPO loss depends on an off-policy dataset.
\end{remark}

\textbf{Reward shaping.}
In the RL tasks where the reward is difficult to learn from (e.g., the reward signal is sparse where most states give zero reward), we can reshape the reward to enable efficient policy learning while staying true to the original task.
Given a task specified by $\Mt=\{\Sc,\Ac,r,\Pc,\tau h\}$, the reward shaping problem \citep{hu2020learning} seeks to find a reshaped reward $r_x$ parameterized by $x\in\Xc$ such that the new MDP with $r_x$ enables more efficient policy learning for the original task. We can define the new MDP as $\Mt(x)=\{\Sc,\Ac,r_x,\Pc,\tau h\}$ and formulate the problem as:
\begin{align}
    \!\!\min_{x,y} -&V_{\Mc_\tau}^{\pi_y}(\rho),~~{\rm s.t.}~~x\!\in\!\Xc,~y\!\in\!\argmin_{y\in\Yc}-V_{\Mt(x)}^{\pi_y}(\rho)
\end{align}
which is a special case of bilevel RL.


\section{Penalty Reformulation of Bilevel RL}\label{sec:penalty reform}
A natural way to solve the bilevel RL problem $\mathcal{BM}$ is through reduction to a single-level problem, that is, to find a single-level problem that shares its local/global solutions with the original problem.
Then by solving the single-level problem, we can recover the original solutions.
In this section, we will perform single-level reformulation of $\mathcal{BM}$ through penalizing the upper-level objective with carefully chosen functions. 

Specifically, we aim to find penalty functions $p(x,y)$ such that the solutions of the following problem recover the solutions of $\mathcal{BM}$:
\begin{align}\label{eq:bmlambdap}
    \mathcal{BM}_{\lambda p}:~\min_{x,y} F_\lambda(x,y)\coloneqq f(x,y)+\lambda p(x,y),~~{\rm s.t.}~~x\in\Xc,~y\in\Yc
\end{align}
where $\lambda$ is the penalty constant. 

\subsection{Value penalty and its landscape property}
In $\mathcal{BM}$, the lower-level problem of finding the optimal policy $\pi_y$ can be rewritten as its optimality condition: $-V_{\Mt(x)}^{\pi_y}(\rho)+\max_{y\in\Yc} V_{\Mt(x)}^{\pi_y}(\rho)=0$.
Therefore, $\mathcal{BM}$ can be rewritten as
\begin{align}
    \min_{x,y} f(x,y),~~{\rm s.t.}~~x\in\Xc,~y\in\Yc,~-V_{\Mt(x)}^{\pi_y}(\rho)+\max_{y\in\Yc} V_{\Mt(x)}^{\pi_y}(\rho)=0.\nonumber
\end{align}
A natural penalty function that we call \emph{value penalty} then measures the lower-level optimality gap:
\begin{align}\label{eq:value penalty}
    p(x,y)=-V_{\Mt(x)}^{\pi_y}(\rho)+\max_{y\in\Yc} V_{\Mt(x)}^{\pi_y}(\rho).
\end{align}
The value penalty specifies the following penalized problem
\begin{align}\label{eq:BMlp value p}
    \mathcal{BM}_{\lambda p}:~\min_{x,y} F_\lambda(x,y)= f(x,y)+\lambda \big(-V_{\Mt(x)}^{\pi_y}(\rho)+\max_{y\in\Yc} V_{\Mt(x)}^{\pi_y}\big),~~{\rm s.t.}~~x\in\Xc,~y\in\Yc.
\end{align}
To capture the relation between solutions of $\mathcal{BM}_{\lambda p}$ and $\mathcal{BM}$, we have the following lemma.
\begin{lemma}[Relation on solutions]\label{lem:value penalty solution}
    Consider choosing $p$ as the value penalty in \eqref{eq:value penalty}. Assume there exists constant $C$ such that $\max_{x\in\Xc,y\in\Yc}|f(x,y)|\!=\!\frac{C}{2}$. Given accuracy $\delta\!>\!0$, choose $\lambda\geq C\delta^{-1}$. If $(x_\lambda,y_\lambda)$ achieves $\epsilon$-minimum of $\mathcal{BM}_{\lambda p}$, it achieves $\epsilon$-minimum of the relaxed $\mathcal{BM}$:
    \begin{align}\label{eq:approxBM1}
         \min_{x,y} f(x,y),~~{\rm s.t.}~~x\in\Xc,~y\in\Yc,~-V_{\Mt(x)}^{\pi_y}(\rho)+\max_{y\in\Yc} V_{\Mt(x)}^{\pi_y}(\rho)\leq \epsilon_\lambda
    \end{align}
    where $\epsilon_\lambda \leq \delta+\lambda^{-1}\epsilon$.
\end{lemma}
The proof is deferred to Appendix \ref{sec:value penalty solution proof}.
Perhaps one restriction of the above lemma is that it requires the boundedness of $f$ on $\Xc\times\Yc$. This assumption is usually mild in RL problems, e.g., it is guaranteed in Stackelberg game provided the reward functions are bounded. 

Since $\mathcal{BM}_{\lambda p}$ is in general a non-convex problem, it is also of interest to connect the local solutions between $\mathcal{BM}_{\lambda p}$ and $\mathcal{BM}$. To achieve this, some structural condition is required.
Suppose we use direct policy parameterization: $y$ is a vector with its $(s,a)$ element $y_{s,a}=\pi_y(a|s)$, and thus $y=\pi_y$ directly. Then we can prove the following structural condition.
\begin{lemma}[Gradient dominance]\label{lemma:gradient dominance direct param}
    Given convex policy class $\Pi$ and any $\tau\geq 0$, $V_{\Mt(x)}^\pi(\rho)$ is gradient dominated in $\pi$:
    $$\max_{\pi'\in\Pi}\ip{\nabla_\pi V_{\Mt(x)}^{\pi}(\rho)}{\pi'-\pi} \geq \frac{1}{(1-\gamma)\min_s \rho(s)} \Big(\max_{\tilde{\pi}\in\Pi}V_{\Mt(x)}^{\tilde{\pi}}(\rho)-V_{\Mt(x)}^\pi(\rho)\Big),~\forall \pi\in\Pi.$$
\end{lemma}
See Appendix \ref{sec:gradient dominance direct param proof} for a proof.
A similar gradient dominance property was first proven in \cite[Lemma 4.1]{agarwal2019optimality} for the unregularized MDPs. The above lemma is a generalization of the result in \citep{agarwal2019optimality} to regularized case. Under such structure of the lower-level problem, we arrive at the following lemma capturing the relation on local solutions.

\begin{lemma}[Relation on local solutions]\label{lemma:local solution relation value p}
 Consider using direct policy parameterization and choosing $p$ as the value penalty in \eqref{eq:value penalty}. Assume $f(x,\cdot)$ is $L$-Lipschitz-continuous on $\Yc$. Given accuracy $\delta\!>\!0$, choose $\lambda\geq L C_u\delta^{-1}$ where $C_u$ is a constant specified in the proof. If $(x_\lambda,y_\lambda)$ is a local solution of $\mathcal{BM}_{\lambda p}$, it is a local solution of the relaxed $\mathcal{BM}$ in \eqref{eq:approxBM1} with an $\epsilon_\lambda \leq \delta$.
\end{lemma}
The proof can be found in Appendix \ref{sec:local solution relation value p proof}. Lemmas \ref{lem:value penalty solution} and \ref{lemma:local solution relation value p} suggest we can recover the local/global solutions of the bilevel RL problem $\mathcal{BM}$ by locally/globally solving its penalty reformulation $\mathcal{BM}_{\lambda p}$ with the value penalty.

\subsection{Bellman penalty and its landscape property}
Next we introduce the Bellman penalty that can be used as an alternative.
To introduce this penalty function, we consider a tabular policy (direct parameterization) $\pi_y$, i.e. $\pi_y(\cdot|s)=y_s$ for all $s$ and $y=(y_s)_{s\in\Sc}\in\Yc= \Pi$. Then we can define the \emph{Bellman penalty} as
\begin{align}\label{eq:bellman penalty}
    p(x,y)=g(x,y)-v(x)~~~~{\rm where}~~v(x)\coloneqq \min_{y\in\Yc}g(x,y).
\end{align}
Here $g(x,y)$ is defined as
\begin{align}
    g(x,y)\coloneqq \E_{s\sim\rho}[\ip{y_s}{q_s(x)}+\tau h_s(y_s)],~~
\end{align}
where $q_s(x)\in\reals^{|\Ac|}$ is the vector of optimal Q functions, which is defined as
\begin{align}
    q_s(x)=(q_{s,a}(x))_{a\in\Ac}~~{\rm where}~~q_{s,a}(x)\coloneqq -\max_{\pi\in\Pi} Q_{\Mt(x)}^\pi (s,a).
\end{align}
It is immediate that $p(x,\cdot)$ is $\tau$-strongly-convex uniformly for any $x\in\Xc$ by the 1-strong-convexity of $h_s$, and $p(x,y)\geq 0$ by definition. 
Moreover, we can show that $p(x,y)=g(x,y)-v(x)$ is a suitable optimality metric of the lower-level RL problem in $\mathcal{BM}$.
Specifically, we prove that the lower-level RL problem is solved whenever $g(x,y)-v(x)$ is minimized in the following lemma.
\begin{lemma}\label{lemma:reformulation}
Assume $\tau>0$, then we have the following holds.
\begin{itemize}
    \item Given any $x\in\Xc$, MDP $\Mt(x)$ has a unique optimal policy $\pi_y^*(x)$. And we have $\arg\min_{y\in\Yc}g(x,y)=\Yc^*(x)=\{\pi_y^*(x)\}$. Therefore, $\mathcal{BM}$ can be rewritten as the following problem with $\epsilon=0$:
    \begin{align}\label{eq:reform0}
    \mathcal{BM}_\epsilon:~~\min_{x,y} f(x,y),~~{\rm s.t.}~&x\in\Xc,~y\in\Yc,\nonumber\\
    &g(x,y)-v(x)\leq \epsilon.
\end{align}
\item Assume $f(x,\cdot)$ is $L$-Lipschitz-continuous on $\Yc$. More generally for $\epsilon \geq 0$, $\mathcal{BM}_\epsilon$ is an $\epsilon$-approximate problem of $\mathcal{BM}$ in a sense that: given any $x\in\Xc$, any feasible policy $y_\epsilon$ of $\mathcal{BM}_\epsilon$ is $\epsilon$-feasible for $\mathcal{BM}$:
\begin{align}
    \|y_\epsilon-\pi_y^*(x)\|^2 \leq \tau^{-1}\epsilon.\nonumber
\end{align}
Moreover, let $f^*$,$f_\epsilon^*$ respectively be the optimal objective value of $\mathcal{BM}$ and $\mathcal{BM}_\epsilon$, then we have $|f^*-f_\epsilon^*|\leq L\sqrt{\tau^{-1}\epsilon}.$
\end{itemize}
\end{lemma}
The proof is deferred to Appendix \ref{sec:reformulation proof}.
Based on Lemma \ref{lemma:reformulation}, $g(x,y)-v(x)$ is a suitable optimality metric for the lower-level problem. It is then natural to consider whether we can use it as a penalty function for the lower-level sub-optimality.
The Bellman penalty specifies the following penalized problem:
\begin{align}\label{eq:BMlp bellman p}
    \mathcal{BM}_{\lambda p}:~\min_{x,y} F_\lambda(x,y)= f(x,y)+\lambda \big(g(x,y)-v(x)\big),~~{\rm s.t.}~&x\in\Xc,~y\in\Yc.
\end{align}
We have the following result that captures the relation between the solution of $\mathcal{BM}_\epsilon$ and $\mathcal{BM}_{\lambda p}$, which proves the Bellman penalty is indeed a suitable penalty function.
\begin{lemma}[Relation on solutions]\label{lemma:generic solution}
Consider choosing $p$ as the Bellman penalty in \eqref{eq:bellman penalty}.
    Assume $f(x,\cdot)$ is $L$-Lipschitz-continuous on $\Yc$. Given some accuracy $\delta>0$, choose $\lambda \geq L\sqrt{\tau^{-1}\delta^{-1}}$. If $(x_\lambda, y_\lambda)$ is a local/global solution of $\mathcal{BM}_{\lambda p}$, then it is a local/global solution of $\mathcal{BM}_{\epsilon_\lambda}$ with $\epsilon_\lambda\leq \delta.$
\end{lemma}
This lemma follows directly from the $\tau$-strong-convexity of $g(x,\cdot)$ and Proposition 3 in \citep{shen2023penalty}.

\section{A Penalty-based Bilevel RL Algorithm}\label{sec:algorithm}
In the previous sections, we have introduced two penalty functions $p(x,y)$ such that the original problem $\mathcal{BM}$ can be approximately solved via solving $\mathcal{BM}_{\lambda p}$. However, it is still unclear how $\mathcal{BM}_{\lambda p}$ can be solved. One challenge is the differentiability of the penalty function $p(x,y)$ in \eqref{eq:bmlambdap}.
In this section, we will first study when $F_\lambda(x,y)$ admits gradients in the generic case, and we will show the specific gradient forms in each application. Based on these results, we propose a penalty-based algorithm and further establish its convergence.

\subsection{Differentiability of the value penalty}
We first consider the value penalty 
$$p(x,y)=-V_{\Mt(x)}^{\pi_y}(\rho)+\max_{y\in\Yc} V_{\Mt(x)}^{\pi_y}(\rho).$$ 
For the differentiability in $y$, it follows $\nabla_y p(x,y)=-\nabla_y V_{\Mt(x)}^{\pi_y}(\rho)$ which can be conveniently evaluated with the policy gradient theorem. The issue lies in the differentiability of $p(x,y)$ with respect to $x$, where $p(x,y)$ may not be differentiable in $x$ due to the optimality function $\max_{y\in\Yc} V_{\Mt(x)}^{\pi_y}(\rho)$.
Fortunately, we will show that in the setting of RL, $p(\cdot,y)$ admits closed-form gradient under relatively mild assumptions below.
\begin{Assumption}\label{asp:diff value p}
Assume
\begin{enumerate}[label=(\alph*)]
    \item $\nabla_x V_{\Mt(x)}^{\pi_y}(\rho)$ is continuous in $(x,y)$; and,
    \item given any $x\in\Xc$ and $y,y'\in\Yc^*(x)$, we have $\nabla_x V_{\Mt(x)}^{\pi_{y}}(\rho)=\nabla_x V_{\Mt(x)}^{\pi_{y'}}(\rho)$.
\end{enumerate}
\end{Assumption}
Assumption \ref{asp:diff value p} (a) is mild in the applications, and can often be guaranteed by the a continuously differentiable reward function $r_x$. A sufficient condition of Assumption \ref{asp:diff value p} (b) is the optimal policy of $\Mt(x)$ on $\Pi$ is unique, e.g., when $\pi_y=\pi_{y'}$ for $y,y'\in\Yc^*(x)$.
As indicated by Lemma \ref{lemma:reformulation}, the uniqueness is guaranteed when $\tau>0$.

\begin{lemma}[Generic gradient form]\label{lemma:derivative value p}
    Consider the value penalty $p$ in \eqref{eq:value penalty}. Suppose Assumption \ref{asp:diff value p} holds. Then $p(x,y)$ is differentiable in $x$ with the gradient
    \begin{align}
         \nabla_x p(x,y)=-\nabla_x V_{\Mt(x)}^{\pi_y}(\rho)+ \nabla_x V_{\Mt(x)}^{\pi}(\rho)|_{\pi=\pi_y^*(x)}
    \end{align}
    where recall $\pi_y^*(x)$ is an optimal policy on policy class $\Pi=\{\pi_y:y\in\Yc\}$ of $\Mt(x)$.
\end{lemma}
The proof can be found in Appendix \ref{sec:derivative value p proof}.
Next, we can apply the generic result from Lemma \ref{lemma:derivative value p} to specify the exact gradient formula in different bilevel RL applications discussed in Section \ref{sec:application}.
\begin{lemma}[Gradient form in the applications]\label{lemma:derivative value p application}
Consider the value penalty $p$ in \eqref{eq:value penalty}.
The gradient of the penalty function in specific applications are listed below.
\begin{enumerate}[label=(\alph*)]
    \item \textit{RLHF/reward shaping}: Assume $r_x$ is continuously differentiable and Assumption \ref{asp:diff value p} (b) holds. Then Lemma \ref{lemma:derivative value p} holds and we have
        \begin{align}
            &\nabla_x p(x,y)=-\E\Big[\sum_{t=0}^\infty \gamma^t \nabla r_x(s_t,a_t)\big| \rho,\pi_y\Big]+\E\Big[\sum_{t=0}^\infty \gamma^t \nabla r_x(s_t,a_t)\big| \rho,\pi_y^*(x)\Big].\nonumber
        \end{align}
        \item \textit{Stackelberg game}: Assume $\pi_x$ is continuously differentiable and  Assumption \ref{asp:diff value p} (b) holds. Then Lemma \ref{lemma:derivative value p} holds and we have
        \begin{align}
        &\nabla_x p(x,y) =
            -\E\Big[\sum_{t=0}^\infty \gamma^t \big( Q_f^{\pi_x,\pi_y}(s_t,a_{l,t},a_{f,t})-\tau h_{f,s_t}(\pi_y(s_t)) \big)\nabla\log\pi_x(a_{l,t}|s_t)\big|s_0=s,\pi_x,\pi_y\Big] \nonumber\\
    &+ \E\Big[\sum_{t=0}^\infty \gamma^t \big( Q_f^{\pi_x,\pi_y^*(x)}(s_t,a_{l,t},a_{f,t})-\tau h_{f,s_t}(\pi_y^*(x)(s_t)) \big)\nabla\log\pi_x(a_{l,t}|s_t)\big|s_0\!=\!s,\pi_x,\pi_y^*(x)\Big]\nonumber
        \end{align}
        Recall in the Stackelberg setting, $\pi_y^*(x)$ is the optimal follower policy given $\pi_x$; and the expectation is taken over the trajectory generated by $\pi_x,\pi_y (\text{or }\pi_y^*(x)),\Pc$.
\end{enumerate}
\end{lemma}
We defer the proof to Appendix \ref{sec:derivative value p application proof}.

\subsection{Differentiability of the Bellman penalty}
For the Bellman penalty in \eqref{eq:bellman penalty},
though it is straightforward to evaluate $\nabla_y p(x,y)=\nabla_y g(x,y)$, the differentiability of $p(x,y)$ in $x$ is unclear. We next identify some sufficient conditions that allow convenient evaluation of $\nabla_x p(x,y)$.
\begin{Assumption}\label{asp:diff}
Assume $\tau>0$ and the following hold:
\begin{enumerate}[label=(\alph*)]
    \item Given any $(s,a)$, $\nabla_x Q_{\Mt(x)}^\pi(s,a)$ exists and is continuous in $(x,\pi)$; and,
    \item Either the discount factor $\gamma=0$ or: Given $x\in\Xc$, for the MDP $\Mt(x)$, the Markov chain induced by any policy $\pi\in\Pi$ is irreducible\footnote{In $\Mt(x)$, the Markov chain induced by policy $\pi$ is irreducible if for any state $s$ and initial state-action pair $s_0,a_0$, there exists time step $t$ such that $P_x^\pi(s_t=s|s_0,a_0)>0$, where $P_x^\pi(s_t=s|s_0,a_0)$ is the probability of reaching $s$ at time step $t$ in MDP $\Mt(x)$ with policy $\pi$.}.
\end{enumerate}
\end{Assumption}
Assumption \ref{asp:diff} (a) is mild and can be satisfied in the applications in Section \ref{sec:application}. Assumption \ref{asp:diff} (b) is a regularity assumption on the MDP \citep{mitrophanov2005sensitivity}, and is often assumed in recent studies on policy gradient algorithms (see e.g., \citep{wu2020finite,qiu2021finite}). 

\begin{lemma}[Generic gradient form]\label{lemma:derivative}
    Consider the Bellman penalty $p$ in \eqref{eq:bellman penalty}. Assume Assumption \ref{asp:diff} holds. Then $p(x,y)$ is differentiable with the gradient $\nabla_x p(x,y)=\nabla_x g(x,y)-\nabla v(x)$ where
    \begin{align}
        &\nabla_x g(x,y)=-\E_{s\sim\rho,a\sim\pi_y(s)}\big[\nabla_x Q_{\Mt(x)}^\pi(s,a)\big]\big|_{\pi=\pi_y^*(x)}\\
        &\nabla v(x) = -\E_{s\sim\rho,a\sim\pi_y^*(x)(s)}\big[\nabla_x Q_{\Mt(x)}^\pi(s,a)\big]\big|_{\pi=\pi_y^*(x)}
    \end{align}
\end{lemma}
The proof can be found in Appendix \ref{sec:derivative proof}.
The above lemma provides the form of gradients for the $\mathcal{BM}$ problem. Next we show that Lemma \ref{lemma:derivative} holds for the example applications in Section \ref{sec:application} and then compute the closed-form of the gradients.

\begin{lemma}[Gradient form in the applications]\label{lemma:derivative application}
Consider the Bellman penalty $p(x,y)$ in \eqref{eq:bellman penalty}.
The gradient form of the bilevel RL applications are listed below.
\begin{enumerate}[label=(\alph*)]
    \item \textit{RLHF/reward shaping}: Assume $r_x$ is continuously differentiable and Assumption \ref{asp:diff} (b) holds. Then Lemma \ref{lemma:derivative} holds and we have
        \begin{align}
            &\nabla_x g(x,y)=-\E\Big[\sum_{t=0}^\infty \gamma^t \nabla r_x(s_t,a_t)\big| s_0\sim\rho,a_0\sim\pi_y(s),\pi_y^*(x)\Big],\nonumber\\
            &\nabla v(x)=-\E\Big[\sum_{t=0}^\infty \gamma^t \nabla r_x(s_t,a_t)\big| s_0\sim\rho,\pi_y^*(x)\Big]\nonumber
        \end{align}
        where the expectation is taken over the trajectory generated by $\pi_y^*(x)$ and $\Pc$.
     \item \textit{Stackelberg game}: Assume $\pi_x$ is continuously differentiable and  Assumption \ref{asp:diff} (b) holds. Then Lemma \ref{lemma:derivative} holds and we have
        \begin{align}
            &\nabla_x g(x,y)
            = -\E\Big[ \sum_{t=0}^\infty \gamma^t Q_f^{\pi_x,\pi_y^*(x)}(s_t,a_{l,t},a_{f,t})\nabla\log\pi_x(a_{l,t}|s_t)\big|s_0\sim\rho,a_{f,0}\sim\pi_y(s_0),\pi_x,\pi_y^*(x)\Big]\nonumber\\
            &+\E\Big[\sum_{t=1}^\infty \gamma^t \tau h_{f,s_t}(\pi_y^*(x)(s_t))\nabla\log\pi_x(a_{l,t}|s_t)\big|s_0\sim\rho,a_{f,0}\sim \pi_y(s_0),\pi_x,\pi_y^*(x)\Big] \nonumber\\
            &\nabla v(x) = -\E\Big[ \sum_{t=0}^\infty \gamma^t Q_f^{\pi_x,\pi_y^*(x)}(s_t,a_{l,t},a_{f,t})\nabla\log\pi_x(a_{l,t}|s_t)\big|s_0\sim\rho,\pi_x,\pi_y^*(x)\Big] \nonumber\\
            &+\E\Big[\sum_{t=1}^\infty \gamma^t \tau h_{f,s_t}(\pi_y^*(x)(s_t))\nabla\log\pi_x(a_{l,t}|s_t)\big|s_0\sim\rho,\pi_x,\pi_y^*(x)\Big]\nonumber
        \end{align}
        Recall in the Stackelberg setting, $\pi_y^*(x)$ is the optimal follower policy given $\pi_x$; and the expectation is taken over the trajectory generated by $\pi_x,\pi_y^*(x)$ and $\Pc$.
\end{enumerate}
\end{lemma}
The proof is deferred to Appendix \ref{sec:derivative application proof} due to space limitation.

\subsection{A gradient-based algorithm and its convergence}
In the previous subsections, we have addressed the challenges of evaluating $\nabla p(x,y)$, enabling the gradient-based methods to optimize $F_\lambda(x,y)$ in \eqref{eq:bmlambdap}.
However, computing $\nabla p(x_k,y_k)$ possibly requires an optimal policy $\pi_y^*(x_k)$ of the lower-level RL problem $\Mt(x_k)$. Given $x_k$, the lower-level RL problem can be solved with a wide range of algorithms, and we can use an approximately optimal policy parameter $\hat{\pi}_k\approx \pi_y^*(x_k)$ to compute the approximate penalty gradient $\hat{\nabla} p(x_k,y_k; \hat{\pi}_k)\approx \nabla p(x_k,y_k)$. The explicit formula of $\hat{\nabla} p(x_k,y_k; \hat{\pi}_k)$ can be straightforwardly obtained by replacing the optimal policy with its approximate $\hat{\pi}_k$ in the formula of $\nabla p(x_k,y_k)$ presented in Lemmas \ref{lemma:derivative value p application} and \ref{lemma:derivative application}. Therefore, we will defer the explicit formula to Appendix \ref{sec:example gradient estimators} for ease of reading.

Given $\hat{\nabla}p(x_k,y_k;\hat{\pi}_k)$, we can compute the approximate gradient of $F_\lambda$ as $\hat{\nabla} F_\lambda(x_k,y_k;\hat{\pi}_k) \coloneqq \nabla f(x_k,y_k)+\lambda \hat{\nabla} p(x_k,y_k; \hat{\pi}_k)$ and update
\begin{align}
    (x_{k+1},y_{k+1})=\Proj_{\Zc}\Big[(x_k,y_k)-\alpha \hat{\nabla} F_\lambda(x_k,y_k;\hat{\pi}_k)\Big]
\end{align}
where $\Zc=\Xc\times\Yc$, and this optimization process is summarized in Algorithm \ref{alg:PBRL}.


\begin{algorithm}[t]
\caption{PBRL: Penalty-based Bilevel RL Gradient-descent}
\begin{algorithmic}[1]
\STATE Select either the value or the Bellman penalty. Select $(x_1,y_1) \in \Zc\coloneqq\Xc \times\Yc$.
Select step size $\alpha$, penalty constant $\gamma$ and iteration number $K$.

\FOR{$k=1$ {\bfseries to} $K$}
\STATE Given RL problem $\Mt(x_k)$, solve for an approximately optimal policy $\hat{\pi}_k\in\Pi$.
\STATE Compute the penalty's approximate gradient $\hat{\nabla} p(x_k,y_k; \hat{\pi}_k)\approx \nabla p(x_k,y_k)$
\STATE Compute the inexact gradient of $F_\lambda$ as $\hat{\nabla} F_\lambda(x_k,y_k;\hat{\pi}_k) = \nabla f(x_k,y_k)+\lambda \hat{\nabla} p(x_k,y_k; \hat{\pi}_k)$
\STATE $(x_{k+1},y_{k+1})=\Proj_{\Zc}\big[(x_k,y_k)-\alpha \hat{\nabla} F_\lambda(x_k,y_k;\hat{\pi}_k)\big]$
\ENDFOR
\end{algorithmic}
\label{alg:PBRL}
\end{algorithm}

We next study the convergence of PBRL.
To bound the error of the update in Algorithm \ref{alg:PBRL}, we make the following assumption on the sub-optimality of the policy $\hat{\pi}_k$.
\begin{Assumption}[Oracle accuracy]\label{asp:orcale accuracy}
    Given some accuracy $\epsilon_{orac}$ and step size $\alpha$, assume the following inequality holds
    \begin{align}
   \frac{1}{K}\sum_{k=1}^K 20\lambda^2\|\hat{\nabla} p(x_k,y_k; \hat{\pi}_k)-\nabla p(x_k,y_k)\|^2 \leq \epsilon_{orac} + \frac{1}{K}\sum_{k=1}^K\frac{1}{\alpha^2}\|(x_{k+1},y_{k+1})-(x_k,y_k)\|^2.
    \end{align}
\end{Assumption}
This assumption only requires the running average of the error to be upper bounded, which is milder than requiring the error to be upper bounded for each iteration. A sufficient condition of the above assumption is $\|\hat{\pi}_k-\pi_y^*(x_k)\|^2 \leq \epsilon_{orac}$ with some constant $c$, which can be achieved by the policy mirror descent algorithm (see e.g., \citep{lan2023policy, zhan2023policy}) with iteration complexity $\mathcal{O}(\log( \lambda^2/\epsilon_{orac}))$ (see a justification in Appendix \ref{sec:example gradient estimators}).

Furthermore, to guarantee worst-case convergence, the regularity condition that $f$ and $p$ are Lipschitz-smooth is required. We thereby identify a set of sufficient conditions for the value penalty or Bellman penalty to be smooth.
\begin{Assumption}[Smoothness assumption]\label{asp:penalty smooth}
     Assume given any $(s,a)$, $h_s(\pi_y(s))$ is $L_h$-Lipschitz smooth on $\Yc$; and $Q_{\Mt(x)}^{\pi_y}(s,a)$, $V_{\Mt(x)}^{\pi_y}(s)$ are $L_v$-Lipschitz-smooth on $\Xc\times\Yc$.
\end{Assumption}
Assumption \ref{asp:penalty smooth} is satisfied under a smooth $r_x$ and a smooth policy (e.g., softmax policy \citep{mei2020softmax}), or a direct policy parameterization paired with smooth regularization function $h_s$. See a detailed justification of this in Appendix \ref{sec:justification of smooth value}.
\begin{lemma}[Lipschitz smoothness of penalty functions]\label{lemma:value smooth}
Under Assumptions \ref{asp:diff} and \ref{asp:penalty smooth}, the value or Bellman penalty function $p(x,y)$ is $L_p$-Lipschitz-smooth on $\Xc\times\Yc$ with constant $L_p$ specified in the proof.
\end{lemma}
We refer the reader to Appendix \ref{sec:smooth proof} for a proof.
Given the smoothness of the penalty terms, we make the final regularity assumption on $f$.
\begin{Assumption}\label{asp:f smooth}
    Assume there exists constant $L_f$ such that $f(x,y)$ is $L_f$-Lipschitz smooth in $(x,y)$.
\end{Assumption}
The projected gradient is a commonly used metric in the convergence analysis of projected gradient type algorithms \citep{ghadimi2016mini}. Define the projected gradient of $F_\lambda(x,y)$ as
\begin{align}\label{eq:nonconvex proximal grad}
    G_\lambda(x_k,y_k) \coloneqq \frac{1}{\alpha}\big((x_k,y_k)-(\Bar{x}_{k+1},\Bar{y}_{k+1})\big),
\end{align}
where $(\Bar{x}_{k+1},\Bar{y}_{k+1})\coloneqq \Proj_{\Zc}((x_k,y_k)-\alpha \nabla F_\lambda (x_k,y_k))$. Now we are ready to present the convergence theorem of PBRL.
\begin{theorem}[Convergence of PBRL]\label{theorem:PBRL convergence}
    Consider running the PBRL algorithm. Suppose Assumptions \ref{asp:diff}--\ref{asp:f smooth} hold. Choose step size $\alpha \leq \frac{1}{L_f+\lambda L_p}$, then we have
    \begin{equation}
    \frac{1}{K}\sum_{k=1}^K \|G_\lambda(x_k,y_k)\|^2 \leq \frac{16\big(F_\lambda(x_1,y_1)-\inf_{(x,y)\in\Zc} f(x,y)\big)}{\alpha K}+\epsilon_{orac}\nonumber
\end{equation}
\end{theorem}
See Appendix \ref{sec:PBRL convergence proof} for the proof of above theorem.
At each outer iteration $k$, let ${\rm com}(\epsilon_{orac})$ be the oracle's iteration complexity. Then the above theorem suggests Algorithm \ref{alg:PBRL} has an iteration complexity of $\mathcal{O}(\lambda\epsilon^{-1}{\rm com}(\epsilon_{orac}))$. When choosing the oracle as policy mirror descent so that ${\rm com}(\epsilon_{orac})=\mathcal{O}(\log (\lambda^2/\epsilon_{\rm orac}))$ \citep{lan2023policy,zhan2023policy}, we have Algorithm \ref{alg:PBRL} has an iteration complexity of $\tilde{\mathcal{O}}(\lambda\epsilon^{-1})$.

\section{Bilevel RL with Lower-level Zero-sum Games}\label{sec:bilevel rl zero sum}
In the previous sections, we have introduced a penalty method to solve the bilevel RL problem with a single-agent lower-level MDP. In this section, we seek to extend the previous idea to the case where the lower-level problem is a zero-sum Markov game \citep{shapley1953stochastic,littman2001friend}. We will first introduce the formulation of bilevel RL with a zero-sum Markov game as the lower-level problem, and then propose its penalty reformulation with a suitable penalty function. Finally, we establish the finite-time convergence for a projected policy gradient-type bilevel RL algorithm.

\subsection{Formulation}\label{sec:bilevel rl zero sum formulation}
Given a parameter $x\in \reals^{d_x}$, consider a parameterized two-player zero-sum Markov game $\Mc_\tau(x)= \{\Sc, \Ac, r_x, \Pc_x, \tau h\}$ where $\Sc$ is a finite state space; $\Ac=\Ac_1\times\Ac_2$ is a finite joint action space, and $\Ac_1, \Ac_2$ are the action spaces of player 1 and 2 respectively; $r_x(s,a)$ ($a=(a_1,a_2)$ is the joint action) is player 1's parameterized reward, and player 2's reward is $-r_x$; the parameterized transition distribution $\Pc_x$ specifies $\Pc_x(s'|s,a)$, which is the probability of the next state being 
$s'$ given when the current state is $s$ and the players take joint action $a$.
Furthermore, we let 
$\pi_i\in\Pi_i$ denote player $i$'s policy, where $\pi_i(a_i|s)$ is the probability of player $i$ taking action $a_i$ given state $s$. 
Here $\Pi_i$ is the  policy class of player $i$ and we assume it is a convex set. We let 
 $\pi\in\Pi=\Pi_1\times\Pi_2$ denote  the joint policy.

Let $\tau h$ be a regularization  parameter and $h_s$ be a regularization function at each state $s\in \mathcal{S} $.  
Given the joint policy $\pi=(\pi_1,\pi_2)$, the (regularized) value function under $\pi$ is defined as 
\begin{align}\label{eq:vpi zero sum}
    V_{\Mt(x)}^{\pi_1,\pi_2}(s)=V_{\Mt(x)}^\pi(s) \coloneqq \E\Big[\sum_{t=0}^\infty \gamma^t \big(r_x(s_t,a_t)-\tau h_{s_t}(\pi_1(s_t))+\tau h_{s_t}(\pi_2(s_t))\big) \big| s_0=s,\pi\Big]
\end{align}
where the expectation is taken over the trajectory generated by $a_t\sim(\pi_1(s_t),\pi_2(s_t)),s_{t+1}\sim\Pc(s_t,a_t)$.
Given some state distribution $\rho$, we write $V_{\Mt(x)}^\pi(\rho)=\E_{s\sim\rho}[V_{\Mt(x)}^\pi(s)]$. 
We can also define the Q function as
\begin{align}\label{eq:qpi zero sum}
    Q_{\Mt(x)}^{\pi_1,\pi_2}(s,a_1,a_2)=Q_{\Mt(x)}^\pi(s,a) \coloneqq r(s,a)+\gamma\E_{s'\sim\Pc(s,a)}\big[V_{\Mt(x)}^\pi(s')\big].
\end{align}

With a state distribution $\rho$ that satisfies $\min_s\rho(s)>0$, the $\epsilon$-Nash-Equilibrium (NE) \citep{ding2022independent,zhang2023model,ma2023decentralized} is a joint policy where $\pi=(\pi_1,\pi_2)$ satisfies
\begin{align}
    NE_\epsilon(x)\coloneqq \Big\{(\pi_1,\pi_2)\in\Pi:&~V_{\Mt(x)}^{\pi_1,\pi_2}(\rho)\geq V_{\Mt(x)}^{\pi_1',\pi_2}(\rho)-\epsilon,~\forall \pi_1'\in\Pi_1 \text{ and }\nonumber\\
    &~V_{\Mt(x)}^{\pi_1,\pi_2}(\rho)\leq V_{\Mt(x)}^{\pi_1,\pi_2'}(\rho)+\epsilon,~\forall \pi_2'\in\Pi_2\Big\}. 
\end{align}
Then Nash equilibrium is defined as $\epsilon$-NE with $\epsilon=0$ and $NE(x)=NE_0(x)$.

\vspace{3pt}

{\noindent \bf Bilevel RL.} In the bilevel RL problem, we are interested in finding the optimal parameter $x$ such that the Nash equilibrium induced by such a parameter, maximizes an objective function $f$. The mathematical formulation is given as follows: 

{\centering
\begin{tcolorbox}[width=0.57\textwidth, right=-20pt]
\leqnomode
\vspace{-0.35cm}
\begin{align}\label{eq:NE problem}
    \min_{x,\pi} f(x,\pi),~~{\rm s.t.}~x\in\Xc,~ \pi\in NE(x).
\end{align}
\vspace*{-0.55cm}
\reqnomode
\end{tcolorbox}}
In the above problem, we aim to find a parameter $x$ and select among all the Nash Equilibria under $x$ such that a certain loss function $f$ is minimized. We next present a motivating example for this general problem.

\noindent\textbf{Motivating example: incentive design.}  Adaptive incentive design \citep{ratliff2019perspective} involves an incentive designer that tries to manipulate self-interested agents by modifying their payoffs with carefully designed incentive functions. In the case where the agents are playing a zero-sum game, the incentive designer's problem \citep{yang2021adaptive} can be formulated as \eqref{eq:NE problem}, given by
\begin{align}\label{eq:incentive}
    \min_{x,\pi} f(\pi)=-\E_{\pi,\Pc_{id}}\Big[\sum_{t=0}^\infty \gamma^t r_{\rm id}(s_t,a_t)-c(s_t)\Big]\,\,\,\,~~~{\rm s.t.}~x\in\Xc,~ \pi\in NE(x)
\end{align}
where $\Pc_{id}(\cdot|s,a)$ is the transition distribution of the designer; $r_{\rm id}$ is the designer's reward, e.g., the social welfare reward \citep{yang2021adaptive}; the function $c(s)$ is the designer's cost; the expectation is taken over the trajectory generated by agents' joint policy $\pi$ and transition $\Pc_{id}$; and in the lower level, the MDP $\Mt(x)$ is parameterized by $x$ via the incentive reward $r_x$. Note $r_x$ is the agents' reward, which is designed by the designer to control the behavior of the agents such that the designer's reward given by $r_{id}$ and $c$ is maximized.

\subsection{Nikaido-Isoda function as a penalty}
Different from a static bilevel optimization problem, the problem in \eqref{eq:NE problem} does not have an optimization problem in the lower level; instead, it has a more abstract constraint set $\pi\in NE(x)$. Our first step is to formulate the problem in \eqref{eq:NE problem} to a bilevel optimization problem with an optimization reformulation of the Nash equilibrium seeking problem. 
In doing so, we will use the Nikaido-Isoda (NI) function first introduced in \citep{nikaido1955note}. It takes a special form in two-player zero sum games:
\begin{align}\label{eq.ni-fun}
    \psi(x,\pi)\coloneqq \max_{\pi_1\in\Pi_1} V_{\Mt(x)}^{\pi_1,\pi_2}(\rho)-\min_{\pi_2\in\Pi_2} V_{\Mt(x)}^{\pi_1,\pi_2}(\rho).
\end{align}
We have the following basic property of this function.
\begin{lemma}[Bilevel formulation]\label{lemma:bilevel formulation of NE}
Given any $x$ and $\pi\in\Pi$, $\psi(x,\pi)\geq 0$, $\psi(x,\pi)\leq 2\epsilon$ if $\pi\in NE_\epsilon(x)$ and $\pi\in NE_\epsilon(x)$ if $\psi(x,\pi)\leq \epsilon$. Therefore, \eqref{eq:NE problem} is equivalent to the following bilevel optimization problem
\begin{align}\label{eq:bilevel NE}
    \mathcal{BZ}:~\min_{x,\pi} f(x,\pi)~~\,\,~\,\,{\rm s.t.}~~x\in\Xc,~ \pi\in\arg\min_{\pi\in\Pi} \psi(x,\pi).
\end{align}
\end{lemma}
\begin{proof}
From the definition \eqref{eq.ni-fun}, we have
    \begin{align}\label{eq.pf-lemma11}
    \psi(x,\pi)= \big(-V_{\Mt(x)}^{\pi_1,\pi_2}(\rho)+\max_{\pi_1\in\Pi_1} V_{\Mt(x)}^{\pi_1,\pi_2}(\rho)\big)+\big(V_{\Mt(x)}^{\pi_1,\pi_2}(\rho)-\min_{\pi_2\in\Pi_2} V_{\Mt(x)}^{\pi_1,\pi_2}(\rho)\big).
\end{align}
The result follows immediately since both terms in the RHS of \eqref{eq.pf-lemma11} are nonnegative on $\Pi$.
\end{proof}
By the above lemma, $\psi(x,\pi)$ is an optimality metric of the lower-level NE-seeking problem. Therefore, it is natural to consider when $\psi$ is a suitable penalty.
Define the penalized problem as
\begin{align}\label{eq:penalized prob zero sum}
    \mathcal{BZ}_{\lambda p}: \min_{x,\pi} f(x,\pi)+\lambda \psi(x,\pi),~{\rm s.t.}~x\in\Xc,~\pi\in\Pi.
\end{align}
To relate $\mathcal{BZ}_{\lambda p}$ with the original problem $\mathcal{BZ}$, certain structures of $\psi(x,\cdot)$ is required. 
Special structure of $\psi$ has been studied in previous works where each player's payoff is non-Markovian (see e.g., \citep{von2009optimization}). While the result relies on certain monotonicity conditions on the payoff functions that do not hold in our Markovian setting. Instead, inspired by the previously discussed single-agent case, we prove a gradient dominance condition under the following assumption.
\begin{Assumption}\label{asp:diff NI p}
    Assume $\tau>0$ and $V_{\Mt(x)}^{\pi_1,\pi_2}(\rho)$ is continuously differentiable in $(x,\pi_1,\pi_2)$.
\end{Assumption}
For a justification of the stronger version of this assumption, please see Appendix \ref{sec:justification of smoothness zero sum}.
Under this assumption, we can prove the following key lemma.
\begin{lemma}[Gradient dominance of $\psi$]\label{lemma:gradient dominance NI}
    If Assumption \ref{asp:diff NI p} holds, then we have the following.
    \begin{enumerate}[label=(\alph*)]
        \item Function $\psi$ is differentiable with $\nabla_\pi \psi(x,\pi)=\big(-\nabla_{\pi_1} V_{\Mt(x)}^{\pi_1,\pi_2^*}(\rho),\nabla_{\pi_2} V_{\Mt(x)}^{\pi_1^*,\pi_2}(\rho)\big)$ where $\pi_1^*\coloneqq \argmax_{\pi_1\in\Pi_1}V_{\Mt(x)}^{\pi_1,\pi_2}(\rho)$ and $\pi_2^*$ defined similarly; and $\nabla_x \psi(x,\pi) = \nabla_x V_{\Mt(x)}^{\pi_1^*,\pi_2}(\rho)-\nabla_x V_{\Mt(x)}^{\pi_1,\pi_2^*}(\rho)$.
        \item There exists a constant $\mu=(1-\gamma)\min_s \rho(s)$ such that given any $x$ and $\tau>0$, $\psi(x,\pi)$ is $\mu$-gradient dominated in $\pi$:
        \begin{align}\label{lemma:gradient dominance NI b}
    \max_{\pi'\in\Pi}\ip{\nabla_\pi \psi(x,\pi)}{\pi-\pi'} \geq \mu \psi(x,\pi),~\forall \pi\in\Pi.
\end{align}
    \end{enumerate}
\end{lemma}
Please see Appendix \ref{sec:gradient dominance NI proof} for the proof.
The proof is based on the gradient dominance condition of the single-agent setting in Lemma \ref{lemma:gradient dominance direct param}, along with the max-min special form of the NI function.

With Lemma \ref{lemma:gradient dominance NI}, we are ready to relate $\mathcal{BZ}_{\lambda p}$ with $\mathcal{BZ}$.

\begin{lemma}[Relation on solutions]\label{lemma:solution relation bilevel rl zero sum}
Assume Assumption \ref{asp:diff NI p} holds and $f(x,\pi)$ is $L$-Lipschitz-continuous in $\pi$. Given accuracy $\delta\!>\!0$, choose $\lambda\geq \delta^{-1}$. If $(x_\lambda,\pi_\lambda)$ is a local/global solution of $\mathcal{BZ}_{\lambda p}$, it is a local/global solution of the relaxed $\mathcal{BZ}$ with some $\epsilon_\lambda \leq \delta$:
\begin{align}
    \mathcal{BZ}_\epsilon: \min_{x,\pi} f(x,\pi),~{\rm s.t.}~x\in\Xc,~\pi\in\Pi,~\psi(x,\pi)\leq \epsilon_\lambda.
\end{align}
\end{lemma}
The proof is deferred to Appendix \ref{sec:solution relation bilevel rl zero sum proof}.
The above lemma shows one can recover the local/global solution of the approximate problem of $\mathcal{BZ}$ by solving $\mathcal{BZ}_{\lambda p}$ instead. To solve for $\mathcal{BZ}_{\lambda p}$, we propose a projected gradient type update next and establish its finite-time convergence.

\subsection{A policy gradient based algorithm and its convergence analysis}
To solve for $\mathcal{BZ}$, we consider a projected gradient update to solve for its penalized problem $\mathcal{BZ}_{\lambda p}$. To evaluate the objective function in $\mathcal{BZ}_{\lambda p}$, one will need to evaluate $\nabla \psi(x,\pi)$. Note that evaluating $\nabla_\pi \psi(x,\pi)$ requires the point $\pi_1^*(\pi_2,x)$ and $\pi_2^*(\pi_1,x)$ (defined in Lemma \ref{lemma:gradient dominance NI}), which are optimal policies of a fixed MDP given parameters $(\pi_2,x)$ and $(\pi_1,x)$ respectively. 

There are various efficient algorithms to find the optimal policy of a regularized MDP. Thus we assume that at each iteration $k$, we have access to some approximate optimal polices $\hat{\pi}_1^k \approx \pi_1^*(\pi_2^k,x^k)$ and $\hat{\pi}_2 \approx \pi_2^*(\pi_1^k,x^k)$ obtained by certain RL algorithms. 
With $\hat{\pi}^k=(\hat{\pi}_1^k,\hat{\pi}_2^k)$, we may denote the estimator of $\nabla \psi(x^k,\pi^k)$ as $\hat{\gr}\psi(x^k,\pi^k;\hat{\pi}^k)$, the definition of which follows from Lemma \ref{lemma:gradient dominance NI} (a) with $\hat{\pi}_1^k$ and $\hat{\pi}_2^k$ in place of $\pi_1^*$ and $\pi_2^*$ respectively:
\begin{equation}
\hat{\gr}\psi(x^k,\pi^k;\hat{\pi}^k)\coloneqq \Big(\nabla_x V_{\Mt(x^k)}^{\hat{\pi}_1^k,\pi_2^k}(\rho)-\nabla_x V_{\Mt(x^k)}^{\pi_1^k,\hat{\pi}_2^k}(\rho),\big(-\nabla_{\pi_1} V_{\Mt(x^k)}^{\pi_1^k,\hat{\pi}_2^k}(\rho),\nabla_{\pi_2} V_{\Mt(x^k)}^{\hat{\pi}_1^k,\pi_2^k}(\rho)\big)\Big).
\end{equation}
We then perform projected gradient type update with this estimator:
\begin{align}\label{eq:projected grad update zero sum}
    (x^{k+1},\pi^{k+1})=\Proj_{\Zc}\Big[(x^k,\pi^k)-\alpha \big(\nabla f(x^k,\pi^k)+\lambda \hat{\gr}\psi(x^k,\pi^k;\hat{\pi}^k)\big)\Big]
\end{align}
where $\Zc=\Xc\times\Pi$. We make the following assumption on the sub-optimality of $\hat{\pi}_1^k$ and $\hat{\pi}_2^k$.
\begin{Assumption}[Oracle accuracy]\label{asp:oracle accuracy zero sum}
    Given some pre-defined accuracy $\epsilon_{\rm orac}>0$ and the step size $\alpha$, assume the approximate policies $\hat{\pi}_1^k$ and $\hat{\pi}_2^k$ satisfy the following inequality
    \begin{align}
   \frac{1}{K}\sum_{k=1}^K 20\lambda^2\|\hat{\nabla} \psi(x^k,\pi^k; \hat{\pi}^k)-\nabla \psi(x^k,\pi^k)\|^2 \leq \epsilon_{\rm orac} + \frac{1}{K}\sum_{k=1}^K\frac{1}{\alpha^2}\|(x^{k+1},\pi^{k+1})-(x^k,\pi^k)\|^2.
    \end{align}
\end{Assumption}
The left-hand side of the above inequality can be upper bounded by the optimality gaps of the approximate optimal policies $\{\hat{\pi}_1^k,\hat{\pi}_2^k\}$. Note that here the policies  $\{\hat{\pi}_1^k,\hat{\pi}_2^k\}$ are not approximate NE. 
Instead, $\hat{\pi}_1^k$ is a   player 1's approximately optimal policy on the Markov model with parameter $x_k$, where player 2 adopts $ \pi_2^k$. 
Thus, to obtain $\hat{\pi}_1^k$, 
one may use efficient single-agent policy optimization algorithms. For example, when using the policy mirror descent algorithm \citep{zhan2023policy}, it will take an iteration complexity of $\mathcal{O}(\log (\lambda^2/\epsilon_{\rm orac}))$ to solve for accurate enough approximate policies.
Similarly, $\hat{\pi}_2^k$ is an approximately optimal policy of player 2 on the Markov game with parameter $x_k$, where player 1 adopts $\pi_1^k$. Thus $\hat{\pi}_2^k$ can similarly be efficiently obtained using a standard single-agent policy optimization algorithm. Furthermore, a  more detailed justification of this assumption is provided in \ref{sec:oracle accuracy zero sum justification}.


We next identify sufficient conditions for the finite-time convergence in \eqref{eq:projected grad update zero sum} as follows.
\begin{Assumption}[Smoothness assumption of $\psi$]\label{asp:smoothness assumption NI}
    Suppose Assumption \ref{asp:diff NI p} holds. Additionally, assume the following arguments hold.
    \begin{enumerate}[label=(\alph*)]
        \item Given any $s$, $V_{\Mt(x)}^\pi(s)$ is $L_v$-Lipschitz-smooth on $\Xc\times\Pi$;
        \item If the discount factor $\gamma>0$ then assume given $x\in\Xc$, for any state $s$ and initial state-action $s_0,a_0$, there exists $t$ such that $P_x^\pi(s_t=s|s_0,a_0)>0$, where $P_x^\pi(s_t=s|s_0,a_0)$ is the probability of reaching $s$ at time $t$ in the MDP $\Mt(x)$ under joint policy $\pi$.
    \end{enumerate}
\end{Assumption}
Assumption \ref{asp:smoothness assumption NI} (a) can be satisfied under a smooth regularization function, and smooth parameterized functions $r_x$ and $\Pc_x$; see the justification in Appendix \ref{sec:justification of smoothness zero sum}. Assumption \ref{asp:smoothness assumption NI} (b) is in the same spirit as Assumption \ref{asp:diff} (b) in the single-agent case. 
Under Assumption \ref{asp:smoothness assumption NI}, we can prove that the NI function is Lipschitz-smooth.
\begin{lemma}[Smoothness of $\psi$]\label{lemma:smooth NI}
    Under Assumption \ref{asp:smoothness assumption NI}, there exists a constant $L_\psi$ such that $\psi(x,\pi)$ is $L_\psi$-Lipschitz-smooth on $\Xc\times\Pi$.
\end{lemma}
The proof of the above lemma can be found in Appendix \ref{sec:smooth NI proof}.
With the above smoothness condition, we are ready to establish the convergence result.
Define the projected gradient of the objective function in $\mathcal{BZ}_{\lambda p}$ \eqref{eq:penalized prob zero sum} as
\begin{align}\label{eq:nonconvex proximal grad zero sum}
    \mathcal{G}_\lambda(x^k,\pi^k) \coloneqq \frac{1}{\alpha}\big((x^k,\pi^k)-(\Bar{x}^{k+1},\Bar{\pi}^{k+1})\big),
\end{align}
where $(\Bar{x}^{k+1},\Bar{\pi}^{k+1})\coloneqq \Proj_{\Zc}\big((x^k,\pi^k)-\alpha \big(\nabla f (x^k,\pi^k)+\lambda \gr \psi(x^k,\pi^k)\big)\big)$. 
Now we are ready to present the convergence theorem of update \eqref{eq:projected grad update zero sum}.
\begin{theorem}[Convergence of PBRL with zero-sum lower-level]\label{theorem:PBRL convergence zero sum}
    Consider running update \eqref{eq:projected grad update zero sum}. Suppose Assumptions \ref{asp:f smooth}, \ref{asp:diff NI p}, \ref{asp:oracle accuracy zero sum} and \ref{asp:smoothness assumption NI} hold. Choose step size $\alpha \leq \frac{1}{L_f+\lambda L_\psi}$, then we have
    \begin{equation}
    \frac{1}{K}\sum_{k=1}^K \|\mathcal{G}_\lambda(x^k,\pi^k)\|^2 \leq \frac{16\big(f(x^1,\pi^1)+\lambda \psi(x^1,\pi^1)-\inf_{(x,y)\in\Zc} f(x,y)\big)}{\alpha K}+\epsilon_{\rm orac}.
\end{equation}
\end{theorem}
The proof is deferred to Appendix \ref{sec:PBRL convergence zero sum proof}.
At each outer iteration $k$, let ${\rm com}(\epsilon_{\rm orac})$ be the oracle's iteration complexity. Then the above theorem suggests update \eqref{eq:projected grad update zero sum} has an iteration complexity of $\mathcal{O}(\lambda\epsilon^{-1}{\rm com}(\epsilon_{\rm orac}))$.
As discussed under Assumption \ref{asp:oracle accuracy zero sum}, one could use policy mirror descent to solve for $\hat{\pi}_1^k,\hat{\pi}_2^k$ when estimating $\nabla \psi(x^k,\pi^k)$, then we have ${\rm com}(\epsilon_{\rm orac})=\mathcal{O}(\log (\lambda^2/\epsilon_{\rm orac}))$. In such cases, update \eqref{eq:projected grad update zero sum} has an iteration complexity of $\tilde{\mathcal{O}}(\lambda\epsilon^{-1})$.

\section{Simulation}
In this section, we test the empirical performance of PBRL in different tasks.
\subsection{Stackelberg Markov game}
We first seek to solve the Stackelberg Markov game formulated as
\begin{align}
    \min_{x} -V_l^{\pi_x,\pi_y^*(x)}(\rho),~~{\rm s.t.}~x\in\reals^{d_x},~ \pi_y^*(x)=\argmin_{\pi_y} -V_f^{\pi_x,\pi_y}(\rho),
\end{align}
where $\pi_x$ and $\pi_y$ is parameterized via the softmax function.
Here the transition distribution and rewards are randomly generated. It has a state space of size $|\Sc|=100$, and the leader, and follower's action space are of size $|\Ac_l|=5$, $|\Ac_f|=5$ respectively. Each entry of the rewards $R_l,R_f\in \reals^{100\times5\times5}$ is uniformly sampled between $[0,1]$ and values smaller than $0.7$ are set to $0$ to promote sparsity. Each entry of the transition matrix is sampled between $[0,1]$ and then is normalized to be a distribution.

\noindent\textbf{Baseline.} 
 We implement PBRL with both value and Bellman penalty, and compare them with the independent policy gradient method \citep{daskalakis2020independent,ding2022independent}.
 In the independent gradient method, each player myopically maximizes its own value function, i.e., the leader maximizes $V_l^{\pi_x,\pi_y}(\rho)$ while the follower maximizes $V_f^{\pi_x,\pi_y}(\rho)$. At each step $k$, leader updates $\pi_{x_k}$ with one-step gradient of $V_l^{\pi_x,\pi_{y_k}}(\rho)$ while the follower updates $\pi_{y_k}$ with one-step gradient of $V_f^{\pi_{x_k},\pi_y}(\rho)$. We test all algorithms across 10 randomly generated MDPs. 

 \begin{figure*}[htp]
\centering
    \includegraphics[width=0.35\textwidth]{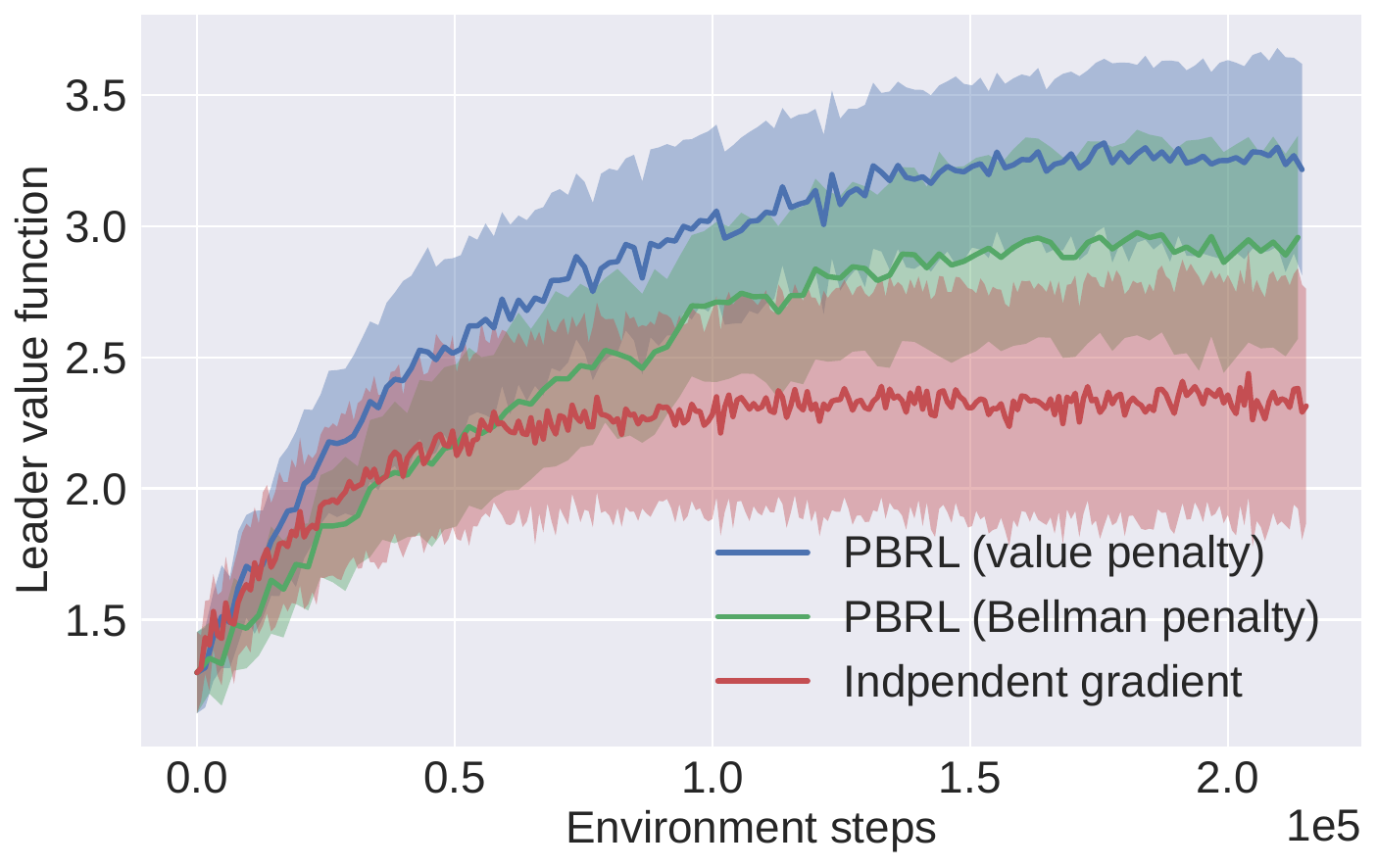}\hspace{0.8cm}
    \includegraphics[width=0.35\textwidth]{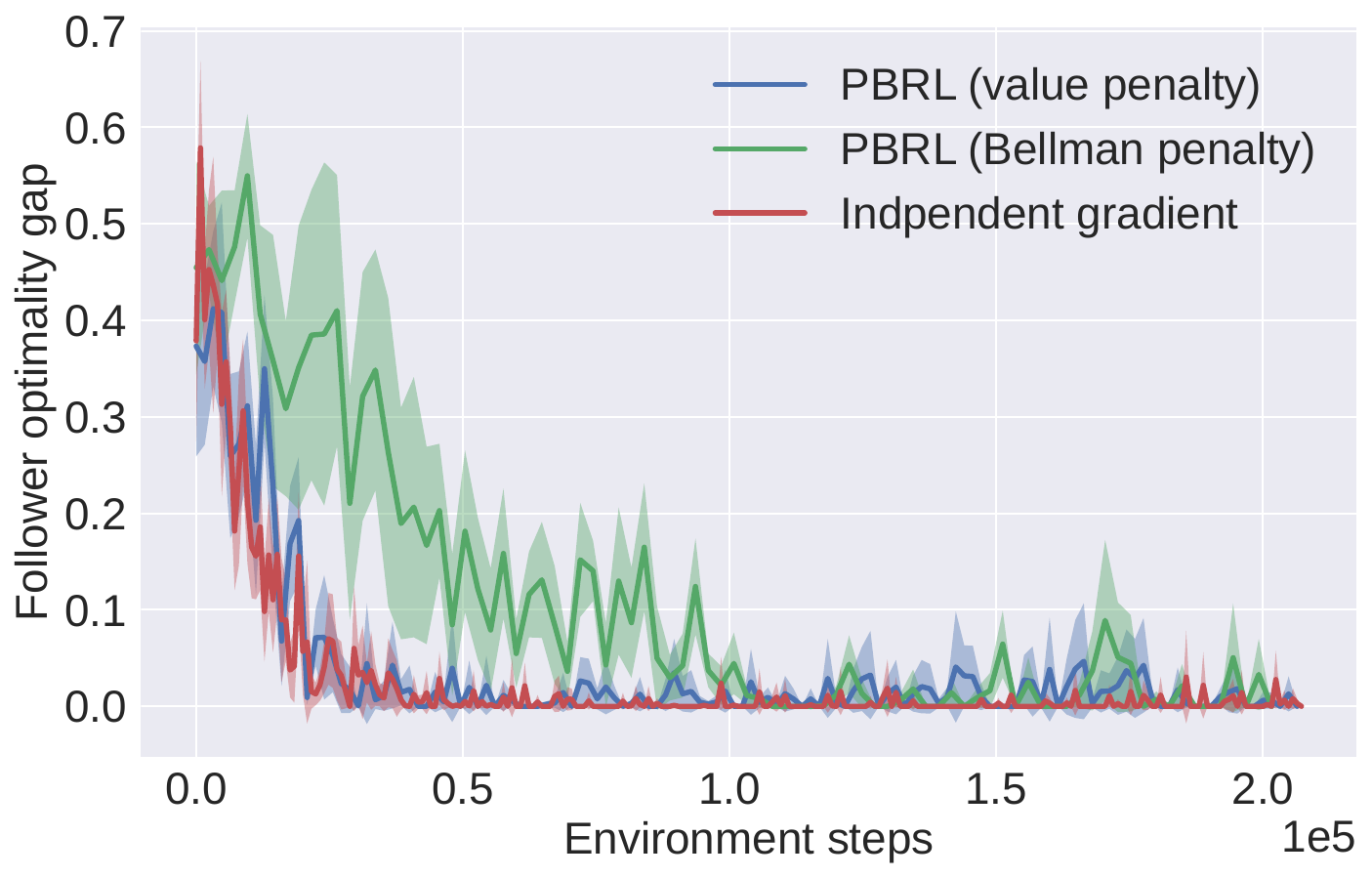}
     \caption{Stackelberg Markov games. The result is generated by running the algorithms in 10 random Stackelberg MDPs. The environment step is the total number of steps taken in the MDP, and is therefore also proportional to the total samples used in training. The leader's value function is $V_l^{\pi_{x_k},\pi_{y_k}}(\rho)$, and the follower's optimality gap is given by $V_f^{\pi_{x_k},\pi_y^*(x_k)}(\rho)-V_f^{\pi_{x_k},\pi_{y_k}}(\rho)$. A zero optimality gap means the follower has found the best response to the leader. }
    \label{fig:name1}
\end{figure*}

 We report the results in Figure \ref{fig:name1}. In the right figure, we can see the follower's optimality gap diminishes to zero, that is, the followers have found their optimal policies. In the mean time, the left figure reports the leaders' total rewards for the three methods.  Overall, we find that both PBRL with value penalty and Bellman penalty outperform the independent gradient: it can be observed from Figure \ref{fig:name1} (left) that PBRL can achieve a higher leader's return than the independent gradient, and the PBRL with value penalty reaches the highest value.

 \begin{figure*}[htp]
\centering
    \includegraphics[width=0.32\textwidth]{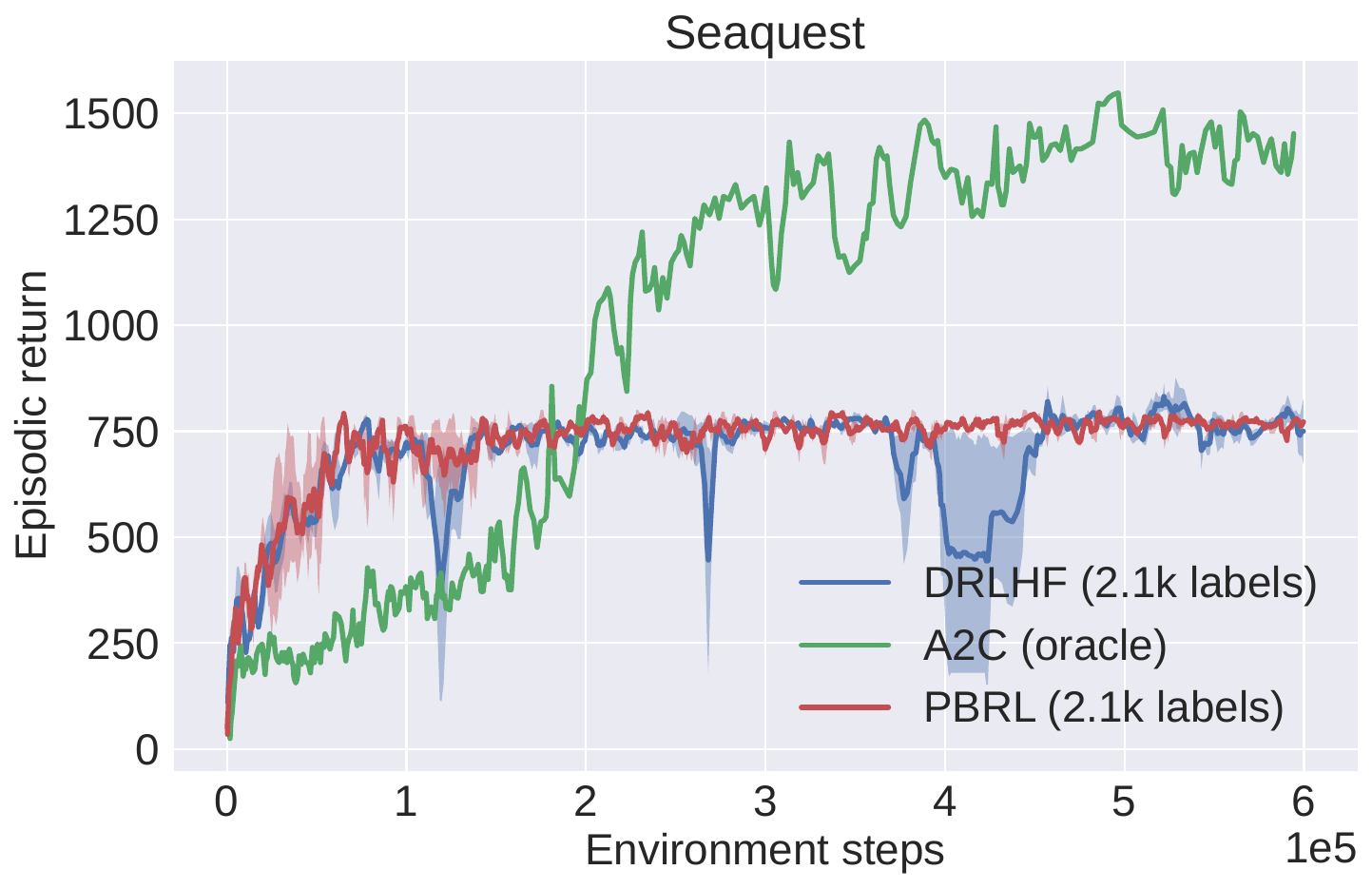}
    \includegraphics[width=0.32\textwidth]{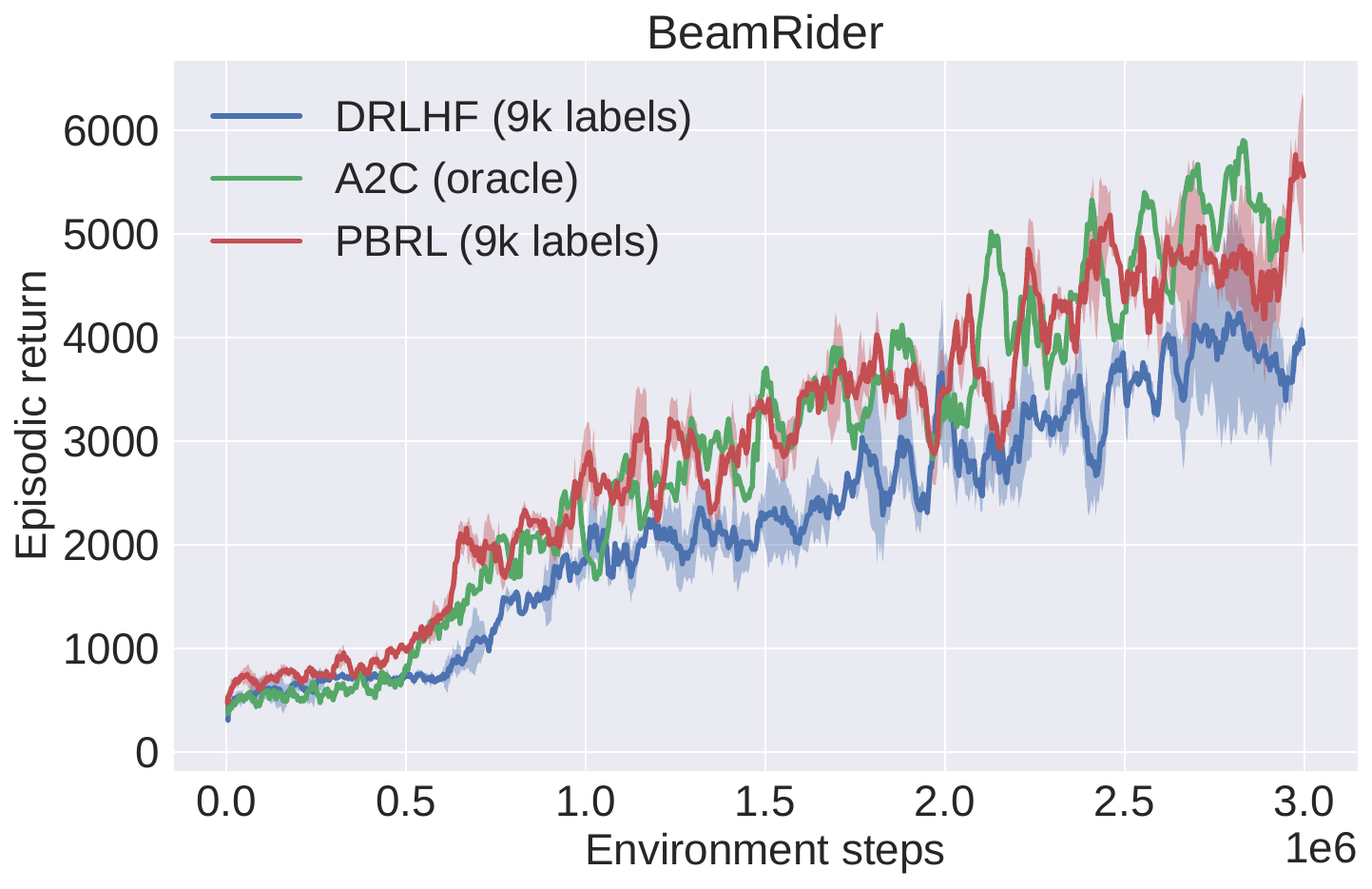}
    \includegraphics[width=0.32\textwidth]{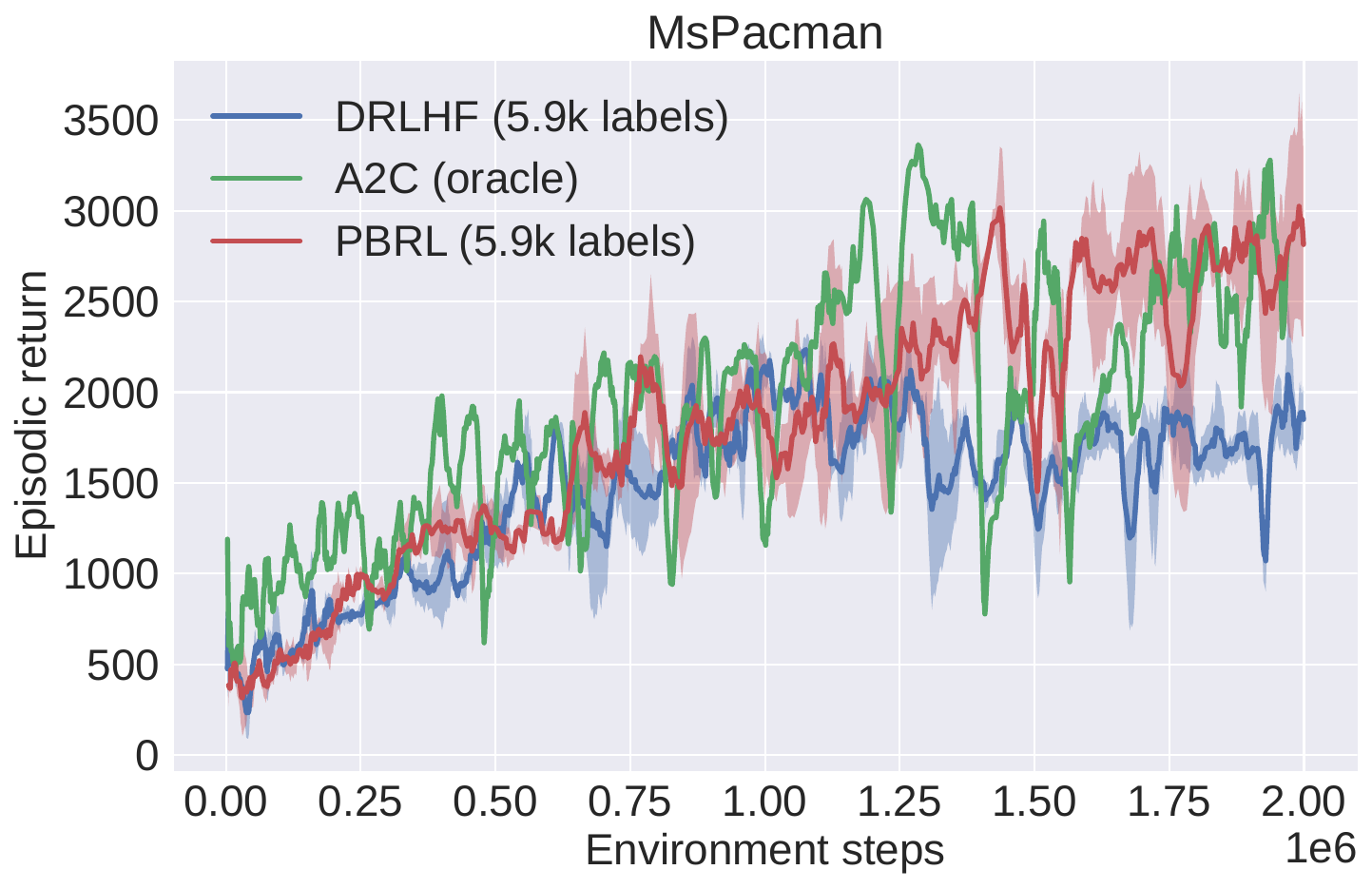} 
     \caption{Performance on Atari games measured by true reward. The `episode return' is the sum of true rewards in an episode. We average the episode return in 5 consecutive episodes. The `environment steps' is the number of steps taken per worker in policy optimization. We compare performance of PBRL (ours) and DRLHF both with few labeled pairs, and A2C with true reward.}
    \label{fig:rlhf}
\end{figure*}

\subsection{Deep reinforcement learning from human feedback}
We test our algorithm in RLHF, following the experiment setting in \citep{christiano2017deep}; see a description of the general RLHF setting in Section \ref{sec:application}.

\noindent\textbf{Environment and preference collection.} We conduct our experiments in the Arcade Learning Environment (ALE) \citep{bellemare2013arcade} through OpenAI gym. The ALE provides the game designer's reward that can be treated as the ground truth reward. For each pair of segments we collect, we assign preference to whichever has the highest ground truth reward. This preference generation process allows us to benchmark our algorithm with DRLHF that also use this process.

\noindent\textbf{Baseline.} We compare PBRL with DRLHF \citep{christiano2017deep} and A2C (A3C \citep{A3C} but synchronous). We use the ground truth reward to train A2C agent, and treat A2C as an oracle algorithm. The oracle algorithm estimates a performance upperbound for other algorithms. 

The results are reported in Figure \ref{fig:rlhf}. The first two games (Seaquest and BeamRider) are also reported in \citep{christiano2017deep}. For Seaquest, the asymptotic performance of DRLHF and PBRL are similar, while DRLHF is more unstable in training. Similar observation can also be made in the original paper of DRLHF. For BeamRider and MsPacman, we find out that PBRL has an advantage over DRLHF on the episode return. It can be observed that PBRL is able to achieve higher best-episode-return than DRLHF, and become comparable to the oracle algorithm.

\subsection{Incentive design}
 \begin{figure*}[htp]
\centering
    \includegraphics[width=0.35\textwidth]{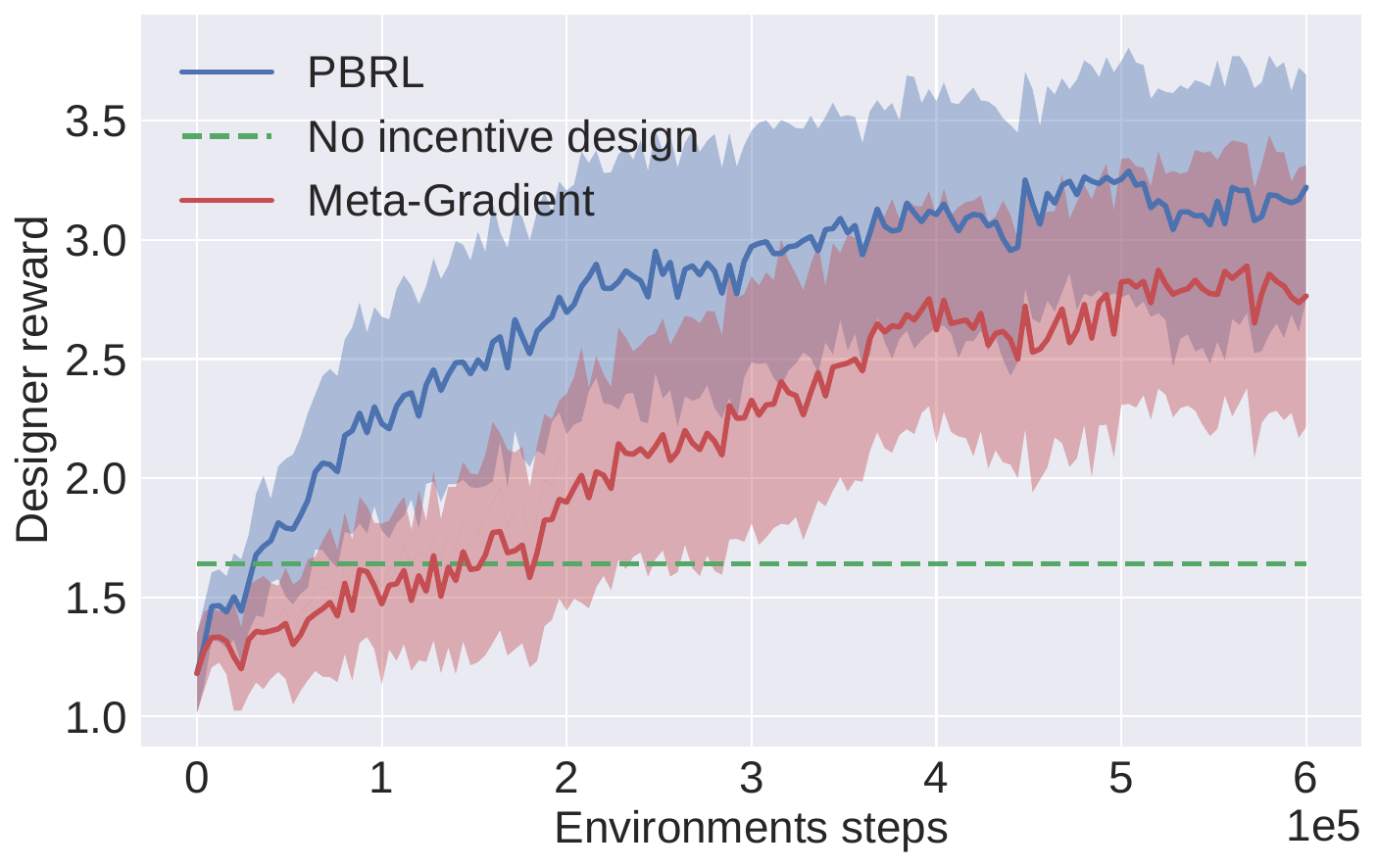}\hspace{0.8cm}
    \includegraphics[width=0.35\textwidth]{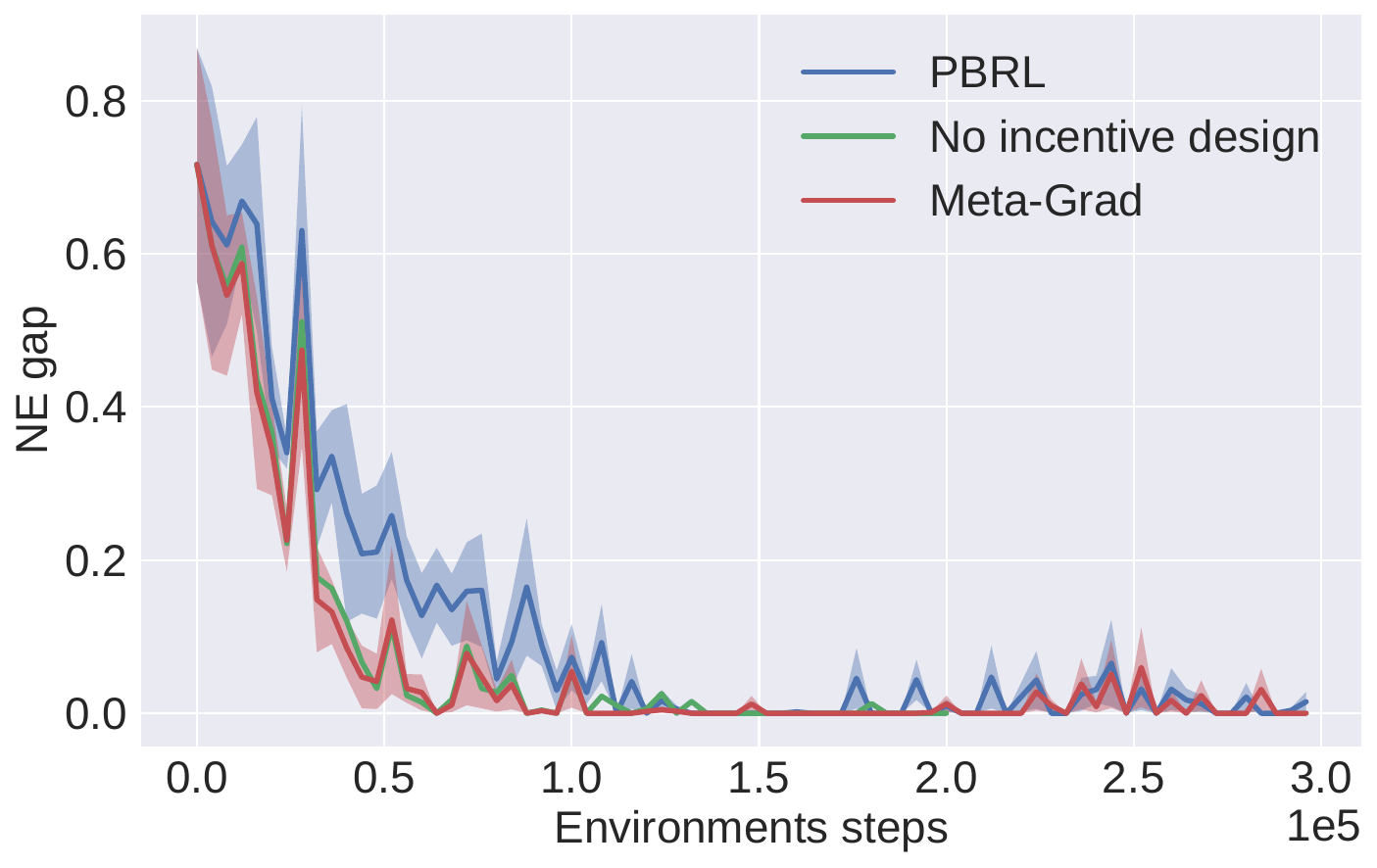}
     \caption{Incentive design. The result is generated by running the algorithms in 5 random environments. The environment step is the total number of steps taken, and is proportional to the total samples used. The designer's reward is $f(\pi)$, which is the expected cumulative designer reward. The NE gap is the estimated value of NI function $\psi(x,\pi)$. A zero NE gap indicates the players have achieved an approximate Nash equilibrium under the $r_x(s,a)$.}
    \label{fig:name3}
\end{figure*}
Here we test our algorithm in the following incentive design problem:
\begin{align}
    \min_{x,\pi} f(\pi)=-\E_{\pi,\Pc_{id}}\Big[\sum_{t=0}^\infty \gamma^t r_{\rm id}(s_t,a_t)\Big]\,\,\,\,~~~{\rm s.t.}~~~\pi\in NE(x).
\end{align}
See a more detailed description of this task in the motivating example of Section \ref{sec:bilevel rl zero sum formulation}.
We have $|\Sc|=10$, $|\Ac_1|=|\Ac_2|=5$. The designer's transition $\Pc_{id}(\cdot|s,a)$ and the lower-level transition $\Pc(\cdot|s,a)$ are randomly generated. Then players' original reward $r(s,a)$ and the designer's reward $r_{id}(s,a)$ are randomly generated between $[0,1]$. The players' reward  $r_x(s,a)=r(s,a)+0.2\sigma(x(s,a))$, where $r$ is the original reward given by the environment, and $\sigma(x(s,a))$ is the incentive reward controlled by the designer. Here $\sigma$ is the sigmoid function and $x\in\reals^{|\Sc|\times|\Ac_1|\times|\Ac_2|}$ is the incentive reward parameter. The $\pi_1,\pi_2$ are softmax policies.

\noindent\textbf{Baseline.}  We implement the PBRL update for zero-sum lower level introduced in Section \ref{sec:bilevel rl zero sum}, and compare it with the Meta-Gradient method \citep{yang2021adaptive}. 
To exclude the case where the original zero-sum game (with no incentive reward) already has a high reward,
we also provide the performance when there is no incentive design, i.e., when $\sigma(x(s,a))$ is kept a constant. This will only return an approximate NE of the lower-level zero-sum problem without incentive reward, and therefore will provide a performance start line. Then an algorithm's output incentive reward is more effective the more it improves over the start line.

    It can be observed from Figure \ref{fig:name3} (right) that both PBRL and Meta-Gradient have found the approximate NE under their respective incentive reward $r_x$. It can be observed from Figure \ref{fig:name3} (left) that the incentive reward $r_x$ of both methods are effective since the designer's reward $f(\pi)$ of both methods exceed the start line (green). While PBRL is able to outperform Meta-Gradient since the incentive reward $r_x$ of PBRL is able to lead to a $\pi\in NE(x)$ with a higher designer reward.


\section{Concluding Remarks}
In this paper, we propose a penalty-based first-order algorithm for the bilevel reinforcement learning problems. In developing the algorithm, we provide results in three aspects: 1) we find penalty function with proper landscape properties such that the induced penalty reformulation admits solutions for the original bilevel RL problem; 2) to develop a gradient-based method, we study the differentiability of the penalty functions and find out their close form gradients; 3) based on the previous findings, we propose the convergent PBRL algorithm and evaluate on the Stackelberg Markov game, RLHF and incentive design.

\bibliography{bob}
\bibliographystyle{plainnat}

\clearpage
\begin{center}
{\large \bf Appendix for\\
``\FullTitle"}
\end{center}
\appendix

\section{Preliminary results}
\begin{lemma}[Lipschitz continuous optimal policy]\label{lemma:lipschitz continuous optimal policy}
    Given $x\in\Xc$, consider the optimal policies in a convex policy class $\Pi$ of a parameterized MDP $\Mc_\tau(x)$. Suppose Assumption \ref{asp:diff} holds, $\tau>0$ and $\Xc$ is compact. Then the optimal policy $\pi_y^*(x)$ is unique and the following inequality hold:
    \begin{align}
        \|\pi_y^*(x)-\pi_y^*(x')\|\leq \tau^{-1}C_J\|x-x'\|,~\forall x,x'\in\Xc
    \end{align}
    where $C_J$ is a constant specified in the proof.
\end{lemma}
 \begin{proof}
    By Lemma \ref{lemma:reformulation}, the optimal policy of $\Mt(x)$ on a convex policy class $\Pi$ is unique, given by
    \begin{align}\label{eq:pistar}
    \pi_y^*(q(x))= \argmin_{\pi\in\Pi}J(q(x),\pi)\coloneqq\E_{s\sim\rho}[\ip{\pi(s)}{q_s(x)}+\tau h_s(\pi(s))]
\end{align}
where recall $q(x)=(q_s(x))_{s\in\Sc}$ with
\begin{align}
    q_s(x)= (-\max_{\pi\in\Pi} Q_{\Mt(x)}^\pi (s,a))_{a\in\Ac}.
\end{align}
We overload the notation $\pi^*$ here with $\pi_y^*(q(x))$ which equals $\pi_y^*(x)$.
In \eqref{eq:pistar},
since $\tau \E_{s\sim\rho}[h_s(\pi(s))]$ is $\tau$-strongly convex at $\pi$ on $\Pi$, $\pi_y^*(q(x))$ satisfies \eqref{eq:pistar} if and only if it is a solution of the following parameterized variational inequality (VI)
\begin{align}\label{eq:VI 1}
    \ip{\nabla_\pi J(q(x),\pi)}{\pi-\pi'} \leq 0,~~\forall \pi'\in\Pi
\end{align}
where
\begin{align}\label{eq:nablaJ}
    \nabla_\pi J(q(x),\pi)=\Big(\rho(s)q_s(x)+\tau\rho(s)\nabla h_s(\pi(s))\Big)_{s\in\Sc}.
\end{align}

First, it can be checked that $\nabla_\pi J(q(x),\pi)$ is continuously differentiable at any $(q(x),\pi)$. Secondly, by the uniform strong convexity of $J(q(x),\cdot)$, given any $q(x)$, it holds that
\begin{align}
    (\pi-\pi')^\top \nabla_\pi^2 J\big(q(x),\pi_y^*(q(x))\big) (\pi-\pi') \geq \tau^{-1}\|\pi-\pi'\|^2.
\end{align}
Given these two properties of the VI, it then follows from \cite[Theorem 2F.7]{dontchev2009implicit} that the solution mapping $\pi_y^*(q(x))$ is $\tau^{-1}$-Lipschitz-continuous locally at any point $q(x)$. Thus 
 $\pi_y^*(q(x))$ is $\tau^{-1}$-Lipschitz-continuous in $q(x)$ globally, yielding
\begin{align}
    \|\pi_y^*(q(x))-\pi_y^*(q(x'))\|
&\leq \tau^{-1}\|q(x)-q(x')\| \nonumber\\
&\leq \tau^{-1}\max_{x\in\Xc}\|\nabla q(x)\|\|x-x'\| \nonumber\\
&= \tau^{-1}C_J\|x-x'\|
\end{align}
where the second inequality follows from $q(x)$ is continuously differentiable, which can be checked by Lemma \ref{lemma:derivative} under Assumption \ref{asp:diff} and the continuity of $\pi_y^*(x)$ we proved earlier; and, $C_J= \max_{x\in\Xc}\|\nabla q(x)\|$ is well-defined by compactness of $\Xc$.
 \end{proof}

\section{Proof in Section \ref{sec:formulation} and \ref{sec:penalty reform}}
\subsection{Proof that Stackelberg Markov game is a bilevel RL problem}\label{sec:stackelberg proof}
\begin{lemma}[Stackelberg game cast as $\mathcal{BM}$]\label{lemma:stackelberg}
    The Stackelberg MDP from the follower's viewpoint can be defined as a parametric MDP:
    $$\Mt(x)=\{\Sc,\Ac_f,r_x(s,a_f)=\E_{a_l\sim\pi_x(s)}[r_l(s,a_l,a_f)],\Pc_x(\cdot|s,a_f)=\E_{a_l\sim\pi_x(s)}[\Pc(\cdot|s,a_l,a_f)],\tau h_f\}.$$
    Then we have $V_f^{\pi_x,\pi_y}(s)=V_{\Mt(x)}^{\pi_y}(s),\forall s$, and thus the original formulation of Stackelberg game in \eqref{eq:SG original} can be rewritten as $\mathcal{BM}$:
    \begin{align}
        \mathcal{SG}: \min_{x,y} -V_l^{\pi_x,\pi_y}(\rho),~~{\rm s.t.}~x\in\Xc,~ y\in \Yc^*(x)=\argmin_{y\in\Yc}-V_{\Mt(x)}^{\pi_y}(\rho).
    \end{align}
\end{lemma}
\begin{proof}
Recall that the follower's value function $V_f^{\pi_x,\pi_y}(s)$ under the leader's policy $\pi_x$ and the follower's policy $\pi_y$ is defined as
\begin{align}\label{eq:idk231}
    V_f^{\pi_x,\pi_y}(s)
    &=\E\Big[\sum_{t=0}^\infty \gamma^t \big(r_f(s_t,a_{l,t},a_{f,t})-\tau h_{f,s_t}(\pi_y(s_t))\big) \big| s_0=s,\pi_x,\pi_y\Big]
\end{align}
where the leader's action $a_{l,t}\sim\pi_l(s_t)$, the follower's action $a_{f,t}\sim\pi_f(s_t)$, and the state transition follows $s_{t+1}\sim\Pc(\cdot|s_t,a_{l,t},a_{f,t})$.

It then follows from a expansion of the expectation in \eqref{eq:idk231} that
\begin{align}
V_f^{\pi_x,\pi_y}(s)
    &=\E_{a_{l,0}\sim\pi_x(s_0),a_{f,0}\sim\pi_y(s_0)}\Big[r_f(s_0,a_{l,0},a_{f,0})-\tau h_{f,s_0}(\pi_y(s_0))\big| s_0=s,\pi_x,\pi_y\Big] \nonumber\\
    &+\gamma \E_{\substack{a_{l,0}\sim\pi_x(s_0),a_{f,0}\sim\pi_y(s_0)\\s_1\sim\Pc(s_0,a_{l,0},a_{f,0})\\a_{l,1}\sim\pi_x(s_1),a_{f,1}\sim\pi_y(s_1)}}\Big[r_f(s_1,a_{l,1},a_{f,1})-\tau h_{f,s_1}(\pi_y(s_1))\big| s_0=s,\pi_x,\pi_y\Big]+\dots\nonumber\\
    &=\E_{a_{f,0}\sim\pi_y(s_0)}\Big[r_x(s_0,a_{f,0})-\tau h_{f,s_0}(\pi_y(s_0))\big| s_0=s,\pi_y\Big] \nonumber\\
    &+\gamma \E_{\substack{a_{f,0}\sim\pi_y(s_0)\\s_1\sim\Pc_x(s_0,a_{f,0})\\a_{f,1}\sim\pi_y(s_1)}}\Big[r_x(s_1,a_{f,1})-\tau h_{f,s_1}(\pi_y(s_1))\big| s_0=s,\pi_y\Big]+\dots \nonumber\\ 
    &= V_{\Mt(x)}^{\pi_y}(s)
\end{align}
where recall $\mathcal{P}_x(s,a_f)=\E_{a_l\sim\pi_x(s)}[\Pc(\cdot|s,a_l,a_f)]$ and $r_x(s,a_f)=\E_{a_l\sim\pi_x(s)}[r_l(s,a_l,a_f)]$.
Thus we have $V_f^{\pi_x,\pi_y}(s)=V_{\Mt(x)}^{\pi_y}(s),\forall s$. Therefore, the Stackelberg Markov game can be written as $\mathcal{BM}$.
\end{proof}

\subsection{Proof of Lemma \ref{lem:value penalty solution}}\label{sec:value penalty solution proof}
\begin{proof}
    Since $(x_\lambda,y_\lambda)$ is an $\epsilon$-minima of $\mathcal{BM}_{\lambda p}$, it holds for any $x\in\Xc$ and $y\in\Yc$ that
    \begin{align}\label{eq:idk24}
        f(x_\lambda,y_\lambda)+\lambda \big(-V_{\Mt(x_\lambda)}^{\pi_{y_\lambda}}(\rho)+\max_{y\in\Yc} V_{\Mt(x_\lambda)}^{\pi_{y}}\big) \leq f(x,y)+\lambda \big(-V_{\Mt(x)}^{\pi_y}(\rho)+\max_{y\in\Yc} V_{\Mt(x)}^{\pi_y}\big)+\epsilon.
    \end{align}
    Choosing $x=x_\lambda$ and $y\in\Yc(x_\lambda)$ in the above inequality and rearranging yields
    \begin{align}
        \max_{y\in\Yc} V_{\Mt(x_\lambda)}^{\pi_y}(\rho)-V_{\Mt(x_\lambda)}^{\pi_{y_\lambda}}(\rho) 
        &\leq \frac{1}{\lambda } \big(f(x_\lambda,y_\lambda)-f(x_\lambda,y)+\epsilon\big)\nonumber\\
        &\leq \frac{1}{\lambda } \big(C+\epsilon\big) \leq \delta+\lambda^{-1}\epsilon.
    \end{align}
    Define $\epsilon_\lambda\coloneqq \max_{y\in\Yc} V_{\Mt(x_\lambda)}^{\pi_y}(\rho)-V_{\Mt(x_\lambda)}^{\pi_{y_\lambda}}(\rho)$ then $\epsilon_\lambda \leq \delta+\lambda^{-1}\epsilon$. It follows from \eqref{eq:idk24} that for any $x,y$ feasible for \eqref{eq:approxBM1} that
    \begin{align}\label{eq:idk25}
        f(x_\lambda,y_\lambda) 
        &\leq f(x,y)+\lambda \big(-V_{\Mt(x)}^{\pi_y}(\rho)+\max_{y\in\Yc} V_{\Mt(x)}^{\pi_y}-\epsilon_\lambda\big)+\epsilon \nonumber\\
        &\leq f(x,y)+\epsilon.
    \end{align}
    This completes the proof.
\end{proof}

\subsection{Proof of Lemma \ref{lemma:gradient dominance direct param}}\label{sec:gradient dominance direct param proof}
\begin{proof}
The following proof holds for any $x$ and thus we omit $x$ in the notations $\Mt(x)$, $\pi_y^*(x)$ and $\Pc_x$ in this proof.
We first prove a policy gradient theorem for the regularized MDP. From the Bellman equation, we have
\begin{align}
    V_{\Mt}^\pi(s)=\sum_a \pi(a|s) Q_{\Mt}^\pi(s,a) - \tau h_s(\pi(s))
\end{align}
Differentiating two sides of the equation with respect to $\pi$ gives
\begin{align}\label{eq:idk213}
    \nabla V_{\Mt}^\pi(s)=\sum_a \nabla \pi(a|s) Q_{\Mt}^\pi(s,a) + \sum_a \pi(a|s) \nabla Q_{\Mt}^\pi(s,a) - \tau \nabla_{\pi} h_s(\pi(s)).
\end{align}
By the definition of $Q$ function, we have $\nabla Q_{\Mt}^\pi(s,a)=\sum_{s'}\Pc(s'|s,a) \nabla V_{\Mt}^\pi (s') $. Substituting this inequality into \eqref{eq:idk213} yields
\begin{align}
    \nabla V_{\Mt}^\pi(s)=\sum_a \nabla \pi(a|s) Q_{\Mt}^\pi(s,a) + \sum_{s'} P_\pi(s_1=s'|s_0=s) \nabla V_{\Mt}^\pi(s,a) - \tau \nabla_{\pi} h_s(\pi(s))
\end{align}
where $P^\pi(s_1=s'|s_0=s)$ is the probability of $s_1=s'$ given $s_0=s$ under policy $\pi$. Note that the above inequality has a recursive structure, thus we can repeatedly applying it to itself and obtain
\begin{align}\label{eq:pg}
    \nabla V_{\Mt}^\pi(s)= \frac{1}{1-\gamma} \E_{\Bar{s}\sim d_s^\pi}[\sum_a Q_{\Mt}^\pi(\Bar{s},a) \nabla \pi(a|\Bar{s})]+\frac{\tau}{1-\gamma} \E_{\Bar{s}\sim d_s^\pi}[-\nabla_{\pi} h_{\Bar{s}}(\pi(\Bar{s}))]
\end{align}
where $d_s^\pi(\Bar{s}) \coloneqq (1-\gamma)\sum_t \gamma^t P^\pi(s_t=\Bar{s}|s_0=s)$ is the discounted visitation distribution. 
Define $d_{\Mt}^\pi(\Bar{s}) \coloneqq \E_{s\sim\rho}[d_s^\pi(\Bar{s})]$. Since $\nabla \pi(a|\Bar{s})=1_{\Bar{s},a}$ where $1_{\Bar{s},a}$ is the indicator vector, we have the regularized policy gradient given by
    \begin{align}\label{eq:rpg direct param}
        \nabla_{\pi} V_{\Mt}^{\pi}(\rho)=\frac{1}{1-\gamma} \big[d_{\Mt}^\pi (s)\big(Q_{\Mt}^\pi(s,\cdot) - \tau \nabla h_s(\pi(s))\big)\big]_{s\in\Sc}.
    \end{align}
    Now we begin the prove the lemma. By the performance difference lemma (see e.g., \cite[Lemma 2]{lan2023policy} and \cite[Lemma 5]{zhan2023policy}), for any $\pi\in\Pi$, we have 
    \begin{align}
        \max_{\tilde{\pi}\in\Pi}V_{\Mt}^{\tilde{\pi}}(\rho)-V_{\Mt}^\pi(\rho) 
        &= \frac{1}{1-\gamma} \E_{s\sim d_{\Mt}^{\pi^*}}\Big[\ip{Q_{\Mt}^\pi(s,\cdot)}{\pi_y^*(s)-\pi(s)}-\tau h_s(\pi_y^*(s))+\tau h_s(\pi(s))\Big] \nonumber\\
        &\leq \frac{1}{1-\gamma} \E_{s\sim d_{\Mt}^{\pi^*}}\Big[\ip{Q_{\Mt}^\pi(s,\cdot)}{\pi_y^*(s)-\pi(s)}-\tau \ip{\nabla h_s(\pi(s))}{\pi_y^*(s)-\pi(s)}\Big] \nonumber
    \end{align}
    where the inequality follows from the convexity of $h_s$. Continuing from the inequality, it follows 
    \begin{align}\label{eq:idk27}
        &\max_{\tilde{\pi}\in\Pi}V_{\Mt}^{\tilde{\pi}}(\rho)-V_{\Mt}^\pi(\rho) \nonumber\\
        &\leq \frac{1}{1-\gamma} \E_{s\sim d_{\Mt}^{\pi^*}}\Big[\max_{\pi'\in\Pi}\ip{Q_{\Mt}^\pi(s,\cdot)}{\pi'(s)-\pi(s)}-\tau \ip{\nabla h_s(\pi(s))}{\pi'(s)-\pi(s)}\Big] \nonumber\\
        &= \frac{1}{1-\gamma} \E_{s\sim d_{\Mt}^{\pi}}\bigg[\frac{d_{\Mt}^{\pi^*}(s)}{d_{\Mt}^{\pi}(s)}\max_{\pi'\in\Pi}\Big(\ip{Q_{\Mt}^\pi(s,\cdot)}{\pi'(s)-\pi(s)}-\tau \ip{\nabla h_s(\pi(s))}{\pi'(s)-\pi(s)}\Big)\bigg] \nonumber\\
        &\leq \frac{1}{1-\gamma} \E_{s\sim d_{\Mt}^{\pi}}\bigg[\Big\|\frac{d_{\Mt}^{\pi^*}}{d_{\Mt}^{\pi}}\Big\|_\infty \max_{\pi'\in\Pi}\Big(\ip{Q_{\Mt}^\pi(s,\cdot)}{\pi'(s)-\pi(s)}-\tau \ip{\nabla h_s(\pi(s))}{\pi'(s)-\pi(s)}\Big)\bigg] 
    \end{align}
    where the last inequality follows from $\frac{d_{\Mt}^{\pi^*}(s)}{d_{\Mt}^{\pi}(s)} \leq \Big\|\frac{d_{\Mt}^{\pi^*}}{d_{\Mt}^{\pi}}\Big\|_\infty$ and
    \begin{align}
        &\max_{\pi'\in\Pi}\Big(\ip{Q_{\Mt}^\pi(s,\cdot)}{\pi'(s)-\pi(s)}-\tau \ip{\nabla h_s(\pi(s))}{\pi'(s)-\pi(s)}\Big) \nonumber\\
        &\geq \ip{Q_{\Mt}^\pi(s,\cdot)}{\pi(s)-\pi(s)}-\tau \ip{\nabla h_s(\pi(s))}{\pi(s)-\pi(s)}=0.
    \end{align}
    Continuing from \eqref{eq:idk27}, we have
    \begin{align}
    &\max_{\tilde{\pi}\in\Pi}V_{\Mt}^{\tilde{\pi}}(\rho)-V_{\Mt}^\pi(\rho) \nonumber\\
        &\leq \frac{1}{1-\gamma} \frac{1}{(1-\gamma)\min_s \rho(s)}\max_{\pi'\in\Pi}\E_{s\sim d_{\Mt}^{\pi}}\bigg[ \Big(\ip{Q_{\Mt}^\pi(s,\cdot)}{\pi'(s)-\pi(s)}-\tau \ip{\nabla h_s(\pi(s))}{\pi'(s)-\pi(s)}\Big)\bigg]\nonumber\\
        &=  \frac{1}{(1-\gamma)\min_s \rho(s)}\max_{\pi'\in\Pi}\ip{\nabla_{\pi} V_{\Mt}^{\pi}(\rho)}{\pi'-\pi}       
    \end{align}
     where the inequality follows from $(1-\gamma) \rho(s)\leq d_{\Mt}^{\pi}(s)\leq 1$ for any $s$ and $\pi$, and the equality follows from \eqref{eq:rpg direct param}. This proves the result.
\end{proof}

\subsection{Proof of Lemma \ref{lemma:local solution relation value p}}\label{sec:local solution relation value p proof}
\begin{proof}
    Given $x_\lambda$, point $y_\lambda$ satisfies the first-order stationary condition:
    \begin{align}
        \ip{\nabla_y f(x_\lambda,y_\lambda)+\lambda \nabla_y p(x_\lambda,y_\lambda)}{y_\lambda - y'} \leq 0,~\forall y'\in\Yc
    \end{align}
    which leads to
    \begin{align}\label{eq:311712}
        \ip{\nabla_y p(x_\lambda,y_\lambda)}{y_\lambda - y'} 
        &\leq -\frac{1}{\lambda}\ip{\nabla_y f(x_\lambda,y_\lambda)}{y_\lambda - y'}\nonumber\\
        &\leq \frac{ L\|y_\lambda - y'\|}{\lambda}\leq \frac{L C_u}{\lambda},~\forall y'\in\Yc
    \end{align}
    where $C_u\coloneqq \max_{y,y'\in\Yc}\|y-y'\|$ which is well defined by compactness of $\Yc$. For the LHS of the above inequality, we have the following inequality hold
    \begin{align}\label{eq:idk3117}
        \min_{y'\in\Yc}\ip{\nabla_y p(x_\lambda,y_\lambda)}{y_\lambda - y'}
        &= \max_{y'\in\Yc}\ip{\nabla_y V_{\Mt(x_\lambda)}^{\pi_{y_\lambda}}(\rho)}{y'-y_\lambda}\nonumber\\
        &\geq \frac{1}{(1-\gamma)\min_s \rho(s)} \big(\max_{y\in\Yc} V_{\Mt(x_\lambda)}^{\pi_y}(\rho)-V_{\Mt(x_\lambda)}^{\pi_{y_\lambda}}(\rho)\big)
    \end{align}
    where the last inequality follows from we are using direct policy parameterization $y=\pi$ and Lemma \ref{lemma:gradient dominance direct param}.

    Substituting \eqref{eq:idk3117} into \eqref{eq:311712} yields
    \begin{align}\label{eq:idk}
        \max_{y\in\Yc} V_{\Mt(x_\lambda)}^{\pi_y}(\rho)-V_{\Mt(x_\lambda)}^{\pi_{y_\lambda}}(\rho) \leq  \frac{L C_u}{\lambda}.
    \end{align}
    Define $\epsilon_\lambda \coloneqq -V_{\Mt(x_\lambda)}^{\pi_{y_\lambda}}(\rho)+\max_{y\in\Yc} V_{\Mt(x_\lambda)}^{\pi_y}(\rho)$ then $\epsilon_\lambda \leq \delta$ by choice of $\lambda$.
    
    By local optimality of $(x_\lambda,y_\lambda)$, it holds for any $x\in\Xc,y\in\Yc$ and in the neighborhood of  $(x_\lambda,y_\lambda)$ that
    \begin{align}
        f(x_\lambda,y_\lambda)+\lambda \big(-V_{\Mt(x_\lambda)}^{\pi_{y_\lambda}}(\rho)+\max_{y\in\Yc} V_{\Mt(x_\lambda)}^{\pi_{y}}\big) \leq f(x,y)+\lambda \big(-V_{\Mt(x)}^{\pi_y}(\rho)+\max_{y\in\Yc} V_{\Mt(x)}^{\pi_y}\big).
    \end{align}
    From the above inequality, it holds for any $(x,y)$ feasible for the relaxed $\mathcal{BM}$ in \eqref{eq:approxBM1} and in neighborhood of $(x_\lambda,y_\lambda)$ that
    \begin{align}
        f(x_\lambda,y_\lambda) 
        &\leq f(x,y)+\lambda \big(-V_{\Mt(x)}^{\pi_y}(\rho)+\max_{y\in\Yc} V_{\Mt(x)}^{\pi_y}-\epsilon_\lambda\big) \nonumber\\
        &\leq f(x,y)
    \end{align}
    which proves the result.
\end{proof}

\subsection{Proof of Lemma \ref{lemma:reformulation}}\label{sec:reformulation proof}
\begin{proof}
    We start with the first bullet. Define 
    $$ V_{\Mt(x)}^*(s)\coloneqq \max_{\pi\in\Pi} V_{\Mt(x)}^\pi(s),~~Q_{\Mt(x)}^*(s,a)\coloneqq r(s,a)+\gamma \E_{s'\sim\Pc_x(s,a)}[V_{\Mt(x)}^*(s')].$$
    Then it follows from the definition of the value function that for any $s_0$,
    \begin{align}
        V_{\Mt(x)}^*(s_0)
        &=\max_{\pi\in\Pi} \E\Big[r_x(s_0,a_0)-\tau h_{s_0}(\pi(s_0))+\sum_{t=1}^\infty \gamma^t \big(r_x(s_t,a_t)-\tau h_{s_t}(\pi(s_t))\big) \big| s_0,\pi\Big] \nonumber\\
        &=\max_{\pi\in\Pi} \E\Big[r_x(s_0,a_0)-\tau h_{s_0}(\pi(s_0))+\E\big[\sum_{t=1}^\infty \gamma^t \big(r_x(s_t,a_t)-\tau h_{s_t}(\pi(s_t))\big)\big| s_0,a_0,\pi,\Pc_x\big] \big| s_0,\pi\Big] \nonumber\\
        &=\max_{\pi\in\Pi} \E_{a_0\sim\pi(s_0)}\Big[r_x(s_0,a_0)-\tau h_{s_0}(\pi(s_0))+\gamma\E_{s_1\sim\Pc_x(s_0,a_0)}\big[V_{\Mt(x)}^\pi(s_1)\big]\Big] \nonumber\\
        \label{eq:idk0}
        &\leq\max_{\pi\in\Pi} \E_{a_0\sim\pi(s_0)}\Big[r(s_0,a_0)-\tau h_{s_0}(\pi(s_0))+\gamma\E_{s_1\sim\Pc_x(s_0,a_0)}\big[V_{\Mt(x)}^*(s_1)\big] \Big]
    \end{align}
    Given $x$, define a policy $\pi_y^*=(\pi_y^*(s))_{s\in\Sc}\in\Pi$ via
    $$\pi_y^*(s_0)\coloneqq\argmax_{\pi(s_0)}\E_{a_0\sim\pi(s_0)}\Big[r(s_0,a_0)-\tau h_{s_0}(\pi(s_0))+\gamma\E_{s_1\sim\Pc_x(s_0,a_0)}\big[V_{\Mt(x)}^*(s_1)\big] \Big],\forall s_0\in\Sc$$
    where the $\argmax$ is a singleton following from the $\tau$-strong convexity of $\tau h$, and we sometimes treat the singleton set as its element for convenience.
    Given the definition of $\pi_y^*$, it then follows from \eqref{eq:idk0} that 
    \begin{align}
         V_{\Mt(x)}^*(s_0)
        &\leq \E_{a_0\sim\pi_y^*(s_0)}\Big[r(s_0,a_0)-\tau h_{s_0}(\pi(s_0))+\gamma\E_{s_1\sim\Pc_x(s_0,a_0)}\big[V_{\Mt(x)}^*(s_1)\big] \Big] \nonumber\\
        &\leq  \E_{a_0\sim\pi_y^*(s_0)}\Big[r(s_0,a_0)-\tau h_{s_0}(\pi(s_0))\nonumber\\
        &+\gamma\E_{s_1\sim\Pc_x(s_0,a_0),a_1\sim\pi_y^*(s_1)}\big[r(s_1,a_1)-\tau h_{s_1}(\pi(s_1))+\gamma\E_{s_2\sim\Pc_x(s_1,a_1)}[V_{\Mt(x)}^*(s_2)]\big] \Big]
    \end{align}
    where the last inequality is a result of applying \eqref{eq:idk0} twice. Continuing to recursively apply \eqref{eq:idk0} and then using the definition of $V_{\Mt(x)}^{\pi}$ in \eqref{eq:vpi} yield
    \begin{align}
        V_{\Mt(x)}^*(s_0) \leq V_{\Mt(x)}^{\pi_y^*}(s_0),~~\forall s_0\in\Sc
    \end{align}
    which proves $\pi_y^*$ is the optimal policy for $\Mt(x)$. In addition, we have 
    \begin{align}
        \pi_y^*(s_0)
        &=\argmax_{\pi(s_0)}\E_{a_0\sim\pi(s_0)}\Big[r(s_0,a_0)-\tau h_{s_0}(\pi(s_0))+\gamma\E_{s_1\sim\Pc_x(s_0,a_0)}\big[V_{\Mt(x)}^{\pi_y^*}(s_1)\big] \Big] \nonumber\\
        &= \argmax_{\pi(s_0)}\E_{a_0\sim\pi(s_0)}\Big[Q_{\Mt(x)}^{\pi_y^*}(s_0,a_0)-\tau h_{s_0}(\pi(s_0))\Big],~~\forall s_0.
    \end{align}
    Then we have
    $\pi_y^*=\arg\min_{y\in\Pi} g(x,y)$ and thus $\arg\min_{y\in\Pi} g(x,y)\in\Yc^*(x)$. To further prove $\arg\min_{y\in\Delta(\Ac)^{|\Sc|}} g(x,y)=\Yc^*(x)$, it then suffices to prove any other policy $\pi\in\Pi$ different from $\pi_y^*$ is not optimal. Let $s_0'$ be the state such that  $\pi_y^*(s_0')\neq \pi(s_0')$. We have
    \begin{align}
         V_{\Mt(x)}^{\pi}(s_0')
         &\leq \E_{a_0\sim\pi(s_0')}\Big[r(s_0',a_0)-\tau h_{s_0'}(\pi(s_0'))+\gamma\E_{s_1\sim\Pc_x(s_0',a_0)}\big[V_{\Mt(x)}^*(s_1)\big] \Big] \nonumber\\
         &< \E_{a_0\sim\pi_y^*(s_0')}\Big[r(s_0',a_0)-\tau h_{s_0'}(\pi_y^*(s_0'))+\gamma\E_{s_1\sim\Pc_x(s_0',a_0)}\big[V_{\Mt(x)}^*(s_1)\big] \Big] \nonumber\\
         &= V_{\Mt(x)}^*(s_0')
    \end{align}
    where the last inequality follows from the strong convexity of $h$ and the definition of $\pi_y^*$; and the last equality follows from $\pi_y^*$ is the optimal policy. This proves the result.

    Next we prove the second bullet. We have
    \begin{align}\label{eq:idk1}
        \|y_\epsilon-\pi_y^*\|^2
        \leq \tau^{-1}\big(g(x,y_\epsilon)-v(x)\big) \leq \tau^{-1} \epsilon
    \end{align}
    where the first inequality follows from $\tau$-strong-convexity of $g(x,\cdot)$.
    Next we prove $|f^*-f_\epsilon^*|\leq L\tau^{-1}\epsilon.$ Let $f_\epsilon^*=f(x_\epsilon^*,y_\epsilon^*)$. We have
    \begin{align}
        f(x_\epsilon^*,\Yc(x_\epsilon^*))-f(x_\epsilon^*,y_\epsilon^*)
        &\leq L\|y_\epsilon^*-\Yc(x_\epsilon^*)\| \leq L\sqrt{\tau^{-1} \epsilon}
    \end{align}
    where the last inequality follows from \eqref{eq:idk1}. The result follows from the fact that $f(x_\epsilon^*,\Yc(x_\epsilon^*)) \geq f^*$ and $f(x_\epsilon^*,y_\epsilon^*) \leq f^*$.
\end{proof}

\section{Proof in Section \ref{sec:algorithm}}
\subsection{Proof of Lemma \ref{lemma:derivative value p}}\label{sec:derivative value p proof}
We first introduce a generalized Danskin's theorem as follows.
 \begin{lemma}[Generalized Danskin's Theorem \citep{clarke1975generalized}]\label{lemma:generalized danskin}
     Let $\Fc$ be a compact set and let a continuous function $\ell:\reals^{d}\times \Fc \mapsto \reals$ satisfy: 1) $\nabla_x \ell(x,y)$ is continuous in $(x,y)$; and 2) given any $x$, for any $y,y'\in\argmax_{y\in\Fc} \ell(x,y)$, $\nabla_x \ell(x,y)=\nabla_x \ell(x,y')$. Then let $h(x)\coloneqq \max_{y\in\Fc}\ell(x,y)$, we have $\nabla h(x)=\nabla_x \ell(x,y^*)$ for any $y^*\in \argmax_{y\in\Fc} \ell(x,y)$.
 \end{lemma}
 Lemma \ref{lemma:generalized danskin} first follows \cite[Theorem 2.1]{clarke1975generalized} where conditions (a)--(d) are guaranteed by  Lemma \ref{lemma:generalized danskin}'s condition 1). Then by \cite[Theorem 2.1 (4)]{clarke1975generalized} that we have the Clarke's generalized gradient set of $h(x)=\max_{y\in\Fc}\ell(x,y)$ is the convex hull of $\{\nabla_x \ell(x,y), y\in\argmax_{y\in\Fc}\ell(x,y)\}$.
 It then follows from Lemma \ref{lemma:generalized danskin}'s condition 2) that this generalized gradient set is a singleton $\{\nabla_x \ell(x,y^*)\}$ with any $y^*\in\argmax_{y\in\Fc}\ell(x,y)$. Finally it follows from \cite[Proposition 1.13]{clarke1975generalized} that $h(x)$ is differentiable with gradient $\nabla_x \ell(x,y^*)$.

Now to prove Lemma \ref{lemma:derivative value p}, it suffices to prove $\nabla \max_{y\in\Yc}V_{\Mt(x)}^{\pi_y}(\rho)=\nabla_x V_{\Mt(x)}^{\pi_{y^*}}(\rho)|_{y^*\in\Yc^*(x)}$. 
This arguments is true following from Assumption \ref{asp:diff value p} and the generalized Danskin's theorem above, with $\ell(x,y)=V_{\Mt(x)}^{\pi_y}(\rho)$.

\subsection{Proof of Lemma \ref{lemma:derivative value p application}}\label{sec:derivative value p application proof}
\begin{proof}
\noindent\textbf{(a).} Under the assumptions in (a), Lemma \ref{lemma:derivative value p} holds. It then follows from 
\begin{align}
    \nabla_x V_{\Mt(x)}^{\pi_y}(\rho) = \E\big[\sum_{t=0}^\infty \gamma^t \nabla r_x(s_t,a_t)|s_0\sim\rho,\pi_y\big]
\end{align}
that the result holds.

\noindent\textbf{(b).}   
Given the follower's policy $\pi_y$, define the Stackelberg MDP from the leader's view as
$$\Mc(\pi_y)=\{\Sc,\Ac_l,r_{\pi_y}(s,a_l)=\E_{a_f\sim\pi_y(s)}[r_f(s,a_f,a_l)]-\tau h_{f,s}(\pi_y(s)), \nonumber\\
\Pc_{\pi_y}(\cdot|s,a_l)=\E_{a_f\sim\pi_y(s)}[\Pc(\cdot|s,a_l,a_f)]\}$$
Note $\Mc(\pi_y)$ does not include a regularization for its policy $\pi_x$.
By Lemma \ref{lemma:stackelberg}, we have the follower's value function $V_f^{\pi_x,\pi_y}(s)$ can be rewritten from the viewpoint that $\pi_y$ is the main policy and $\pi_x$ is part of the follower's MDP, that is, $V_f^{\pi_x,\pi_y}(s)=V_{\Mt(x)}^{\pi_y}(s)$.
It can be proven similarly that $V_f^{\pi_x,\pi_y}(s)=V_{\Mc(\pi_y)}^{\pi_x}(s)$. Therefore, we have $V_{\Mt(x)}^{\pi_y}(s)=V_{\Mc(\pi_y)}^{\pi_x}(s)$ and
\begin{align}\label{eq:idk4}
    \nabla_x V_{\Mt(x)}^{\pi_y}(s)&=\nabla_x V_{\Mc(\pi_y)}^{\pi_x}(s) \nonumber\\
    &=\E\Big[\sum_{t=0}^\infty \gamma^t Q_{\Mc(\pi_y)}^{\pi_x}(s_t,a_{l,t})\nabla\log\pi_x(a_{l,t}|s_t)\big|s_0=s,\pi_x\Big]
\end{align}
where the last equality follows from the policy gradient theorem \citep{SuttonPG}. We have
\begin{align}
    Q_{\Mc(\pi_y)}^{\pi_x}(s,a_l)&=r_{\pi_y}(s,a_l)+\gamma\E_{s'\sim\Pc_{\pi_y}(s,a_l)}[V_{\Mc(\pi_y)}^{\pi_x}(s')]\nonumber\\
    &=\E_{a_f\sim\pi_y(s)}[r_f(s,a_f,a_l)]-\tau h_{f,s}(\pi_y(s))+\gamma \E_{s'\sim\Pc(s,a_l,a_f),a_f\sim\pi_y(s)}[V_f^{\pi_x,\pi_y}(s')] \nonumber\\
    &= \E_{a_f\sim\pi_y(s)}[Q_f^{\pi_x,\pi_y}(s,a_l,a_f)]-\tau h_{f,s}(\pi_y(s))
\end{align}
where the last equality follows from the definition of $Q_f^{\pi_x,\pi_y}(s,a_l,a_f)$ in Section \ref{sec:application}.
Substituting the above equality into \eqref{eq:idk4} yields
\begin{align}\label{eq:idk5}
    \nabla_x V_{\Mt(x)}^{\pi_y}(s)
    &=\E\Big[\sum_{t=0}^\infty \gamma^t Q_f^{\pi_x,\pi_y}(s_t,a_{l,t},a_{f,t})\nabla\log\pi_x(a_{l,t}|s_t)\big|s_0=s,\pi_x,\pi_y\Big] \nonumber\\
    &-\tau \E\Big[\sum_{t=0}^\infty \gamma^t h_{f,s_t}(\pi_y(s_t))\nabla\log\pi_x(a_{l,t}|s_t)\big|s_0=s,\pi_x,\pi_y\Big]
\end{align}
It then follows from Lemma \ref{lemma:derivative value p} that
\begin{align}
    \nabla_x p(x,y) 
    &= -\nabla_x V_{\Mt(x)}^{\pi_y}(\rho)+ \nabla_x V_{\Mt(x)}^{\pi_y}(\rho)|_{\pi_y=\pi_y^*(x)} \nonumber\\
    &= -\E\Big[\sum_{t=0}^\infty \gamma^t \big( Q_f^{\pi_x,\pi_y}(s_t,a_{l,t},a_{f,t})-\tau h_{f,s_t}(\pi_y(s_t)) \big)\nabla\log\pi_x(a_{l,t}|s_t)\big|s_0=s,\pi_x,\pi_y\Big] \nonumber\\
    &\quad+ \E\Big[\sum_{t=0}^\infty \gamma^t \big( Q_f^{\pi_x,\pi_y^*(x)}(s_t,a_{l,t},a_{f,t})-\tau h_{f,s_t}(\pi_y^*(x)(s_t)) \big)\nabla\log\pi_x(a_{l,t}|s_t)\big|s_0=s,\pi_x,\pi_y^*(x)\Big]
\end{align}
where $\pi_y^*(x)$ is the follower's optimal policy given leader's policy $\pi_x$.
\end{proof}

\subsection{Proof of Lemma \ref{lemma:derivative}}\label{sec:derivative proof}
\begin{proof}
We first consider $\nabla_x g(x,y)$. To prove $\nabla_x g(x,y)$ exist, it suffices to show $\nabla q_{s,a}(x)$ exist for any $(s,a)$.
    By Lemma \ref{lemma:generalized danskin}, to show $q_{s,a}(x)=-\max_{\pi\in\Pi} Q_{\Mt(x)}^\pi (s,a)$ is differentiable, it remains to show that $\argmax_{\pi\in\Pi} Q_{\Mt(x)}^\pi (s,a)$ is a singleton. 
    By Lemma \ref{lemma:reformulation}, the optimal policy of $\Mt(x)$ is unique. Since the unique optimal policy $\pi_y^*(x)\in\argmax_{\pi\in\Pi} Q_{\Mt(x)}^\pi (s,a)$, it suffices to show any policy $\pi$ different from $\pi_y^*(x)$ leads to $ Q_{\Mt(x)}^\pi (s,a) < Q_{\Mt(x)}^{\pi_y^*(x)} (s,a)$. Next we prove this result.
    
    By the uniqueness of the optimal policy, the policies different from $\pi_y^*(x)$ are non-optimal, that is, for any non-optimal $\pi$, there exists state $\Bar{s}$ such that $V_{\Mt(x)}^\pi(\Bar{s})<V_{\Mt(x)}^{\pi_y^*(x)}(\Bar{s})$.
    By the Bellman equation, we have for any $T$,
    \begin{align}
         Q_{\Mt(x)}^\pi (s,a) 
         &= \E\Big[\sum_{t=0}^{T-1} \gamma^t r_x(s_t,a_t)|\pi,s_0=s,a_0=a\Big]+\gamma^T \E_{s_T\sim P_x^\pi(\cdot|s_0=s,a_0=a)}[V_{\Mt(x)}^\pi(s_T)] 
    \end{align}
    By the irreducible Markov chain assumption, there exists $i$ such that $P_x^\pi(s_i=\Bar{s}|s_0=s,a_0=a)>0$. Choosing $T=i$ in the above equality yields
    \begin{align}\label{eq:idk2}
         Q_{\Mt(x)}^\pi (s,a) 
         &= \E\Big[\sum_{t=0}^{i-1} \gamma^t r_x(s_t,a_t)|\pi,s_0=s,a_0=a\Big]+\gamma^{i} \E_{s_i\sim P_x^\pi(\cdot|s_0=s,a_0=a)}[V_{\Mt(x)}^\pi(s_i)] \nonumber\\
         &< \E\Big[\sum_{t=0}^{i-1} \gamma^t r_x(s_t,a_t)|\pi,s_0=s,a_0=a\Big]+\gamma^{i} \E_{s_i\sim P_x^\pi(\cdot|s_0=s,a_0=a)}[V_{\Mt(x)}^{\pi_y^*(x)}(s_i)] \nonumber\\
         &\leq  Q_{\Mt(x)}^{\pi_y^*(x)} (s,a)
    \end{align}
    where the first inequality follows from $V_{\Mt(x)}^\pi(\Bar{s})<V_{\Mt(x)}^{\pi_y^*(x)}(\Bar{s})$ and $P_x^\pi(s_i=\Bar{s}|s_0=s,a_0=a)>0$; and the last inequality follows from the optimality of $\pi_y^*(x)$. 

    Given \eqref{eq:idk2}, we can conclude that $q_{s,a}(x)$ is differentiable with the gradient
    \begin{align}
        \nabla q_{s,a}(x) = -\nabla_x Q_{\Mt(x)}^\pi (s,a)|_{\pi=\pi_y^*(x)}.
    \end{align}
    Then $\nabla_x g(x,y)$ can be computed as
    \begin{align}
        \nabla_x g(x,y)=-\E_{s\sim\rho,a\sim\pi_y(s)}\big[\nabla_x Q_{\Mt(x)}^\pi(s,a)\big]\big|_{\pi=\pi_y^*(x)}.
    \end{align}
    Since $g(x,\cdot)$ is smooth and strongly-convex, we can use the Danskins' theorem to obtain
    \begin{align}
        \nabla v(x)=\nabla_x g(x,y)|_{y=\argmin_{y\in\Yc}g(x,y)}=\nabla_x g(x,y)|_{y=\pi_y^*(x)}
    \end{align}
    where the last equality follows from Lemma \ref{lemma:reformulation}. This completes the proof.
\end{proof}

\subsection{Proof of Lemma \ref{lemma:derivative application}}\label{sec:derivative application proof}
\begin{proof}
    We first prove the first bullet. We have
    \begin{align}\label{eq:idk3}
        \nabla_x Q_{\Mt(x)}^\pi(s,a) = \E\Big[\sum_{t=0}^\infty \gamma^t \nabla_x r_x(s_t,a_t)|\pi,s_0=s,a_0=a \Big].
    \end{align}
    It can be checked that Assumption \ref{asp:diff} holds and then $\nabla_x g(x,y)$, $\nabla v(x)$ follow from Lemma \ref{lemma:derivative} with \eqref{eq:idk3}.

    We next prove the second bullet.
    By \eqref{eq:idk5}, we have
    \begin{align}
        \nabla_x V_{\Mt(x)}^{\pi_y}(s)
        &=\E\Big[\sum_{t=0}^\infty \gamma^t Q_f^{\pi_x,\pi_y}(s_t,a_{l,t},a_{f,t})\nabla\log\pi_x(a_{l,t}|s_t)\big|s_0=s,\pi_x,\pi_y\Big] \nonumber\\
        &-\tau \E\Big[\sum_{t=0}^\infty \gamma^t h_{f,s_t}(\pi_y(s_t))\nabla\log\pi_x(a_{l,t}|s_t)\big|s_0=s,\pi_x,\pi_y\Big]
    \end{align}
    Then
    \begin{align}
        \nabla_x Q_{\Mt(x)}^{\pi_y}(s,a_f) &= \nabla_x \big(r_x(s,a_f)+\gamma\E_{s'\sim\Pc_x(s,a_f)}[V_{\Mt(x)}^{\pi_y}(s')\big) \nonumber\\
        &= \nabla_x \big(\E_{a_l\sim\pi_x(s)}[r_l(s,a_l,a_f)]+\gamma\E_{a_l\sim\pi_x(s),s'\sim\Pc(s,a_l,a_f)}[V_{\Mt(x)}^{\pi_y}(s')]\big)
    \end{align}
    where the last eqaulity follows from the definition of $\Mt(x)$ in Lemma \ref{lemma:stackelberg}. Using the log-trick, we can write
    \begin{align}
        \nabla_x Q_{\Mt(x)}^{\pi_y}(s,a_f) &= \E_{a_l\sim\pi_x(s)}\Big[ \big(r(s,a_l,a_f)+\gamma \E_{s'\sim\Pc(s,a_l,a_f)}[V_f^{\pi_x,\pi_y}(s')]\big)\nabla\log\pi_x(a_l|s)\Big]\nonumber\\
        &+ \gamma\E_{a_l\sim\pi_x(s),s'\sim\Pc(s,a_l,a_f)}[\nabla_x V_{\Mt(x)}^{\pi_y}(s')]
    \end{align}
    Substituting \eqref{eq:idk5} into the above equality yields
    \begin{align}\label{eq:idk6}
         \nabla_x Q_{\Mt(x)}^{\pi_y}(s,a_f) &= \E_{a_l\sim\pi_x(s)}\Big[ \big(r(s,a_l,a_f)+\gamma \E_{s'\sim\Pc(s,a_l,a_f)}[V_f^{\pi_x,\pi_y}(s')]\big)\nabla\log\pi_x(a_l|s)\Big]\nonumber\\
        &\quad+ \gamma\E_{a_l\sim\pi_x(s),s'\sim\Pc(s,a_l,a_f)}\E\Big[\sum_{t=0}^\infty \gamma^t Q_f^{\pi_x,\pi_y}(s_t,a_{l,t},a_{f,t})\nabla\log\pi_x(a_{l,t}|s_t)\big|s_0=s',\pi_x,\pi_y\Big] \nonumber\\
        &\quad- \tau\gamma\E_{a_l\sim\pi_x(s),s'\sim\Pc(s,a_l,a_f)} \E\Big[\sum_{t=0}^\infty \gamma^t h_{f,s_t}(\pi_y(s_t))\nabla\log\pi_x(a_{l,t}|s_t)\big|s_0=s',\pi_x,\pi_y\Big] \nonumber
     \end{align}
     Using the definition of $Q_f^{\pi_x,\pi_y}$ in the first term, and
     taking $\gamma$ of the second and third term inside the expectation gives
     \begin{align}
     \nabla_x Q_{\Mt(x)}^{\pi_y}(s,a_f)
        &= \E_{a_l\sim\pi_x(s)}\big[Q_f^{\pi_x,\pi_y}(s,a_l,a_f)\nabla\log\pi_x(a_l|s)\big]\nonumber\\
        &\quad+ \E_{a_l\sim\pi_x(s),s'\sim\Pc(s,a_l,a_f)}\E\Big[\sum_{t=1}^\infty \gamma^t Q_f^{\pi_x,\pi_y}(s_t,a_{l,t},a_{f,t})\nabla\log\pi_x(a_{l,t}|s_t)\big|s_1=s',\pi_x,\pi_y\Big] \nonumber\\
        &\quad-  \tau\E_{a_l\sim\pi_x(s),s'\sim\Pc(s,a_l,a_f)} \E\Big[\sum_{t=1}^\infty \gamma^t h_{f,s_t}(\pi_y(s_t))\nabla\log\pi_x(a_{l,t}|s_t)\big|s_1=s',\pi_x,\pi_y\Big] \nonumber
    \end{align}
    Continuing from above, combining the first and second term yields
    \begin{align}
    \nabla_x Q_{\Mt(x)}^{\pi_y}(s,a_f)
        &=  \E_{a_l\sim\pi_x(s)}\E\Big[\sum_{t=0}^\infty \gamma^t Q_f^{\pi_x,\pi_y}(s_t,a_{l,t},a_{f,t})\nabla\log\pi_x(a_{l,t}|s_t)\big|s_0=s,a_{l,0}=a_l,a_{f,0}=a_f,\pi_x,\pi_y\Big] \nonumber\\
        &\quad- \tau\E_{a_l\sim\pi_x(s),s'\sim\Pc(s,a_l,a_f)} \E\Big[\sum_{t=1}^\infty \gamma^t h_{f,s_t}(\pi_y(s_t))\nabla\log\pi_x(a_{l,t}|s_t)\big|s_1=s',\pi_x,\pi_y\Big] \nonumber\\
        &=  \E\Big[\sum_{t=0}^\infty \gamma^t Q_f^{\pi_x,\pi_y}(s_t,a_{l,t},a_{f,t})\nabla\log\pi_x(a_{l,t}|s_t)\big|s_0=s,a_{f,0}=a_f,\pi_x,\pi_y\Big] \nonumber\\
        &\quad- \tau\E_{a_l\sim\pi_x(s),s'\sim\Pc(s,a_l,a_f)} \E\Big[\sum_{t=1}^\infty \gamma^t h_{f,s_t}(\pi_y(s_t))\nabla\log\pi_x(a_{l,t}|s_t)\big|s_1=s',\pi_x,\pi_y\Big]
    \end{align}
    It can then be checked that Assumption \ref{asp:diff} holds and the result follows from Lemma \ref{lemma:derivative} and \eqref{eq:idk6}.
\end{proof}

\subsection{Sufficient conditions of the smoothness assumption}\label{sec:justification of smooth value}
\begin{lemma}\label{lemma:justification of smooth value}
Suppose the following conditions hold.
\begin{enumerate}[label=(\alph*)]
    \item For any $(s,a)$, the policy parameterization $\pi_y$ satisfies 1) $\sum_a\| \nabla\pi_y(a|s)\| \leq B_\pi$; and, 2) $\pi_y(a|s)$ is $L_y$-Lipschitz-smooth.
    \item If $\tau>0$ then: 1) for any $s$, assume $|h_s(\pi_y(s))|\leq B_h$ and $\|\nabla_y h_s(\pi_y(s))\|\leq B_h'$ on $\Yc$; and, 2)  $h_s(\pi_y(s))$ is $L_h$-Lipschitz-smooth on $\Yc$.
    \item For any $(s,a,s')$, we have for any $x\in\Xc$ that 1) $|r_x(s,a)|\leq B_r$; and, 2) $V_{\Mt(x)}^{\pi_y}(\rho)$ is $L_{vx}$-Lipschitz-smooth on $\Xc$ uniformly for $y\in\Yc$.
\end{enumerate}
Then it holds for any $s$ that $V_{\Mt(x)}^{\pi_y}(s)$ is Lipschitz-smooth on $\Xc\times\Yc$:
\begin{align}
    \|\nabla V_{\Mt(x)}^{\pi_y}(s)-\nabla V_{\Mt(x)}^{\pi_{y'}}(s)\| \leq \max\{L_{vx},L_{vy}\}\|(x,y)-(x',y')\|,~~\forall x,x'\in\Xc\text{ and }y,y'\in\Yc
\end{align}
where $L_{vy}=\mathcal{O}\Big(\frac{B_\pi^2(B_r+\tau B_h)}{(1-\gamma)^3}+\frac{\tau B_h'B_\pi + |\Ac|L_y(B_r+\tau B_h)}{(1-\gamma)^2}+\frac{\tau( B_h'+L_h)}{1-\gamma}\Big)$.
\end{lemma}
Condition (a) holds for direct parameterization, where $\sum_a \|\nabla \pi_y(a|s)\| \leq |\Ac|$ and $L_y=0$; and it also holds for softmax parameterization where $\sum_a \|\nabla \pi_y(a|s)\|=\sum_a \pi_y(a|s)\|\nabla \log\pi_y(a|s)\| \leq 1$ and $L_y=2$. Condition (b) holds for smooth composite of regularization function and policy, e.g., softmax and entropy \citep[Lemma 14]{mei2020softmax}, or direct policy with a smooth regularization. function. Condition (c) 1) is guaranteed since $\Xc$ is compact and $r_x$ is continuous, and 2) needs to be checked for specific applications. For example, in RLHF/Reward shaping, it can be checked from the formula of $\nabla_x V_{\Mt(x)}^{\pi_y}(s)$ in Lemma \ref{lemma:derivative value p application} that there exists $L_{vx}=\frac{L_r}{1-\gamma}$ if $r_x$ is $L_r$-Lipschitz-smooth.
\begin{proof}
We start the proof by showing $V_{\Mt(x)}^{\pi_y}(s)$ is Lipschitz-smooth in $y$ on uniformly for any $x$, that is 
\begin{align}\label{eq:idk goal1}
    \| \nabla_y V_{\Mt(x)}^{\pi_y}(s)- \nabla_y V_{\Mt(x)}^{\pi_{y'}}(s)\| \leq L_{vy}\|y-y'\|
\end{align}
where $L_{vy}$ is a constant independent of $x$.
By the regularized policy gradient in \eqref{eq:pg}, we have
\begin{align}\label{eq:pg in this proof}
    \nabla_y V_{\Mt(x)}^{\pi_y}(s)= \frac{1}{1-\gamma} \E_{\Bar{s}\sim d_{s,x}^{\pi_y}}\Big[\sum_a Q_{\Mt(x)}^{\pi_y}(\Bar{s},a) \nabla \pi_y(a|\Bar{s})\Big]+\frac{\tau}{1-\gamma} \E_{\Bar{s}\sim d_{s,x}^\pi}[-\nabla_y h_{\Bar{s}}(\pi_y(\Bar{s}))]
\end{align}
where $d_{s,x}^{\pi_y}(\Bar{s}) \coloneqq (1-\gamma)\sum_{t=0}^\infty \gamma^t P_x^{\pi_y}(s_t=\Bar{s}|s_0=s)$ is the discounted visitation distribution, and recall $P_x^{\pi_y}(s_t=\Bar{s}|s_0=s)$ is the probability of reaching state $\Bar{s}$ at time step $t$ under $\Pc_x$ and $\pi_y$. Towards proving \eqref{eq:idk goal1}, we prove the following results:

\noindent\textbf{(1)} We have $Q_{\Mt(x)}^{\pi_y}(s,a)$ is uniformly bounded, and $V_{\Mt(x)}^{\pi_y}(s)$ and $Q_{\Mt(x)}^{\pi_y}(s,a)$ are Lipschitz continuous in $y$ uniformly for any $x$. 

By the definition of $Q_{\Mt(x)}^{\pi_y}(s,a)$, we have
\begin{align}\label{eq:q upperbound}
    |Q_{\Mt(x)}^{\pi_y}(s,a)|\leq \sum_{t=0}^\infty \gamma^t |r_x(s_t,a_t)|+\tau|h_{s_t}(\pi_y(s_t))| \leq \frac{B_r+\tau B_h}{1-\gamma},
\end{align}
therefore it follows from \eqref{eq:pg in this proof} that
\begin{align}\label{eq:v lipschitz}
    \|\nabla_y V_{\Mt(x)}^{\pi_y}(s)\| \leq B_\pi\frac{B_r+\tau B_h}{(1-\gamma)^2}+\frac{\tau B_h'}{1-\gamma}
\end{align}
Then by the definition of Q function
\begin{align}
    Q_{\Mt(x)}^{\pi_y}(s,a) = r_x(s,a)+\gamma \E_{s'\sim\Pc_x(s,a)}\big[V_{\Mt(x)}^{\pi_y}(s')\big] \nonumber
\end{align}
we have
\begin{align}\label{eq:q lipschitz}
    \|\nabla_y Q_{\Mt(x)}^{\pi_y}(s,a)\| \leq  \gamma \big(B_\pi\frac{B_r+\tau B_h}{(1-\gamma)^2}+\frac{\tau B_h'}{1-\gamma}\big)
\end{align}

\noindent\textbf{(2)} We have $d_{s_0,x}^{\pi_y}(s)$ is Lipschitz-continuous in $y$ uniformly for any $x$. 
Define $\Mc$ as a MDP with $\tau=0$, $r(s,a)=\textbf{1}_s$ which is an indicator function of $s$, and transition $\Pc_x$. 
Then we can write $d_{s_0,x}^{\pi_y}(s)$ as
\begin{align}
    d_{s_0,x}^{\pi_y}(s) &= \sum_{s'\in\Sc} \sum_{a'\in\Ac} d_{s_0,x}^{\pi_y}(s') \pi_y(a'|s') \textbf{1}_s\nonumber\\
    &=\E_{s\sim d_{s_0,x}^{\pi_y},a\sim\pi_y(s)}[r(s,a)] \nonumber\\
    &= (1-\gamma)V_{\Mc}^{\pi_y}(s_0) \nonumber
\end{align}
where the last equality follows from substituting in $d_{s_0,x}^{\pi_y}(s)=(1-\gamma)\sum_{t=0}^\infty \gamma^t P_x^{\pi_y}(s_t=s|s_0)$. It then follows from \eqref{eq:v lipschitz} with $\tau=0$ (since $V_{\Mc}^{\pi_y}(s_0)$ has $\tau=0$) that $d_{s_0,x}^{\pi_y}(s)$ is also uniformly Lipschitz continuous with constant $B_\pi$:
\begin{align}\label{eq:d lipschitz}
    \sup_{s\in\Sc}\|d_{s_0,x}^{\pi_y}(s)-d_{s_0,x}^{\pi_{y'}}(s)\| \leq B_\pi \|y-y'\|.
\end{align}

To this end, we can decompose the difference as
\begin{align}
    &\nabla_y V_{\Mt(x)}^{\pi_y}(s)-\nabla_y V_{\Mt(x)}^{\pi_{y'}}(s)\nonumber\\
    &=\frac{1}{1-\gamma} \E_{\Bar{s}\sim d_{s,x}^{\pi_y}}\Big[\sum_a Q_{\Mt(x)}^{\pi_y}(\Bar{s},a) \nabla \pi_y(a|\Bar{s})\Big]-\frac{1}{1-\gamma} \E_{\Bar{s}\sim d_{s,x}^{\pi_{y'}}}\Big[\sum_a Q_{\Mt(x)}^{\pi_y}(\Bar{s},a) \nabla \pi_y(a|\Bar{s})\Big] \nonumber\\
    &+\frac{1}{1-\gamma} \E_{\Bar{s}\sim d_{s,x}^{\pi_{y'}}}\Big[\sum_a Q_{\Mt(x)}^{\pi_y}(\Bar{s},a) \nabla \pi_y(a|\Bar{s})\Big]-\frac{1}{1-\gamma} \E_{\Bar{s}\sim d_{s,x}^{\pi_{y'}}}\Big[\sum_a Q_{\Mt(x)}^{\pi_{y'}}(\Bar{s},a) \nabla \pi_y(a|\Bar{s})\Big] \nonumber\\
    &+\frac{1}{1-\gamma} \E_{\Bar{s}\sim d_{s,x}^{\pi_{y'}}}\Big[\sum_a Q_{\Mt(x)}^{\pi_{y'}}(\Bar{s},a) \nabla \pi_y(a|\Bar{s})\Big] -\frac{1}{1-\gamma} \E_{\Bar{s}\sim d_{s,x}^{\pi_{y'}}}\Big[\sum_a Q_{\Mt(x)}^{\pi_{y'}}(\Bar{s},a) \nabla \pi_{y'}(a|\Bar{s})\Big] \nonumber\\
    &+\frac{\tau}{1-\gamma} \E_{\Bar{s}\sim d_{s,x}^{\pi_y}}[-\nabla_y h_{\Bar{s}}(\pi_y(\Bar{s}))]-\frac{\tau}{1-\gamma} \E_{\Bar{s}\sim d_{s,x}^{\pi_{y'}}}[-\nabla_y h_{\Bar{s}}(\pi_y(\Bar{s}))] \nonumber\\
    &+\frac{\tau}{1-\gamma} \E_{\Bar{s}\sim d_{s,x}^{\pi_{y'}}}[-\nabla_y h_{\Bar{s}}(\pi_y(\Bar{s}))]-\frac{\tau}{1-\gamma} \E_{\Bar{s}\sim d_{s,x}^{\pi_{y'}}}[-\nabla_y h_{\Bar{s}}(\pi_{y'}(\Bar{s}))] \nonumber
\end{align}
Continuing from the above inequality, we have
\begin{align}
    \|\nabla_y V_{\Mt(x)}^{\pi_y}(s)-\nabla_y V_{\Mt(x)}^{\pi_{y'}}(s)\|
    &\leq\frac{1}{1-\gamma} 2\sup_{s}\| d_{s,x}^{\pi_y}(s)-d_{s,x}^{\pi_{y'}}(s)\|\sup\big\|\sum_a Q_{\Mt(x)}^{\pi_y}(\Bar{s},a) \nabla \pi_y(a|\Bar{s})\big\|\nonumber\\
    &+\frac{1}{1-\gamma} \sup_a \big|Q_{\Mt(x)}^{\pi_y}(\Bar{s},a)-Q_{\Mt(x)}^{\pi_{y'}}(\Bar{s},a)\big|\sum_a\big\|  \nabla \pi_y(a|\Bar{s})\big\|\nonumber\\
    &+\frac{1}{1-\gamma} \E_{\Bar{s}\sim d_{s,x}^{\pi_{y'}}}\Big[\sup_a \big|Q_{\Mt(x)}^{\pi_{y'}}(\Bar{s},a)\big|\sum_a \big\| \nabla \pi_y(a|\Bar{s})-\nabla \pi_{y'}(a|\Bar{s})\big\|\Big]\nonumber\\
    &+\frac{\tau}{1-\gamma} 2\sup_{s}\| d_{s,x}^{\pi_y}(s)-d_{s,x}^{\pi_{y'}}(s)\|\sup\|\nabla_y h_{\Bar{s}}(\pi_y(\Bar{s}))\| \nonumber\\
    &+\frac{\tau}{1-\gamma} \E_{\Bar{s}\sim d_{s,x}^{\pi_{y'}}}\big[\|\nabla_y h_{\Bar{s}}(\pi_y(\Bar{s}))-\nabla_y h_{\Bar{s}}(\pi_{y'}(\Bar{s}))\|\big].
\end{align}
Then given the assumptions (a), (b) in this lemma, along with the \eqref{eq:q upperbound}--\eqref{eq:d lipschitz}, we can get
\begin{align}\label{eq:v smooth in y}
    \|\nabla_y V_{\Mt(x)}^{\pi_y}(s)-\nabla_y V_{\Mt(x)}^{\pi_{y'}}(s)\| \leq L_{vy} \|y-y'\|
\end{align}
where $L_{vy}=\mathcal{O}\Big(\frac{B_\pi^2(B_r+\tau B_h)}{(1-\gamma)^3}+\frac{\tau B_h'B_\pi + |\Ac|L_y(B_r+\tau B_h)}{(1-\gamma)^2}+\frac{\tau( B_h'+L_h)}{1-\gamma}\Big)$.
Thus we conclude
\begin{align}
     &\|\nabla V_{\Mt(x)}^{\pi_y}(s)-\nabla V_{\Mt(x)}^{\pi_{y'}}(s)\|^2 \nonumber\\
     &= \|\nabla_y V_{\Mt(x)}^{\pi_y}(s)-\nabla_y V_{\Mt(x)}^{\pi_{y'}}(s)\|^2+\|\nabla_x V_{\Mt(x)}^{\pi_{y'}}(s)-\nabla_x V_{\Mt(x')}^{\pi_{y'}}(s)\|^2\nonumber\\
     &\leq L_{vy}^2\|y-y'\|^2+L_{vx}^2\|x-x'\|^2 \leq \max\{L_{vy}^2,L_{vx}^2\}\|(x,y)-(x',y')\|^2
\end{align}
which proves the result.
\end{proof}

\subsection{Proof of Lemma \ref{lemma:value smooth}}\label{sec:smooth proof}
\subsubsection{Smoothness of the value penalty}
\begin{proof}
Under the two assumptions, Lemma \ref{lemma:lipschitz continuous optimal policy} holds and thus $\pi_y^*(x)$ is unique and is $\tau^{-1}C_J$-Lipschitz continuous on $\Xc$.
 Thus for any $y,y'\in\Yc^*(x)$, we have $\pi_y=\pi_{y'}=\pi_y^*(x)$. With Lemma \ref{lemma:derivative value p}, we have
\begin{align}
    \|\nabla\max_{y\in\Yc} V_{\Mt(x)}^{\pi_y}(\rho) - \nabla\max_{y\in\Yc} V_{\Mt(x')}^{\pi_y}(\rho)\|
    &=\|\nabla V_{\Mt(x)}^{\pi}(\rho)|_{\pi=\pi_y^*(x)}-\nabla V_{\Mt(x')}^{\pi}(\rho)|_{\pi=\pi_y^*(x')}\|\nonumber\\
    &\leq L_v(\|x-x'\|+\|\pi_y^*(x)-\pi_y^*(x')\|)\nonumber\\
    &\leq L_v(1+\tau^{-1}C_J)\|x-x'\|.
\end{align}
It then follows from $V_{\Mt(x)}^{\pi_y}(\rho)$ is $L_v$-Lipschitz smooth that the value penalty is $L_v(2+\tau^{-1}C_J)$-Lipschitz smooth.
\end{proof}
\subsubsection{Smoothness of the Bellman penalty}
\begin{proof}
First note that Lemma \ref{lemma:lipschitz continuous optimal policy} holds and thus $\pi_y^*(x)$ is $\tau^{-1}C_J$-Lipschitz continuous on $\Xc$. 
 We have $p(x,y)=g(x,y)-v(x)$ where
\begin{align}
    g(x,y)\coloneqq \E_{s\sim\rho}[\ip{y_s}{q_s(x)}+\tau h_s(y_s)].
\end{align}
By Lemma \ref{lemma:derivative},
\begin{align}
        \nabla_x g(x,y)=-\E_{s\sim\rho,a\sim y_s}\big[\nabla_x Q_{\Mt(x)}^\pi(s,a)\big]\big|_{\pi=\pi_y^*(x)}.
\end{align}
Since $\nabla_x Q_{\Mt(x)}^\pi(s,a)$ is $L_v$-Lipschitz continuous by the assumption, and $\pi_y^*(x)$ is $\tau^{-1}C_J$-Lipschitz continuous, we have $\nabla_x g(x,y)$ is $L_v(1+\tau^{-1}C_J)$-Lipschitz continuous at $x\in\Xc$ uniformly for any $y$. We also have $\nabla_x g(x,y)$ is $C_J$-Lipschitz continuous at $y\in\Pi$ uniformly for any $x\in\Xc$. Therefore, we conclude $\nabla_x g(x,y)$ is $(C_J+L_v(1+\tau^{-1}C_J))$-Lipschitz continuous at $(x,y)$ on $\Xc\times\Pi$.

Next we have 
\begin{align}
        \nabla_y g(x,y)=\Big(\rho(s)q_s(x)+\tau\rho(s)\nabla h_s(y_s)\Big)_{s\in\Sc}.
\end{align}
Since $q_s$ is $C_J$-Lipschitz continuous, and $h_s$ is $L_h$-Lipschitz smooth, we have $\nabla_y g(x,y)$ is $(C_J+L_h)$-Lipschitz continuous at $(x,y)$ on $\Xc\times\Pi$.

Collecting the Lipschitz continuity of $\nabla_x g(x,y)$ and $\nabla_y g(x,y)$ yields $g(x,y)$ is Lipschitz smooth with modulus $L_g=2 C_J+L_v(1+\tau^{-1}C_J)+L_h$. Then we have
\begin{align}
    \|v(x)-v(x')\|=\|g(x,\pi_y^*(x))-g(x',\pi_y^*(x'))\| \leq L_g (\|x-x'\|+\tau^{-1}C_J\|x-x'\|).
\end{align}
Then we have $p(x,y)=g(x,y)-v(x)$ is Lipschitz smooth with modulus $L_g(2+\tau^{-1}C_J)$. Together with the assumption that $f$ is $L_f$-Lipschitz smooth gives $F_\lambda$ is $L_v$-Lipschitz smooth with $L_v=L_f+\lambda L_g(2+\tau^{-1}C_J)$.
\end{proof}

\subsection{Example gradient estimators of the penalty functions}\label{sec:example gradient estimators}
In this section, we give examples of $\hat{\nabla}p(x,y;\hat{\pi})$ that is an estimator of $\nabla p(x,y)$.

\noindent\textbf{Value penalty.} Consider choosing the value penalty $p(x,y)=- V_{\Mt(x)}^{\pi_y}(\rho)+  \max_{y\in\Yc}V_{\Mt(x)}^{\pi_y}(\rho)$. Then by Lemma \ref{lemma:derivative value p}, we have
\begin{align}
     \nabla_x p(x,y)=-\nabla_x V_{\Mt(x)}^{\pi_y}(\rho)+ \nabla_x V_{\Mt(x)}^{\pi}(\rho)|_{\pi=\pi_y^*(x)}\nonumber
\end{align}
where recall $\pi_y^*(x)$ is the optimal policy of MDP $\Mt(x)$ on the policy class $\Pi=\{\pi_y:y\in\Yc\}$. A natural choice of $\hat{\nabla}p(x,y;\hat{\pi})$ is then
\begin{align}
    \hat{\nabla}p(x,y;\hat{\pi}) \coloneqq \Big(-\nabla_x V_{\Mt(x)}^{\pi_y}(\rho)+ \nabla_x V_{\Mt(x)}^{\pi}(\rho)|_{\pi=\hat{\pi}},\nabla_y p(x,y) \Big)
\end{align}
By \citep[Lemma D.3.]{agarwal2019optimality}, there exists constant $L_v=2\gamma|\Ac|/(1-\gamma)^3$ that $V^\pi_{\Mt(x)}(\rho)$ is $L_v$-Lipschitz-smooth in $\pi$ for any $x$. Then the estimation error can be quantified by
\begin{align}\label{eq:idk36}
    \|\hat{\nabla}p(x,y;\hat{\pi})-\nabla p(x,y)\| \leq L_v \|\pi_y^*(x)-\hat{\pi}\|.
\end{align}
Therefore, the estimation error is upper bounded by the policy optimality gap $\|\pi_y^*(x)-\hat{\pi}\|$. One may use efficient algorithms (e.g., policy mirror descent \citep{zhan2023policy}) to solve for $\hat{\pi}$, which has an iteration complexity of $\mathcal{O}(\log (1/\epsilon))$ to achieve $\|\pi_y^*(x)-\hat{\pi}\| \leq \epsilon$. Then Assumption \ref{asp:orcale accuracy} is guaranteed with complexity $\mathcal{O}(\log( \lambda^2/\epsilon_{orac}))$.

\noindent\textbf{Bellman penalty.} Consider choosing the Bellman penalty $p(x,y)=g(x,y)-v(x)$ where recall $g(x,y)=\E_{s\sim\rho}[\ip{y_s}{q_s(x)}+\tau h_s(y_s)]$ and $v(x)=\min_{y\in\Yc} g(x,y)$. Then by Lemma \ref{lemma:derivative}, we have
\begin{align}
     \nabla_x p(x,y)&=-\E_{s\sim\rho,a\sim\pi_y(s)}\big[\nabla_x Q_{\Mt(x)}^\pi(s,a)\big]\big|_{\pi=\pi_y^*(x)}\nonumber\\
     &\quad+\E_{s\sim\rho,a\sim\pi(s)}\big[\nabla_x Q_{\Mt(x)}^\pi(s,a)\big]\big|_{\pi=\pi_y^*(x)}
\end{align}
Therefore, a natural choice of $\hat{\nabla}p(x,y;\hat{\pi})$ is then
\begin{align}
    \hat{\nabla}p(x,y;\hat{\pi}) \coloneqq \Big(-\E_{s\sim\rho,a\sim\pi_y(s)}\big[\nabla_x Q_{\Mt(x)}^{\hat{\pi}}(s,a)\big]+\E_{s\sim\rho,a\sim\hat{\pi}(s)}\big[\nabla_x Q_{\Mt(x)}^{\hat{\pi}}(s,a)\big],\nabla_y p(x,y) \Big)
\end{align}
It then follows similarly to \eqref{eq:idk36} that Assumption \ref{asp:orcale accuracy} is guaranteed with complexity $\mathcal{O}(-\log( \epsilon_{orac}/\lambda^2))$.

\noindent\textbf{Example algorithms to get $\hat{\pi}$.}
Finally, we also explicitly write down the update to obtain $\hat{\pi}$ to be self-contained. If we are using policy mirror descent, then at each outer-iteration $k$, for $i=1,...T$ where $T$ is the inner iteration number, we run
\begin{align}
    \pi_k^{i+1}(\cdot|s) = \argmin_{p \in \Pi}\Big\{-\ip{p}{Q_{\Mt(x)}^{\pi_k^i}(s,\cdot)}+\tau  h_s(p) +\frac{1}{\eta}D_h(p,\pi_k^i;\xi_k^i)\Big\},~\text{for any }s\in\Sc
\end{align}
where $\eta$ is a learning rate, $D_h$ is the Bregman divergence, and $\xi_k^i$ is given by
\begin{align}
    \xi_k^{i+1}(s,a) = \frac{1}{1+\eta\tau}\xi_k^i(s,a)+\frac{\eta}{1+\eta \tau}Q_{\Mt(x)}^{\pi_k^i}(s,a).
\end{align}
Finally, we set the last iterate $\pi_k^{T+1}(\cdot|s)$ as the approximate optimal policy $\hat{\pi}_k$.
For theoretical reasons, we use this update in the analysis to gain fast rate. While practically our update scheme is not limited to policy mirror descent. As a simple example, the policy gradient based algorithms can also be used:
\begin{align}
    \hat{y}_k^{i+1} = \Proj_{\Yc}\Big[\hat{y}_k^{i} + \eta \nabla_{\hat{y}}V_{\Mt(x)}^{\pi_{\hat{y}_k^i}}(\rho)\Big],~\text{for }i=1,2,\dots,T.
\end{align}
We use the last iterate as the approximate optimal policy parameter: $\hat{\pi}_k=\pi_{\hat{y}_k^{T+1}}$.
In the above update, the policy gradient $\nabla_{\hat{y}}V_{\Mt(x)}^{\pi_{\hat{y}_k^i}}(\rho)$ can be estimated by a wide range of algorithms including the basic Reinforce \citep{baxter2001infinite}, and the advantage actor-critic \citep{A3C}.

\subsection{Proof of Theorem \ref{theorem:PBRL convergence}}\label{sec:PBRL convergence proof}
\begin{proof}
 In this proof, we write $z=(x,y)$.
Consider choosing either the value penalty or the Bellman penalty, then Lemma \ref{lemma:value smooth} holds under the assumptions of this theorem. Therefore, $F_\lambda$ is $L_\lambda$-Lipschitz-smooth with $L_\lambda = L_f + \lambda L_p$.
Then by Lipschitz-smoothness of $F_\lambda$, it holds that
\begin{align}\label{eq:00}
    F_\lambda(z_{k+1}) 
    &\leq F_\lambda(z_k) + \ip{\gr F_\lambda(z_k)}{z_{k+1}-z_k} + \frac{L_\lambda}{2}\|z_{k+1}-z_k\|^2 \nonumber\\
    &\stackrel{\alpha\leq \frac{1}{L_\lambda}}{\leq} F_\lambda(z_k) + \ip{\hat{\gr}F_\lambda (z_k;\hat{\pi}_k)}{z_{k+1}-z_k} + \frac{1}{2\alpha}\|z_{k+1}-z_k\|^2 +\ip{\gr F_\lambda(z_k)-\hat{\gr}F_\lambda (z_k;\hat{\pi}_k)}{z_{k+1}-z_k}.
\end{align}
Consider the second term in the RHS of \eqref{eq:00}.
It is known that $z_{k+1}$ can be written as
$$z_{k+1}=\arg\min_{z\in\Zc}\ip{\hat{\gr}F_\lambda (z_k;\hat{\pi}_k)}{z}+\frac{1}{2\alpha}\|z-z_k\|^2.$$
By the first-order optimality condition of the above problem, it holds that
\begin{align}
    \ip{\hat{\gr}F_\lambda (z_k;\hat{\pi}_k)+\frac{1}{\alpha}(z_{k+1}-z_k)}{z_{k+1}-z} \leq 0,~\forall z \in \Zc.\nonumber
\end{align}
Since $z_k\in\Zc$, we can choose $z=z_k$ in the above inequality and obtain
\begin{align}\label{eq:bb00}
    \ip{\hat{\gr}F_\lambda (z_k;\hat{\pi}_k)}{z_{k+1}-z_k} \leq -\frac{1}{\alpha}\|z_{k+1}-z_k\|^2.
\end{align}
Consider the last term in the RHS of \eqref{eq:00}. By Young's inequality, we first have
\begin{align}\label{eqn.app.proof.b8}
    \ip{\gr F_\lambda(z_k)-\hat{\gr}F_\lambda (z_k;\hat{\pi}_k)}{z_{k+1}-z_k} 
    &\leq \alpha \norm{\gr F_\lambda(z_k)-\hat{\gr}F_\lambda (z_k;\hat{\pi}_k)}^2 + \frac{1}{4\alpha}\norm{z_{k+1}-z_k}^2 \nonumber\\
    &\leq \alpha \lambda^2\norm{\gr p(z_k)-\hat{\gr}p (z_k;\hat{\pi}_k)}^2 + \frac{1}{4\alpha}\norm{z_{k+1}-z_k}^2
\end{align}

Substituting \eqref{eqn.app.proof.b8} and \eqref{eq:bb00} into \eqref{eq:00} and rearranging the resulting inequality yield
\begin{align}\label{eq:trash00}
\frac{1}{4\alpha}\|z_{k+1}-z_k\|^2 \leq F_\lambda(z_k)-F_\lambda(z_{k+1}) + \alpha \lambda^2 \norm{\gr p(z_k)-\hat{\gr}p (z_k;\hat{\pi}_k)}^2.
\end{align}
With $\Bar{z}_{k+1}$ defined in \eqref{eq:nonconvex proximal grad}, we have
\begin{align}\label{eqn.app-proof.b12}
    \norm{\Bar{z}_{k+1}-z_k}^2 &\leq 2\norm{\Bar{z}_{k+1}-z_{k+1}}^2+2\|z_{k+1}-z_k\|^2 \nonumber\\
    &\leq 2\alpha^2 \norm{\gr F_\lambda(z_k)-\hat{\gr}F_\lambda (z_k;\hat{\pi}_k)}^2\!+\!2\|z_{k+1}\!-\!z_k\|^2 \nonumber\\
    &\leq 2\alpha^2 \lambda^2\norm{\gr p(z_k)-\hat{\gr}p (z_k;\hat{\pi}_k)}^2 + 2\|z_{k+1}\!-\!z_k\|^2
\end{align}
where the second inequality uses non-expansiveness of $\Proj_{\Zc}$.

Together \eqref{eq:trash00} and \eqref{eqn.app-proof.b12} imply
\begin{align}
    \norm{\Bar{z}_{k+1}-z_k}^2 \leq 10\alpha^2\lambda^2\norm{\gr p(z_k)-\hat{\gr}p (z_k;\hat{\pi}_k)}^2 + 8\alpha(F_\lambda(z_k)-F_\lambda(z_{k+1})).\nonumber
\end{align}
Since $p(x,y) \geq 0$, $F_\lambda(z) \geq \inf_{z\in\Zc} f(z)$ for any $z\in\Zc$.
Taking a telescope sum of the above inequality and using $G_\lambda(z_k)=\frac{1}{\alpha}(z_k-\Bar{z}_{k+1})$ yield
\begin{align}
    \sum_{k=1}^K \|G_\lambda(z_k)\|^2 
    &\leq \frac{8\big(F_\lambda(z_1)-\inf_{z\in\Zc} f(z)\big)}{\alpha}+\sum_{k=1}^K 10\lambda^2 \norm{\gr p(z_k)-\hat{\gr}p (z_k;\hat{\pi}_k)}^2\nonumber\\
    &\leq \frac{8\big(F_\lambda(z_1)-\inf_{z\in\Zc} f(z)\big)}{\alpha}+\sum_{k=1}^K \frac{1}{2} \|G_\lambda(z_k)\|^2 + \frac{K}{2}\epsilon_{orac}
\end{align}
where the last inequality follows from Assumption \ref{asp:orcale accuracy}.
Rearranging gives
\begin{align}\label{eq:result nonconvex}
    \sum_{k=1}^K \|G_\lambda(z_k)\|^2 \leq \frac{16\big(F_\lambda(z_1)-\inf_{z\in\Zc} f(z)\big)}{\alpha}+K\epsilon_{orac}.
\end{align}
This proves the first inequality in this theorem. The result for $\mathcal{OS}$ follows similarly with $F_\lambda (y)$ being $L_v$-Lipschitz-smooth and $\epsilon_{orac}=0$ since no oracle is needed.
\end{proof}

\section{Proof in Section \ref{sec:bilevel rl zero sum}}
\subsection{Proof of Lemma \ref{lemma:gradient dominance NI}}\label{sec:gradient dominance NI proof}
    \noindent\textbf{(a).} Treating $(\pi_2,x)$ (or $(\pi_1,x)$) as the parameter, it follows from Lemma \ref{lemma:reformulation} that $\pi_1^*$ (or $\pi_2^*$) is unique. Under Assumption \ref{asp:diff NI p}, it then follows from Lemma \ref{lemma:generalized danskin} that (a) holds.

    \noindent\noindent\textbf{(b).} We have
    \begin{align}
        \max_{\pi'\in\Pi}\ip{\nabla_\pi \psi(x,\pi)}{\pi-\pi'}
        &= \max_{\pi'\in\Pi}\ip{\nabla_{\pi_2} V_{\Mt(x)}^{\pi_1^*,\pi_2}(\rho)}{\pi_2-\pi_2'}+\ip{\nabla_{\pi_1}-V_{\Mt(x)}^{\pi_1,\pi_2^*}(\rho)}{\pi_1-\pi_1'}\nonumber\\
        &= \max_{\pi_2'\in\Pi_2}\ip{\nabla_{\pi_2} -V_{\Mt(x)}^{\pi_1^*,\pi_2}(\rho)}{\pi_2'-\pi_2}+\max_{\pi_1'\in\Pi_1}\ip{\nabla_{\pi_1}V_{\Mt(x)}^{\pi_1,\pi_2^*}(\rho)}{\pi_1'-\pi_1}\nonumber\\
        &\geq \mu \Big[\max_{\pi_2\in\Pi_2}\big(-V_{\Mt(x)}^{\pi_1^*,\pi_2}(\rho)\big)+V_{\Mt(x)}^{\pi_1^*,\pi_2}(\rho) +\max_{\pi_1\in\Pi_1}\big(V_{\Mt(x)}^{\pi_1,\pi_2^*}(\rho)\big)-V_{\Mt(x)}^{\pi_1,\pi_2^*}(\rho)\Big] \nonumber\\
        &= \mu \Big[V_{\Mt(x)}^{\pi_1^*,\pi_2}(\rho)-V_{\Mt(x)}^{\pi_1,\pi_2^*}(\rho) -\big(-\max_{\pi_1\in\Pi_1}V_{\Mt(x)}^{\pi_1,\pi_2^*}(\rho)+\min_{\pi_2\in\Pi_2}V_{\Mt(x)}^{\pi_1^*,\pi_2}(\rho)\big)\Big] \nonumber\\
        &=\mu\big(\psi(x,\pi)-\min_{\pi\in\Pi} \psi(x,\pi)\big)
    \end{align}
    where the inequality follows from Lemma \ref{lemma:gradient dominance direct param}.

\subsection{Proof of Lemma \ref{lemma:solution relation bilevel rl zero sum}}\label{sec:solution relation bilevel rl zero sum proof}
\begin{proof}
    Given $x_\lambda$, point $ \pi_\lambda$ satisfies the first-order stationary condition:
    \begin{align}
        \ip{\nabla_\pi f(x_\lambda, \pi_\lambda)+\lambda \nabla_\pi \psi(x_\lambda, \pi_\lambda)}{ \pi_\lambda - \pi'} \leq 0,~\forall \pi' \nonumber
    \end{align}
    which leads to
    \begin{align}
        \ip{\nabla_\pi \psi(x_\lambda, \pi_\lambda)}{ \pi_\lambda - \pi'} 
        &\leq -\frac{1}{\lambda}\ip{\nabla_\pi f(x_\lambda, \pi_\lambda)}{ \pi_\lambda - \pi'}\nonumber\\
        &\leq \frac{ L\| \pi_\lambda - \pi'\|}{\lambda}\leq \frac{L}{\lambda},~\forall \pi' .
    \end{align}
    Combining the above inequality with Lemma \ref{lemma:gradient dominance NI} \ref{lemma:gradient dominance NI b} yields
    \begin{align}
        \psi(x_\lambda,\pi_\lambda) \leq  \frac{L}{\lambda}.\nonumber
    \end{align}
    Define $\epsilon_\lambda \coloneqq \psi(x_\lambda,\pi_\lambda)$ then $\epsilon_\lambda \leq \delta$ by choice of $\lambda$.
    
    Since $(x_\lambda, \pi_\lambda)$ is a local solution of $\mathcal{BZ}_{\lambda p}$, it holds for any feasible $(x,\pi)$ in the region where $(x_\lambda, \pi_\lambda)$ attains its minimum that
    \begin{align}
        f(x_\lambda, \pi_\lambda)+\lambda \psi(x_\lambda,\pi_\lambda) \leq f(x,\pi)+\lambda \psi(x,\pi).
    \end{align}
    From the above inequality, it holds for any $(x,\pi)$ feasible for $\mathcal{BM}_\epsilon$ and in the region that
    \begin{align}
        f(x_\lambda, \pi_\lambda) 
        &\leq f(x,\pi)+\lambda \big(\psi(x,\pi)-\epsilon_\lambda\big) \leq f(x,\pi)
    \end{align}
    which proves the result.
    \end{proof}

\subsection{Justification of the smoothness assumption in the two-player case}\label{sec:justification of smoothness zero sum}
\begin{lemma}
    Consider the following conditions.
    \begin{enumerate}[label=(\alph*)]
        \item For any $s$, assume $|h_s(\pi_1(s))|\leq B_h$ and $\|\nabla h_s(\pi_1(s))\|\leq B_h'$ for any $\pi_1\in\Pi_1$, and;  $h_s(\pi_1(s))$ is $L_h$-Lipschitz-smooth on $\Pi_1$. Also assume this holds for player 2's policy $\pi_2\in\Pi_2$.
        \item For any $(s,a_1,a_2)$, we have for any $x\in\Xc$ that 1) $|r_x(s,a_1,a_2)|\leq B_r$, and; 2) $V_{\Mt(x)}^{\pi_1,\pi_2}(\rho)$ is $L_{vx}'$-Lipschitz-smooth on $\Xc$ uniformly for $(\pi_1,\pi_2)$.
    \end{enumerate}
\end{lemma}
Then there exists a universal constant $L_v=\mathcal{O}\Big(\frac{|\Ac|(B_r+\tau B_h)}{(1-\gamma)^3}+\frac{\tau B_h'|\Ac|}{(1-\gamma)^2}+\frac{\tau( B_h'+L_h)}{1-\gamma}+L_{vx}'\Big)$ such that $V_{\Mt(x)}^{\pi_1,\pi_2}(s)$ is $L_v$-Lipschitz-smooth on $\Xc\times\Pi_1\times\Pi_2$.
\begin{proof}
Recall the definition of the value function
\begin{align}
    V_{\Mt(x)}^{\pi_1,\pi_2}(s)=V_{\Mt(x)}^\pi(s) = \E\Big[\sum_{t=0}^\infty \gamma^t \big(r_x(s_t,a_t)-\tau h_{s_t}(\pi_1(s_t))+\tau h_{s_t}(\pi_2(s_t))\big) \big| s_0=s,\pi\Big].
\end{align}
where the expectation is taken over the trajectory generated by $a_t\sim(\pi_1(s_t),\pi_2(s_t)),s_{t+1}\sim\Pc(s_t,a_t)$. When viewing $\pi_1$ as the main policy and player 1 as the main player, we can view $(\pi_2,x)$ as the parameter of player 1's MDP, where the reward function is given by $\E_{a_2\sim\pi_2(s)}[r_x(s,a_1,a_2)]$, and the transition is $\E_{a_2\sim\pi_2(s)}[\Pc(\cdot|s,a_1,a_2)]$. To prove $V_{\Mt(x)}^{\pi_1,\pi_2}(s)$ is Lipschitz-smooth in $\pi_1$, it is then natural to use the previous results on single-agent parameterized MDP in Lemma \ref{lemma:justification of smooth value}. Specifically, we hope to use \eqref{eq:v smooth in y}.

Under the assumptions of this lemma, for \eqref{eq:v smooth in y} to hold, we additionally need to check Lemma \ref{lemma:justification of smooth value} (a). 
\begin{align}
    \sum_a \|\nabla \pi_1(a_1|s)\| = \sum_a \|\textbf{1}_{a_1,s}\|=|\Ac|,~~\nabla^2\pi(a|s)=0\text{ thus }L_y=0.
\end{align}
Then there exists constant $L_{v,1}$ that
\begin{align}
    \|\nabla_{\pi_1}V_{\Mt(x)}^{\pi_1,\pi_2}(s)-\nabla_{\pi_1}V_{\Mt(x)}^{\pi_1',\pi_2}(s)\| \leq L_{v,1}\|\pi_1-\pi_1'\|
\end{align}
where $L_{v,1}=\mathcal{O}\Big(\frac{|\Ac|(B_r+\tau B_h)}{(1-\gamma)^3}+\frac{\tau B_h'|\Ac|}{(1-\gamma)^2}+\frac{\tau( B_h'+L_h)}{1-\gamma}\Big)$, which is a uniform constant for any $\pi_2,x$. Similarly, there exists a uniform constant $L_{v,2}$ that
\begin{align}
    \|\nabla_{\pi_2}V_{\Mt(x)}^{\pi_1,\pi_2}(s)-\nabla_{\pi_2}V_{\Mt(x)}^{\pi_1,\pi_2'}(s)\| \leq L_{v,2}\|\pi_2-\pi_2'\|.
\end{align}
The above two inequalities along with assumption (b) 2) gives
\begin{align}
    \|\nabla V_{\Mt(x)}^{\pi_1,\pi_2}(s)-\nabla V_{\Mt(x')}^{\pi_1',\pi_2'}(s)\| 
    &\leq \|\nabla_{\pi_1}V_{\Mt(x)}^{\pi_1,\pi_2}(s)-\nabla_{\pi_1}V_{\Mt(x)}^{\pi_1',\pi_2}(s)\| \nonumber\\
    &+\|\nabla_{\pi_2}V_{\Mt(x)}^{\pi_1,\pi_2}(s)-\nabla_{\pi_2}V_{\Mt(x)}^{\pi_1,\pi_2'}(s)\| \nonumber\\
    &+\|\nabla_{x}V_{\Mt(x)}^{\pi_1,\pi_2}(s)-\nabla_{x}V_{\Mt(x)}^{\pi_1,\pi_2'}(s)\| \nonumber\\
    &\leq L_{vx}'\|x-x'\|+L_{v,1}\|\pi_1-\pi_1'\|+L_{v,2}\|\pi_2-\pi_2'\|.
\end{align}
Then the result holds with $L_v=\max\{L_{vx}',L_{v,1},L_{v,2}\}$.
\end{proof}

\subsection{Proof of Lemma \ref{lemma:smooth NI}}\label{sec:smooth NI proof}
\begin{proof}
Denote $\pi_1^*(x,\pi_2)\coloneqq \arg\max_{\pi_1\in\Pi_1} V_{\Mt(x)}^{\pi_1,\pi_2}(\rho)$. Treating $(x,\pi_2)$ as the parameter of player 1's MDP, then we have Lemma \ref{lemma:lipschitz continuous optimal policy} holds under Assumption \ref{asp:smoothness assumption NI}. It then follows that there exits a constant $L_\pi$ such that $\pi_1^*(x,\pi_2)$ is $L_\pi$-Lipschitz-continuous. Similar results also hold for $\pi_2^*(x,\pi_1)$.
 With Lemma \ref{lemma:gradient dominance NI} (a), we have
\begin{align}
    \|\nabla \psi(x,\pi) - \nabla \psi(x',\pi')\|^2
    &= \|\nabla_{(x,\pi_1)} V_{\Mt(x)}^{\pi_1,\pi_2^*(x,\pi_1)}(\rho)-\nabla_{(x,\pi_1)} V_{\Mt(x')}^{\pi_1',\pi_2^*(x',\pi_1')}(\rho)\|^2\nonumber\\
    &~~~~+\|\nabla_{(x,\pi_2)} V_{\Mt(x)}^{\pi_1^*(x,\pi_2),\pi_2}(\rho)-\nabla_{(x,\pi_2)} V_{\Mt(x')}^{\pi_1^*(x',\pi_2'),\pi_2'}(\rho)\|^2\nonumber\\
    &\leq L_v^2(2 \|x-x'\|^2+\|\pi_1^*(x,\pi_2)-\pi_1^*(x',\pi_2')\|^2+\|\pi_2^*(x,\pi_1)-\pi_2^*(x',\pi_1')\|^2)\nonumber\\
    &\leq 2 L_v^2(1+ L_\pi^2)\|x-x'\|+L_v^2 L_\pi^2(\|\pi_1-\pi_1'\|^2+\|\pi_2-\pi_2'\|^2).
\end{align}
which proves $\nabla \psi(x,\pi)$ is $L_v\sqrt{2(1+L_\pi^2)}$-Lipschitz continuous.
\end{proof}

\subsection{Gradient Estimator Accuracy}\label{sec:oracle accuracy zero sum justification}
We omit the iteration index $k$ here. The following arguments hold for any iteration $k$.

Recall that
$$\hat{\gr}\psi(x,\pi;\hat{\pi})= \Big(\nabla_x V_{\Mt(x)}^{\hat{\pi}_1,\pi_2}(\rho)-\nabla_x V_{\Mt(x)}^{\pi_1,\hat{\pi}_2}(\rho),\big(-\nabla_{\pi_1} V_{\Mt(x)}^{\pi_1,\hat{\pi}_2}(\rho),\nabla_{\pi_2} V_{\Mt(x)}^{\hat{\pi}_1,\pi_2}(\rho)\big)\Big)$$
and the formula of $\nabla \psi$ is (see Lemma \ref{lemma:gradient dominance NI} (a)):
$$\nabla \psi(x,\pi) = \Big(\nabla_x V_{\Mt(x)}^{\pi_1^*,\pi_2}(\rho)-\nabla_x V_{\Mt(x)}^{\pi_1,\pi_2^*}(\rho),\big(-\nabla_{\pi_1} V_{\Mt(x)}^{\pi_1,\pi_2^*}(\rho),\nabla_{\pi_2} V_{\Mt(x)}^{\pi_1^*,\pi_2}(\rho)\big)\Big)$$
where $\pi_1^*\coloneqq \argmax_{\pi_1\in\Pi_1}V_{\Mt(x)}^{\pi_1,\pi_2}(\rho)$ and $\pi_2^*$ defined similarly.
It then follows that
\begin{align}
    \|\hat{\nabla}\psi(x,\pi;\hat{\pi}) - \nabla \psi(x,\pi)\| 
    &\leq 2 \|\nabla V_{\Mt(x)}^{\pi_1^*,\pi_2}(\rho) - \nabla V_{\Mt(x)}^{\hat{\pi}_1,\pi_2}(\rho)\| + 2\|\nabla V_{\Mt(x)}^{\pi_1,\pi_2^*}(\rho)-\nabla V_{\Mt(x)}^{\pi_1,\hat{\pi}_2}(\rho)\| \nonumber\\
    &\leq 2 L_v (\|\hat{\pi}_1-\pi_1^*\|+\|\hat{\pi}_2-\pi_2^*\|)
\end{align}
where the last inequality follows from Assumption \ref{asp:smoothness assumption NI} (a). Fixing $x,\pi_2$, it takes the policy mirror descent algorithm \citep{zhan2023policy} an iteration complexity of $\mathcal{O}(-\log \epsilon)$ to solve for a $\hat{\pi}_1$ such that $\|\hat{\pi}_1-\pi_1^*\|\leq \epsilon$ (and similarly for $\hat{\pi}_2$). Therefore, the iteration complexity to guarantee Assumption \ref{asp:oracle accuracy zero sum} is $\mathcal{O}(-\log (\epsilon_{orac}/\lambda^2))$.

\subsection{Proof of Theorem \ref{theorem:PBRL convergence zero sum}}\label{sec:PBRL convergence zero sum proof}
\begin{proof}
 In this proof, we write $z=(x,\pi)$. Given $\hat{\pi}$, we also define
 \begin{align}
     \hat{\nabla} F_\lambda(z;\hat{\pi}) \coloneqq \nabla f(z) +\lambda \hat{\nabla}\psi(z;\hat{\pi}).
 \end{align}
 Thus the update we are analyzing can be written as
 \begin{align}
     z^{k+1} = \Proj_{\Zc}\big[z^k-\alpha \hat{\nabla} F_\lambda(z^k;\hat{\pi}^k) \big]
 \end{align}
 Now we start proving the result.
Under the assumptions, we have $F_\lambda$ is $L_\lambda$-Lipschitz-smooth with $L_\lambda = L_f + \lambda L_\psi$, thus it holds that
\begin{align}\label{eq:00 zs}
    F_\lambda(z^{k+1}) 
    &\leq F_\lambda(z^k) + \ip{\gr F_\lambda(z^k)}{z^{k+1}-z^k} + \frac{L_\lambda}{2}\|z^{k+1}-z^k\|^2 \nonumber\\
    &\stackrel{\alpha\leq \frac{1}{L_\lambda}}{\leq} F_\lambda(z^k) + \ip{\hat{\gr}F_\lambda (z^k;\hat{\pi}^k)}{z^{k+1}-z^k} + \frac{1}{2\alpha}\|z^{k+1}-z^k\|^2 +\ip{\gr F_\lambda(z^k)-\hat{\gr}F_\lambda (z^k;\hat{\pi}^k)}{z^{k+1}-z^k}.
\end{align}
Consider the second term in the RHS of \eqref{eq:00 zs}.
It is known that $z^{k+1}$ can be written as
$$z^{k+1}=\arg\min_{z\in\Zc}\ip{\hat{\gr}F_\lambda (z^k;\hat{\pi}^k)}{z}+\frac{1}{2\alpha}\|z-z^k\|^2.$$
By the first-order optimality condition of the above problem, it holds that
\begin{align}
    \ip{\hat{\gr}F_\lambda (z^k;\hat{\pi}^k)+\frac{1}{\alpha}(z^{k+1}-z^k)}{z^{k+1}-z} \leq 0,~\forall z \in \Zc.\nonumber
\end{align}
Since $z^k\in\Zc$, we can choose $z=z^k$ in the above inequality and obtain
\begin{align}\label{eq:bb00 zs}
    \ip{\hat{\gr}F_\lambda (z^k;\hat{\pi}^k)}{z^{k+1}-z^k} \leq -\frac{1}{\alpha}\|z^{k+1}-z^k\|^2.
\end{align}
Consider the last term in the RHS of \eqref{eq:00 zs}. By Young's inequality, we first have
\begin{align}\label{eqn.app.proof.b8 zs}
    \ip{\gr F_\lambda(z^k)-\hat{\gr}F_\lambda (z^k;\hat{\pi}^k)}{z^{k+1}-z^k} 
    &\leq \alpha \norm{\gr F_\lambda(z^k)-\hat{\gr}F_\lambda (z^k;\hat{\pi}^k)}^2 + \frac{1}{4\alpha}\norm{z^{k+1}-z^k}^2 \nonumber\\
    &\leq \alpha \lambda^2\norm{\gr \psi(z^k)-\hat{\gr}\psi (z^k;\hat{\pi}^k)}^2 + \frac{1}{4\alpha}\norm{z^{k+1}-z^k}^2
\end{align}
Substituting \eqref{eqn.app.proof.b8 zs} and \eqref{eq:bb00 zs} into \eqref{eq:00 zs} and rearranging the resulting inequality yield
\begin{align}\label{eq:trash00 zs}
\frac{1}{4\alpha}\|z^{k+1}-z^k\|^2 \leq F_\lambda(z^k)-F_\lambda(z^{k+1}) + \alpha \lambda^2 \norm{\gr \psi(z^k)-\hat{\gr}\psi (z^k;\hat{\pi}^k)}^2.
\end{align}
With $\Bar{z}^{k+1}$ defined in \eqref{eq:nonconvex proximal grad zero sum}, we have
\begin{align}\label{eqn.app-proof.b12 zs}
    \norm{\Bar{z}^{k+1}-z^k}^2 &\leq 2\norm{\Bar{z}^{k+1}-z^{k+1}}^2+2\|z^{k+1}-z^k\|^2 \nonumber\\
    &\leq 2\alpha^2 \norm{\gr F_\lambda(z^k)-\hat{\gr}F_\lambda (z^k;\hat{\pi}^k)}^2\!+\!2\|z^{k+1}\!-\!z^k\|^2 \nonumber\\
    &\leq 2\alpha^2 \lambda^2\norm{\gr \psi(z^k)-\hat{\gr}\psi (z^k;\hat{\pi}^k)}^2 + 2\|z^{k+1}\!-\!z^k\|^2
\end{align}
where the second inequality uses non-expansiveness of $\Proj_{\Zc}$.

Together \eqref{eq:trash00 zs} and \eqref{eqn.app-proof.b12 zs} imply
\begin{align}
    \norm{\Bar{z}^{k+1}-z^k}^2 \leq 10\alpha^2\lambda^2\norm{\gr \psi(z^k)-\hat{\gr}\psi (z^k;\hat{\pi}^k)}^2 + 8\alpha(F_\lambda(z^k)-F_\lambda(z^{k+1})).\nonumber
\end{align}
Since $p(x,y) \geq 0$, $F_\lambda(z) \geq \inf_{z\in\Zc} f(z)$ for any $z\in\Zc$.
Taking a telescope sum of the above inequality and using $\mathcal{G}_\lambda(z^k)=\frac{1}{\alpha}(z^k-\Bar{z}^{k+1})$ yield
\begin{align}
    \sum_{k=1}^K \|\mathcal{G}_\lambda(z^k)\|^2 
    &\leq \frac{8\big(F_\lambda(z_1)-\inf_{z\in\Zc} f(z)\big)}{\alpha}+\sum_{k=1}^K 10\lambda^2 \norm{\gr \psi(z^k)-\hat{\gr}\psi (z^k;\hat{\pi}^k)}^2\nonumber\\
    &\leq \frac{8\big(F_\lambda(z_1)-\inf_{z\in\Zc} f(z)\big)}{\alpha}+\sum_{k=1}^K \frac{1}{2} \|\mathcal{G}_\lambda(z^k)\|^2 + \frac{K}{2}\epsilon_{orac}
\end{align}
where the last inequality follows from Assumption \ref{asp:oracle accuracy zero sum}.
Rearranging gives
\begin{align}\label{eq:result nonconvex zs}
    \sum_{k=1}^K \|\mathcal{G}_\lambda(z^k)\|^2 \leq \frac{16\big(F_\lambda(z_1)-\inf_{z\in\Zc} f(z)\big)}{\alpha}+K\epsilon_{orac}.
\end{align}
This proves the theorem.
\end{proof}

\section{Additional Experiment Details}
\subsection{Stackelberg Markov game}
For the independent policy gradient method \citep{daskalakis2020independent,ding2022independent}, we set the learning rate as $0.1$, and both the follower and the leader use Monte Carlo sampling with trajectory length $5$ and batch size $16$ to estimate the policy gradient.
For the PBRL algorithms, to estimate a near-optimal policy $\hat{\pi}$ at each outer iteration, we run the policy gradient algorithm for $T$ steps at every outer iteration.
For PBRL with value penalty, we set learning rate $0.1$, penalty constant $\lambda=2$, inner iteration number $T=1$, and we use Monte Carlo sampling with trajectory length $5$ and batch size $16$ to estimate the policy gradient. For PBRL with the Bellman penalty, we use $\lambda=7$ and inner iteration number $T=10$ instead.

\subsection{Deep reinforcement learning from human feedback}
We conduct our experiments in the Arcade Learning Environment (ALE) \citep{bellemare2013arcade} by OpenAI gymnasium which is also used in \citep{A3C} and \citep{christiano2017deep}. 

For the Atari games, we use A2C, which is a synchronous version of \citep{A3C}, as the policy gradient estimator in both DRLHF and PBRL. The policy and the critic share a common base model: The input is fed through 4 convolutional layers of size $8\times8$, $5\times5$, $4\times 4$, $4\times4$, strides $4,2,1,1$ and number of filters $16,32,32,32$, with ReLU activation. This is followed by a fully connected layer of output size $256$ and a ReLU non-linearity. The output of the base model is fed to a fully connected layer with scalar output as critic, and another fully connected layer of action space size as policy. The reward predictor has the same input ($84\times84\times4$ stacked image) as the actor-critic. The input is fed through 4 convolutional layers of size $7\times7$, $5\times5$, $3\times 3$, $3\times3$, strides $3,2,1,1$ with $16$ filters each and ReLU activation. It is followed by a fully connected layer of size $64$, ReLU activation and another fully connected layer of action space size that gives the reward function. We use random dropout (probability $0.5$) between fully connected layers to prevent over-fitting (only in reward predictor). 
The reward predictor and the policy are trained synchronously. The reward predictor is updated for one epoch every $300$ A2C update.

We compare trajectories of $25$ time steps. At the start of training, we collect $576$ pairs of trajectories and warm up the reward predictor for $500$ epochs. After training starts, we collect $16$ new pairs per reward learning epoch. We only keep the last collected $3000$ pairs in a buffer.

For policy learning, we set the actor-critic learning rate $0.0003$, the entropy coefficient $0.01$, the actor-critic batch size $16$, initial upper-level loss coefficient $0.001$ which decays every $3000$ actor-critic gradient steps. We find out that the learning procedure is very sensitive to this coefficient, so we generally select this coefficient so that the upper-level loss converges stably; for reward learning, we set reward predictor learning rate $0.0003$, reward predictor batch size $64$, and the reward predictor is trained for one epoch every $500$ actor-critic gradient steps. For Beamrider, we change the actor-critic learning rate to $7\times10^{-5}$.


\subsection{Incentive design}
For the PBRL algorithms, we set the learning rate as $0.1$ and a penalty constant $\lambda=4$. The policy gradients are given by Monte Carlo sampling with trajectory length 5 and batch size 24. To obtain $\hat{\pi}_1^k,\hat{\pi}_2^k$ at each outer iteration $k$, we run the policy gradient algorithm for a single iteration with a learning rate $0.1$ at every outer iteration. For the meta-gradient method, we use the same learning rate, trajectory length and batch size as PBRL. The inner iteration number is 1.

\end{document}